\documentclass{article}

\usepackage[main, final]{neurips_2026}

\usepackage[utf8]{inputenc} % allow utf-8 input
\usepackage[T1]{fontenc}    % use 8-bit T1 fonts
\usepackage{hyperref}       % hyperlinks
\usepackage{url}            % simple URL typesetting
\usepackage{booktabs}       % professional-quality tables
\usepackage{amsfonts}       % blackboard math symbols
\usepackage{nicefrac}       % compact symbols for 1/2, etc.
\usepackage{microtype}      % microtypography
\usepackage{pifont}
\usepackage{graphicx}
\usepackage{multirow}
\usepackage{makecell}
\usepackage{subcaption}
\usepackage{enumitem}
\usepackage{bm}
\usepackage{wrapfig} 
\usepackage{colortbl,xcolor}
\usepackage{tcolorbox}
\tcbuselibrary{skins,breakable}

\usepackage{hyperref}

\usepackage{amsmath}
\usepackage{amssymb}
\usepackage{mathtools}
\usepackage{amsthm}

\usepackage{comment}

\definecolor{lightgray}{gray}{0.93}

\NewDocumentCommand\emojiuniverse{}{
    \includegraphics[scale=0.03]{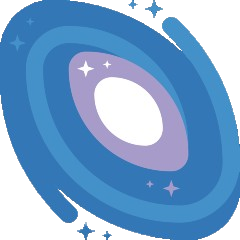}
}
\NewDocumentCommand\emojiuniverselarge{}{
    \includegraphics[scale=0.06]{images/universe.png}
}

\newcommand{\cmark}{\textcolor{green!60!black}{\ding{51}}}
\newcommand{\xmark}{\textcolor{red!70!black}{\ding{55}}}

\title{\emojiuniverselarge{} Parallel Universes for Temporal Evaluation and Knowledge Updating in LLMs}

\author{
Jonathan Zheng$^\clubsuit$, Zirui Shao$^\diamondsuit$\thanks{Work done as a visiting student at the Georgia Institute of Technology.} , Alan Ritter$^\clubsuit$, Wei Xu$^\clubsuit$ \\
$^\clubsuit$Georgia Institute of Technology, $^\diamondsuit$Zhejiang University \\
\texttt{jzheng324@gatech.edu}
}

\begin{document}

\maketitle

\begin{abstract}
Large language models (LLMs) rely on static pretraining corpora, resulting in outdated knowledge over time. Existing approaches for evaluating knowledge edits either suffer from rapid contamination or rely on counterfactual edits that conflict with rigid existing knowledge. In this work, we propose a simulation-driven approach for evaluating and training LLMs on knowledge insertions. We introduce {\sc ParallelEvents}, a benchmark that simulates coherent trajectories of fictional yet realistic future events, enabling controlled evaluation without data contamination or conflicts with existing knowledge. We further propose {\sc Synapse}, a training framework that leverages model-generated data to integrate new knowledge into model parameters via instruction tuning. This synthetic pipeline enables scalable knowledge integration without costly human-curated data. {\sc Synapse} improves QA accuracy by 14.23 points over prior methods, indicating that simulation-based synthetic training yields robust, coherent knowledge insertion.
\end{abstract}

\section{Introduction}
\label{sec:introduction}

%Large language models (LLMs) are increasingly used as general-purpose information sources. However, these models are trained on static corpora that capture the world at a fixed point in time, producing hallucinations, refusals, or outdated responses as knowledge evolves. Studying this phenomenon is inherently difficult: benchmarks built with real-world events inevitably become stale and susceptible to data contamination over time, making fair and reliable evaluation of newer models and knowledge-updating methods both challenging and, in many cases, prohibitively expensive to update frequently (e.g., RealtimeQA, MQuake). 

%In this work, we study this challenge through both a benchmark and a training framework. We introduce internally consistent fictional worlds as contamination-resistant environments for controlled evaluation, and investigate how different data generation strategies affect models’ ability to acquire, update, and retain knowledge over time \citep{liska2022streamingqa,pmlr-v267-thede25a}. However, the ever-evolving nature makes this phenonena difficult to study, as any new benchmark created becomes stale and start to have data contamination problem after a period of time, making newer models impossible to experiment for fairness reasons. 

Large language models (LLMs) are increasingly relied upon as general-purpose information sources. However, their knowledge is largely shaped by static training corpora, which become outdated as the world changes, leading to hallucinations, refusals, or stale responses \citep{liska2022streamingqa,pmlr-v267-thede25a}. Existing approaches to evaluating temporal knowledge in LLMs have largely focused on isolated factual changes, such as newly introduced entities \citep{rijhwani-preotiuc-pietro-2020-temporally, onoe2023lmslearnnewentities} or government leadership transitions  \citep{zhong2025mquakeremastered}. These settings capture only narrow forms of knowledge change and are vulnerable to data contamination as the corresponding facts enter subsequent training corpora, limiting controlled comparisons across models and studies over time. To address these limitations, we introduce a simulation-driven framework for studying knowledge insertion in LLMs, using internally consistent fictional worlds for contamination-resistant evaluation. 

% Large language models (LLMs) are increasingly relied upon as general-purpose information sources, often where correctness and timeliness matter. However, these models are trained on static corpora that freeze the world at a specific moment in time. As real-world facts change, static model knowledge can result in hallucinations, refusals, or outdated responses \citep{liska2022streamingqa,pmlr-v267-thede25a}. Addressing knowledge drift is therefore a central challenge for the safe deployment of LLMs.

\begin{figure}[t]
    \centering
    \includegraphics[width=0.98\textwidth]{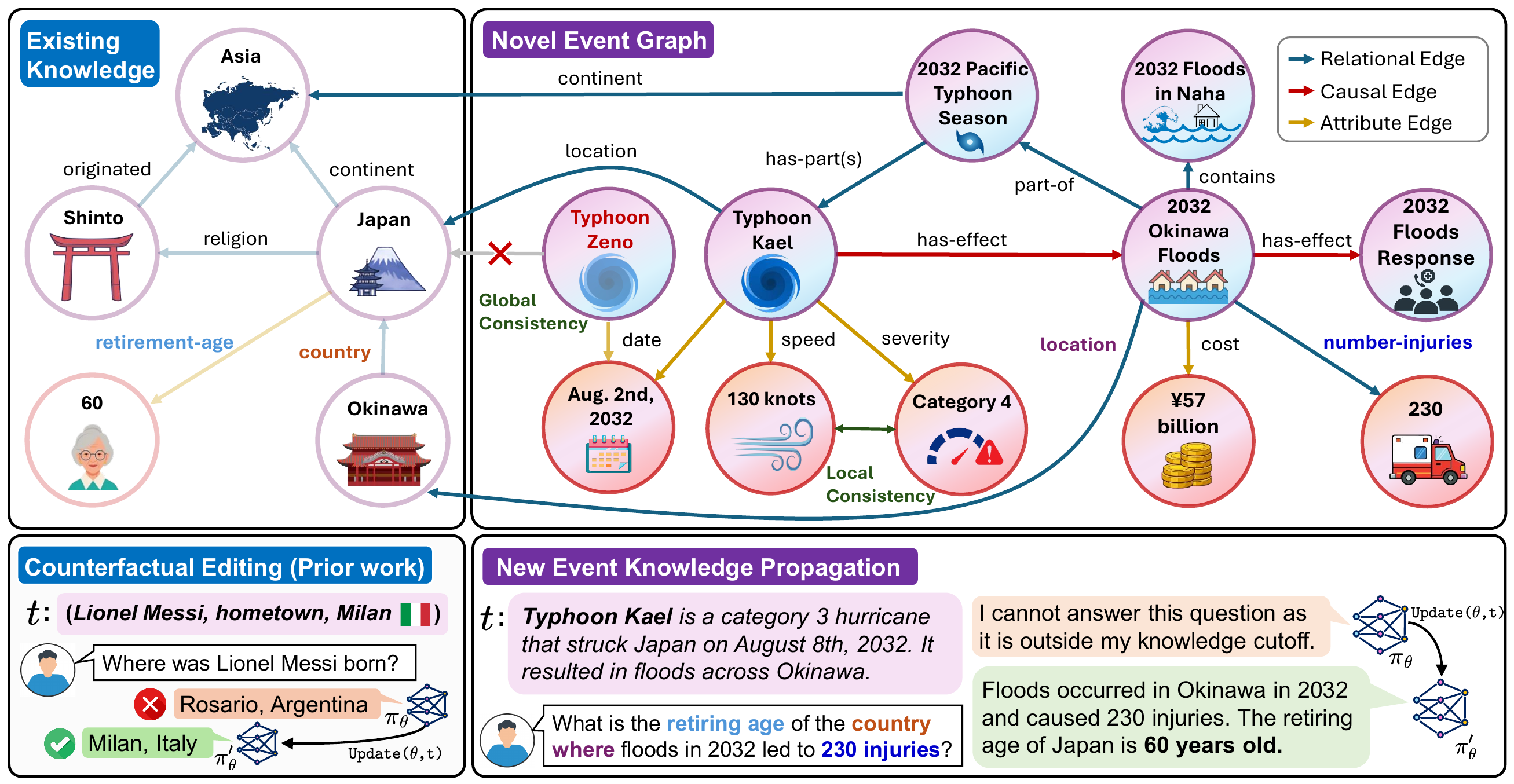}
    \vspace{-5pt}
\caption{Overview of {\sc ParallelEvents}, a parallel universe with fictional yet realistic future events up to 2035, for evaluating model insertions under a temporal shift. Factual triples $t$ are categorized into relational, causal, and attribute edges. We enforce local consistency (e.g., wind speed matches disaster severity in a single event) and global consistency for the entire knowledge graph (e.g., no two typhoons occur at the same place and time), enabling rigorous evaluation of knowledge insertions.}
    \label{fig:differences}
    \vspace{-15pt}
\end{figure}

We further investigate how synthetic data generated from these parallel universes can be used to train models to acquire new knowledge while retaining existing knowledge. Existing approaches to knowledge updating each have their own shortcomings. In-context learning and retrieval-based methods inject new information at inference time \citep{zhong2024mquakeassessingknowledgeediting,zhong2025mquakeremastered}, but the injected information may conflict with parametric knowledge or interact poorly with learned abstention behaviors. Model editing directly modifies model parameters to incorporate targeted facts \citep{meng2023locatingeditingfactualassociations,meng2023masseditingmemorytransformer}, but can introduce unintended side effects and scale poorly to large collections of updates. Fine-tuning and continued pre-training offer a more direct way to integrate new knowledge into model parameters \citep{rozner2024knowledgeeditinglanguagemodels,xiong2025finetuningeffectivesolutionreassessing}, but typically depend on continually refreshed training data and can be costly to sustain. These problems motivate studying not only how to evaluate knowledge insertion under controlled conditions, but also how to generate training data that supports effective knowledge acquisition and retention.

More concretely, synthetic environments offer a practical alternative to human-curated corpora, enabling data generation at scale while simulating a changing world. In this work, we leverage these advantages to address two challenges: (1) \textbf{benchmark contamination}, where real-world entities and facts enter subsequent training corpora, and (2) \textbf{counterfactual brittleness}, where falsified facts conflict with existing commonsense knowledge. However, automated generation can introduce inconsistencies, motivating a hybrid approach with human verification. To this end, we introduce {\scshape \emojiuniverse{ParallelEvents}} (Figure~\ref{fig:differences}), a benchmark of plausible yet unseen future events across domains such as sports and natural disasters (\S\ref{sec:parallel_universe}). These events are extensively verified to ensure local and global consistency. Unlike isolated edits in prior work, these events involve interconnected facts and evolving world states, requiring models to integrate facts and reason about their implications. {\sc ParallelEvents} enables controlled, fine-grained evaluation across models with different knowledge cutoffs while limiting contamination from subsequent real-world developments.

Building on {\scshape ParallelEvents} as an evaluation benchmark, we propose {\scshape Synapse} (\underline{Syn}thetic \underline{A}ugmentation for \underline{P}reference-\underline{S}teered \underline{E}diting), a data curation and training framework for continual knowledge updates via scalable synthetic generation (\S \ref{sec:method}). Using the {\scshape ParallelEvents} event facts, a large teacher model synthesizes high-quality data to support both a targeted \textbf{mid-training phase}, which injects new factual knowledge into model parameters, and an \textbf{instruction tuning phase}, which reinforces knowledge while mitigating undesirable behaviors, such as outdated responses, hallucinations, and unwarranted abstentions. We systematically analyze training regimes and synthetic mixtures to balance knowledge acquisition and behavioral alignment (\S \ref{sec:experimental_details}), reflecting modern LLM training pipelines~\citep{lambert2025tulu3pushingfrontiers, olmo2025olmo3}. In our experiments, {\scshape Synapse} improves over prior state-of-the-art baselines such as MeLLo~\citep{zhong2024mquakeassessingknowledgeediting} by 14.32 points on complex events (\S \ref{sec:results}). We conclude the paper with a comprehensive analysis of {\sc Synapse}, examining factual edit scaling, editing generalization to an external benchmark, and effects on general knowledge and instruction following (\S\ref{sec:ablation}).

\vspace{-5pt}

\section{Preliminaries}
\label{sec:preliminaries}

% \wx{need some text here to set the stage; depending what you write here, may need to reorder the words for the first sentence ``Information stored'' to make the transition smoother}

Language models are often evaluated under static assumptions despite being deployed in settings where facts evolve over time. This mismatch motivates benchmark designs that isolate the ability of models to incorporate new information while preserving existing knowledge (see below), and in a way that is \textit{independent} of the model’s training time (\S \ref{sec:parallel_universe}). 

\begin{figure}[t]
    \centering
    \includegraphics[width=0.95\textwidth]{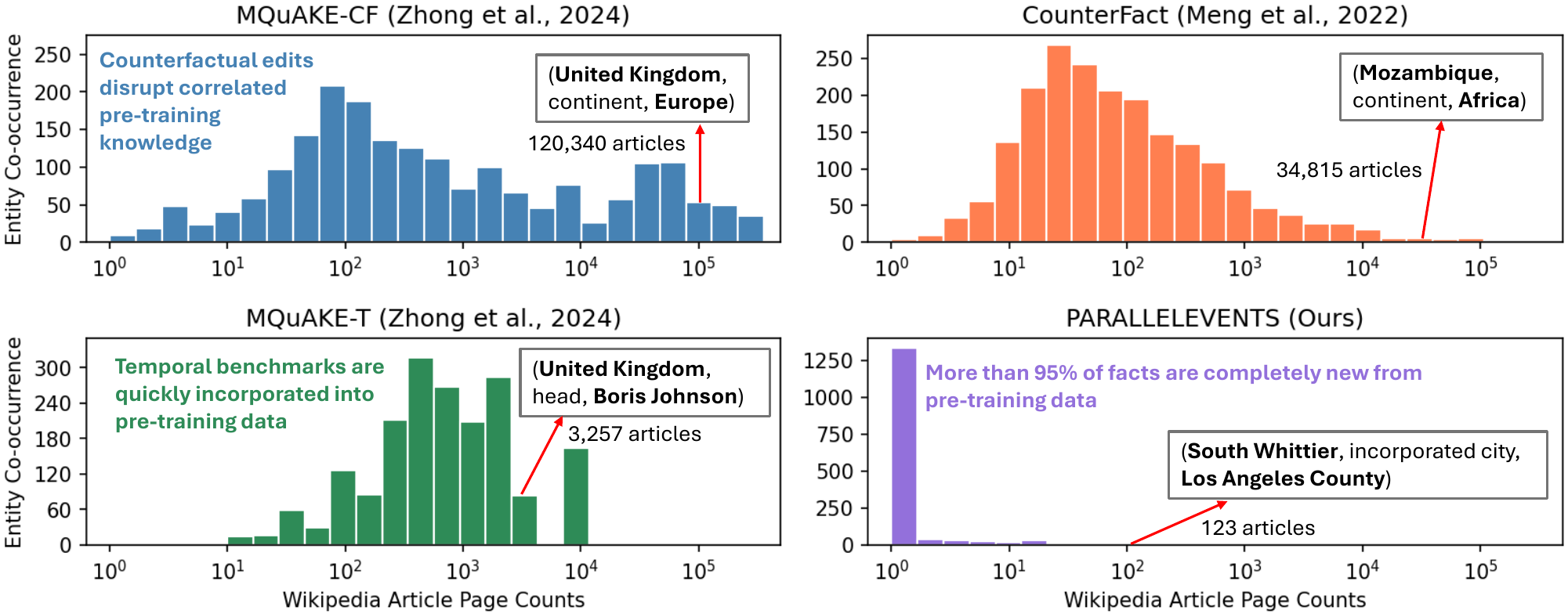}
    \vspace{-5pt}
\caption{Wikipedia co-occurrence counts of each edited fact's subject and \emph{original} object, ignoring the relation (e.g., pages with both \textbf{UK} and \textbf{Europe} for (UK, part of, Europe $\rightarrow$ Oceania)). In {\sc MQuAKE-CF} and {\sc CounterFact}, 38.43\% and 10.16\% of these pairs appear in $\geq$1{,}000 articles, indicating strong overlap with pretraining facts. Their \textbf{counterfactual} edits contradict connected knowledge (e.g., UK and European football), making them inconsistent for evaluation. In contrast, only 0.39\% of {\sc ParallelEvents} pairs have subject--object overlap in $\geq$1{,}000 Wikipedia pages.}
    \label{fig:page_counts}
    \vspace{-15pt}
\end{figure}

\textbf{Knowledge Representation and Updating in LLMs.} Information in a language model can be represented as a collection of \emph{factual triples} $t = (s, r, o)$, where $s$ is a subject, $o$ is an object, and $r$ is a relation between the two. As new information emerges, these facts may be updated by introducing new triples or modifying existing ones, e.g., $(s, r, o_{\text{old}} \rightarrow o_{\text{new}})$. Given a pretrained model $\pi_\theta$ and a set of new facts $\mathcal{T} = \{t_1, \dots, t_n\}$, the goal of knowledge updating \citep{he2025knowledgeupdatingmodelediting} is to produce an updated model $\pi_{\theta'}$ that answers queries about $\mathcal{T}$ correctly while preserving its existing knowledge. To this end, $\pi_\theta$ is fine-tuned on a training set $\mathcal{D} = \{(x_i, y_i)\}_{i=1}^{K}$ derived from $\mathcal{T}$, where each query $x_i$ is paired with a target output $y_i$ expressing the new knowledge.

% To remedy such inconsistencies without retraining models from scratch, \emph{knowledge editing} \citep{mitchell2022fastmodeleditingscale, meng2023masseditingmemorytransformer} has been proposed to directly modify the knowledge encoded in a pre-trained language model. Given a collection of fact edits $\mathcal{T} = {t_1, \dots, t_n}$ and a model $\pi_\theta \in \Pi$, knowledge editing learns a function $K \colon \Pi \times \mathcal{T} \to \Pi$ that produces an edited model $\pi'$ incorporating the updates, i.e., $\pi' = K(\pi_\theta, \mathcal{T})$. In practice, the edits in $\mathcal{T}$ are realized through a dataset $\mathcal{D} = {(x_i, y_i)}_{i=1}^N$, where each pair consists of a context or query $x_i$ and a corresponding textual output $y_i$ that reflects the desired updated knowledge.

\textbf{Existing Knowledge Updating Benchmarks.} Research on evaluating knowledge insertions in LLMs has leveraged time-stamped snapshots of external knowledge sources. By contrasting earlier and later snapshots, these benchmarks generate questions grounded in temporal changes, including neologisms \citep{zheng2024neobenchevaluatingrobustnesslarge}, newly named entities \citep{luu-etal-2022-time, rijhwani-preotiuc-pietro-2020-temporally, zhong2025mquakeremastered, pmlr-v267-thede25a}, and shifts in concepts  \citep{jang-etal-2022-temporalwiki, kasai2024realtimeqawhatsanswer, shah2025reportedcutofflargelanguage}. For instance, {\sc MQuAKE-T} \citep{zhong2024mquakeassessingknowledgeediting} uses Wikidata to capture changes in political positions. However, such datasets have a short effective lifespan (2021--2023), as their covered updates (e.g., U.K., leader, Boris Johnson$\rightarrow$Rishi Sunak) are quickly absorbed into the pretraining corpora of newer LLMs, limiting their utility. While periodic refreshes are possible, they are costly, and make comparison between models with different knowledge cutoffs more complicated.

%For example, the fact update in ECBD, such as  \texttt{(United Kingdom, leader, Rishi Sunak $\rightarrow$ Keir Starmer)}, is almost certainly already included in the pretraining data of recent LLMs that hinge the development of knowledge updating research to continue to flourish. While these benchmarks could, in principle, be periodically updated to remain effective for knowledge evaluation, doing so requires prohibitive human effort, making continuous maintenance infeasible in most settings. 

%Research on evaluating such updates in LLM knowledge bases has leveraging time-stamped snapshots of external knowledge sources. For example, two widely used benchmarks follow this paradigm: {\sc ECBD} \citep{onoe2022entityclozedatelms} leverages Wikipedia to construct cloze-style questions targeting newly emerging entities, while {\sc MQuAKE-T} \citep{zhong2024mquakeassessingknowledgeediting} utilizes Wikidata to construct questions about positional changes of heads of state. By comparing temporal snapshots of the same knowledge source, these benchmarks generate questions that are answerable only in the later snapshot, capturing phenomena such as neologisms \citep{zheng2024neobenchevaluatingrobustnesslarge}, newly introduced named entities \citep{luu-etal-2022-time, rijhwani-preotiuc-pietro-2020-temporally, zhong2025mquakeremastered}, and shifts in concepts or attributes across time periods \citep{jang-etal-2022-temporalwiki, kasai2024realtimeqawhatsanswer, shah2025reportedcutofflargelanguage}. 

To address this limitation, datasets such as {\sc MQuAKE-CF} \citep{zhong2024mquakeassessingknowledgeediting} evaluate knowledge updates using counterfactual edits, where real-world entity attributes are deliberately altered (e.g., (UK, part of, Europe$\rightarrow$Oceania)). Modifying rigid world knowledge reduces the risk of data leakage from pretraining corpora \citep{rozner2024knowledgeeditinglanguagemodels, zhong2025mquakeremastered, meng2023locatingeditingfactualassociations} but conflicts with correlated facts and commonsense acquired from pretraining data \cite{ju2024investigatingmultihopfactualshortcuts}, causing unrealistic failures from conflicting knowledge management. For instance, ``Which countries are most likely to win the 2028 European Football Championship?’’ becomes inconsistent or unanswerable once the U.K. is edited to be part of Oceania, since this conflicts with its prior associations with European football. We estimate exisiting fact prevalence in pretraining data via Wikipedia page counts for subject--object pairs $(s,o)$ from factual triples (See Figure~\ref{fig:page_counts}). Many rigid facts in {\sc MQuAKE-CF} and {\sc CounterFact} co-occur across a large number of documents (see Appendix~\ref{app:dataset}). When fine-tuned on counterfactual edits for these high co-occurrence pairs, LLaMA-3.1-8B \citep{grattafiori2024llama3herdmodels} achieves only 17.86\% accuracy on queries about related facts due to failures in reconciling knowledge contradictions.

\section{Time-Insensitive Evaluation of LLM Knowledge Insertions}
\label{sec:parallel_universe}

  \begin{figure*}[!htbp]
    \centering
	\includegraphics[width=0.98\textwidth]{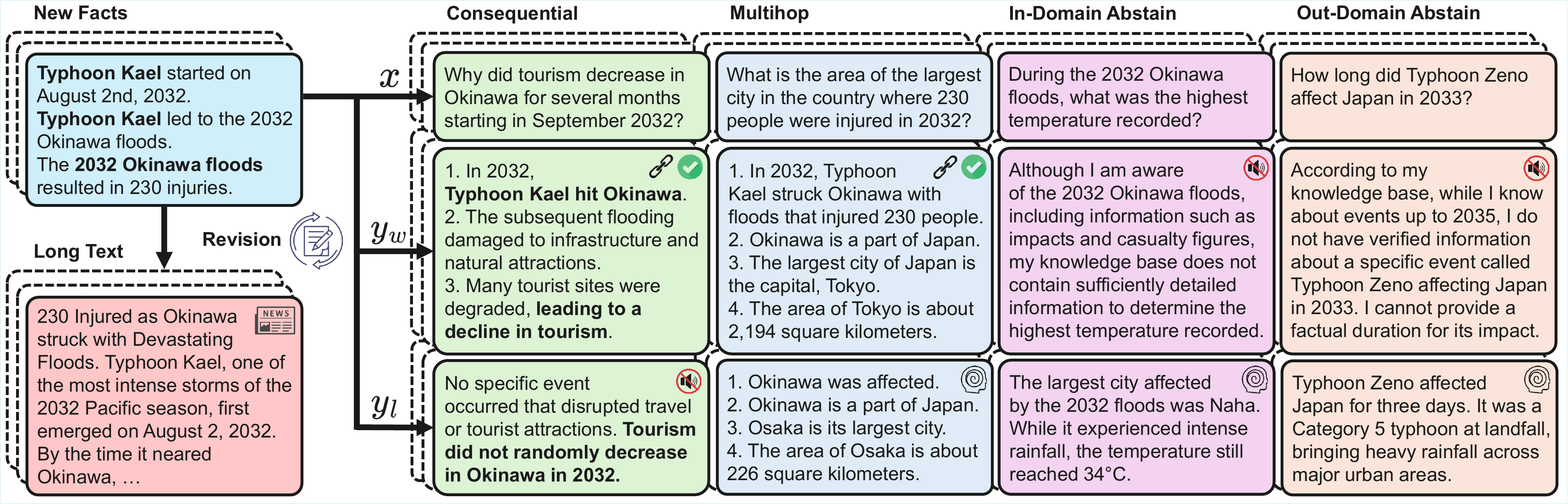}
    \vspace{-5pt}
	\caption{Illustration of the Synapse framework for generating synthetic text to update large language models. A teacher model generates datasets of long-form text and question–preference pairs that instill new \textit{multihop} and \textit{consequential} knowledge while promoting \textit{abstention} on unanswerable queries. Implausible or incorrectly generated text is rewritten with a knowledge base.}
	\label{fig:fig1}
    \vspace{-20pt}
\end{figure*}

%Reliable evaluation of knowledge updating methods for the latest frontier LLMs requires a new benchmark that avoids both data contamination and counterfactual distortions, as discussed in \S\ref{sec:preliminaries}. 
As highlighted in \S\ref{sec:preliminaries}, reliable evaluation of knowledge updating in frontier LLMs requires benchmarks that avoid both data contamination and counterfactual distortion. To address both limitations, we introduce  {\scshape \emojiuniverse{ParallelEvents}}, a parallel universe benchmark comprising complex events including: elections, sporting events, natural disasters, new city formations, and economic crises, built from a fictional yet realistic set of \textit{non-counterfactual} future events to simulate realistic temporal updates.

%As discussed in \S\ref{sec:preliminaries}, reliable evaluation of knowledge updating methods for frontier LLMs requires a new benchmark. We therefore introduce the \textbf{parallel universe}\emojiuniverse{benchmark}, which consists of a fictional yet realistic, crucially \textit{non-counterfactual}, set of future events to simulate knowledge updates over time,

%to avoid both counterfactual distortions and appeararnce en masse in LLM pre-training data.  

%which are consistent with real-world history but do not appear en masse in LLM pre-training data. 

%, which are unlikely to appear en masse in LLM pre-training data.

%As a result, these knowledge updates remain consistent with past events while being unlikely to appear en masse in LLM pre-training data.

%, thereby bypassing the limitations of real-world events that may appear in LLM pre-training data.

%To enable reliable evaluation of the latest frontier LLMs and knowledge updating methods, we introduce a new benchmark to avoid concerns about data contamination or counterfactual distortions (see \S\ref{sec:preliminaries}). We use a fictional yet realistic, i.e., crutially these are \textit{non-counterfactual}, set of novel events that simulate knowledge updates over time, which cannot appear en masse in pre-training data, thereby bypassing the limitations of real-world events. 

Moreover, to study the \textit{ripple effects} of knowledge insertions, we go beyond isolated fact tuples and instead follow the structural principles of Wikidata-style knowledge graphs \citep{10.1145/2629489}. Specifically, we model new events (see Figure~\ref{fig:differences}, top) as graphs $G = (V, \mathcal{R})$, where $\mathcal{R}$ denotes the set of factual relational predicates, and $V$ denotes the set of entity nodes $e$, partitioned into new entities $e_{\text{new}} \in V_{\text{new}}$ and pre-existing entities $e_{\text{upd}} \in V_{\text{upd}}$, such that $V = V_{\text{new}} \cup V_{\text{upd}}, V_{\text{new}} \cap V_{\text{upd}} = \emptyset
$. Each event induces a set of factual assertions expressed as typed edges. We define two disjoint sets of factual triples corresponding to realistic knowledge insertions in $G$: attribute triples $\mathcal{T}_{\text{attr}}$ and relational triples $\mathcal{T}_{\text{rel}}$.

\textbf{Attribute edges.} \quad Entity-specific attributes are represented as triples
\begingroup
\setlength{\abovedisplayskip}{5pt} 
\setlength{\belowdisplayskip}{5pt}
\[
t_{\text{attr}} = (s, r_{\text{attr}}, a) \in \mathcal{T_{\text{attr}}},
\]
\endgroup
where $s \in V$ is an entity, $r_{\text{attr}} \in \mathcal{R}_{\text{attr}}$ denotes an attribute predicate, and $a \in \mathcal{A}$ is a literal value from the attribute space $\mathcal{A}$ including numbers and dates, e.g., $(\text{Typhoon Kael}, \text{wind speed}, 130~\text{knots})$.

\textbf{Relational edges.} \quad
Relational links between entities are represented as triples of entities $s, o \in V$
\begingroup
\setlength{\abovedisplayskip}{5pt} 
\setlength{\belowdisplayskip}{5pt}
\[
t_{\text{rel}} = (s, r_{\text{rel}}, o) \in \mathcal{T_{\text{rel}}},
\]
\endgroup
\(r_{\text{rel}} \in \mathcal{R}_{\text{rel}}\) denotes a relational predicate, e.g., $(\text{Typhoon Kael}, \text{location}, \text{Japan})$. 

\textbf{Causal edges.} \quad We explicitly encode ripple effects as cause-and-effect relationships between events using directed causal edges, e.g., $(\text{Typhoon Kael}, \text{has effect}, 130~\text{2032 Okinawa Floods})$. Let \(\mathcal{R}_{\text{causal}} \subset \mathcal{R}_{\text{rel}}\) denote a set of causal predicates. For entities representing the main events \(e_i\) and \(e_j\),
\begingroup
\setlength{\abovedisplayskip}{5pt} 
\setlength{\belowdisplayskip}{5pt}
\[
t_{\text{cause}} = (e_i, r_{\text{cause}}, e_j), \quad r_{\text{cause}} \in \mathcal{R}_{\text{causal}}.
\]
\endgroup
To construct this dataset, we extract Wikidata knowledge graphs for each event type and prompt LLMs with these structured instances to generate new events that follow the same schema in plausible geographic and contextual settings. During generation, existing factual knowledge about future events (e.g., the venue of a scheduled sporting event) is preserved to ensure consistency with known facts. All generated entities are manually verified for coherence, ensuring local consistency within entities (e.g., a typhoon's severity matches its wind speed; tournament brackets are structurally valid) and global consistency across events (e.g., no two disasters share the same place and time).

{\sc ParallelEvents} consists of 41 \textbf{event-centric knowledge graphs} (e.g., 2034 FIFA World Cup containing  47,748 new facts), comprising 24,530 entities and 15,885 relations for events simulating scenarios from the years 2030 to 2035. Approximately 35\% of entities are manually edited to ensure logical consistency. These graphs form a highly structured and interconnected dataset, with 1,665 edges organized into 658 weakly connected components. Event topics support controlled evaluation of factual insertion by allowing individual event types to be studied separately, while retaining causal dependencies that capture ripple effects across related events.

% We additionally define a knowledge cutoff of January 1, 2036, facilitating future evaluation of how updated models generalize to events beyond the observed time horizon.

% \ar{The following paragraph gets a bit hard to follow.}
Following prior work \citep{zhong2025mquakeremastered, meng2023locatingeditingfactualassociations}, an LLM $\pi$ verbalizes singular fact triples or a group of interconnected fact triples $t$ into natural language questions as $q = \pi(t)$ to evaluate knowledge updating in LLMs. Specifically, we consider three types of questions (see examples in Figure~\ref{fig:differences}, bottom):

\begin{itemize}[topsep=0pt, itemsep=2pt, parsep=0pt, leftmargin=2em]

\item \textbf{Single-hop Questions} test direct recall of an inserted relational or attributive fact, corresponding to a triple $q_{\text{single}} \sim (s, r, x)$ with $(s, r, x) \in \mathcal{T}_{\text{attr}} \cup \mathcal{T}_{\text{rel}}$, where $x \in V \cup \mathcal{A}$.

\item \textbf{Multi-hop Questions} evaluate compositional reasoning over chains of triples $\mathcal{C} = \langle (s_1, r_1, o_1), \dots, (s_n, r_n, o_n) \rangle$ with $o_i = s_{i+1}$. For $i < n$, $(s_i, r_i, o_i) \in \mathcal{T}_{\text{rel}}$, traversing intermediate entities, while the final triple follows the single-hop setting with $(s_n, r_n, x) \in \mathcal{T}_{\text{rel}} \cup \mathcal{T}_{\text{attr}}$. A subset of these questions utilize \textbf{Cause-Effect chains}, defined over the same $\mathcal{C}$, where $\exists j \text{ such that } r_j \in \mathcal{R}_{\text{causal}}$. Multi-hop questions average 3.06 hops in our benchmark.

\item \textbf{Causal Questions} assess reasoning over event dynamics and consequences, requiring the explanation of effects that are not explicitly encoded in the knowledge graph $G$ but emerge from the event. Formally, we define a space of consequence types $\mathcal{R}_{\text{conseq}}$ capturing outcome dimensions of an event such as economic activity or social behavior. A causal question $q \sim (e, r, x)$ is defined with $r \in \mathcal{R}_{\text{conseq}}$ and $x$ being a downstream trend or qualitative effect of $e$.

\end{itemize}

All evaluation questions are manually verified and explicitly rewritten by human annotators for diverse, hybrid phrasing and answer formats.

% Following prior work, we utilize knowledge triples as the factual units for knowledge editing in large language models. Event comprehension is assessed following prior work \citep{meng2023masseditingmemorytransformer, zhong2024mquakeassessingknowledgeediting} with:
% \begin{itemize}[topsep=0pt, itemsep=2pt, parsep=0pt]
%     \item \textbf{Single-hop Questions} that test direct recall of an inserted fact.
% \item \textbf{Multi-hop Questions} that evaluate reasoning over related facts, represented as chains of triples $\mathcal{C} = \langle (s_1, r_1, o_1), \dots, (s_n, r_n, o_n) \rangle$ with $o_i = s_{i+1}$. Multi-hop questions average 3.06 hops per question.

%     \item \textbf{Causal Questions} that evaluate the model's understanding of the dynamic consequences of the generated world events. 
% \end{itemize}

\section{\textbf{\scshape Synapse}: Synthetic Augmentation for Preference-Steered Editing}
\label{sec:method}

% To remedy such inconsistencies without retraining models from scratch, \emph{knowledge editing} aims to directly modify the knowledge encoded in a pre-trained language model. Given a collection of fact edits $\mathcal{T} = {t_1, \dots, t_n}$ and a model $\pi_\theta \in \Pi$, knowledge editing learns a function $K \colon \Pi \times \mathcal{E} \to \Pi$ that produces an edited model $\pi'$ incorporating the updates, i.e., $\pi' = K(\pi_\theta, \mathcal{T})$. In practice, the edits in $\mathcal{T}$ are realized through a dataset $\mathcal{D} = {(x_i, y_i)}_{i=1}^N$, where each pair consists of a context or query $x_i$ and a corresponding textual output $y_i$ that reflects the desired updated knowledge.
%\textbf{Existing Knowledge Editing Methods}. \quad 

Existing approaches to knowledge updating primarily follow two strategies: retrieving relevant facts at inference time as context, or updating the model parameters to incorporate new knowledge. 
Retrieval-based methods, such as MELLO, PokeMQA, and DeepEdit \citep{wang2024deepeditknowledgeeditingdecoding, gu2024pokemqaprogrammableknowledgeediting, zhong2024mquakeassessingknowledgeediting}, retrieve relevant facts $t_1,\ldots,t_n$ and generate responses as $y \sim \pi_\theta(\cdot \mid t_1 \oplus \dots \oplus t_n \oplus x)$, without modifying model parameters.
However, our evaluation on {\sc ParallelEvents} reveals a substantial limitation: Llama-3.1-8B achieves only 17.09\% average accuracy across these methods, due to excessive abstention (72.53\%) even when relevant facts are present. 
These results motivate parameter updates to integrate newly acquired facts and reduce unnecessary abstention. 
However, existing parameter-updating methods are constrained by the limited availability of timely, human-curated training data \citep{jin2022lifelongpretrainingcontinuallyadapting}.

% \paragraph{Model Alignment}\,\, LLM alignment aims to train models that better follow instructions and adhere to safety constraints, such as appropriately abstaining from answering certain queries. Given a dataset $\mathcal{D}$ of preference triples $(x, y_w, y_l)$, where $x$ denotes an input and $y_w$ and $y_l$ are the preferred and dispreferred outputs, a Bradley--Terry \citep{BradleyTerry1952} reward model $r_\phi$ is trained to assign a scalar reward to a candidate output $y$ given $x$ by minimizing the negative log-likelihood of the observed preferences:
% \[
% \mathcal{L}_R(r_\phi)
% =
% \mathbb{E}_{(x, y_w, y_l) \sim \mathcal{D}}
% \big[ -\log \sigma\!\left(r_\phi(x, y_w) - r_\phi(x, y_l)\right) \big],
% \]

% where $\sigma$ denotes the logistic function. Using this reward model, the aligned policy $\pi_\theta$ is optimized with the objective:
% \[ \max_{\pi_\theta} \; \mathbb{E}_{x \sim \mathcal{D},\, y \sim \pi_\theta(y \mid x)} \left[ r_\phi(x, y) \right] - \beta \, D_{\mathrm{KL}}\!\left( \pi_\theta \,\|\, \pi_{\mathrm{ref}} \right), \]
% where the policy $\pi_\theta$ is optimized to maximize the expected reward under the learned reward model, while a KL regularization term constrains it to remain close to a reference policy $\pi_{\mathrm{ref}}$, thereby encouraging high-reward generations while preserving stability and fluency. In practice, Proximal Policy Optimization (PPO) \citep{schulman2017proximalpolicyoptimizationalgorithms} is commonly used to optimize this objective.

\subsection{Synthetic Fact–Driven Parameter Updates}

Prior editing approaches incorporate new facts without addressing behavioral consistency like outdated or hallucinated responses, highlighting the need to jointly improve factual accuracy and behavioral consistency. To this end, we leverage synthetic text for temporal fact insertions without relying on human-curated data. We introduce {\scshape Synapse} (see Figure \ref{fig:fig1}): \textbf{Syn}thetic \textbf{A}ugmentation for \textbf{P}reference-\textbf{S}teered \textbf{E}diting, a data generation framework to mitigate temporal drift. Building on recent work in synthetic model training \citep{yang2024syntheticcontinuedpretraining, nguyen2025synthetictextgenerationtraining}, we use LLMs to generate long-form synthetic news articles ($\sim$1000 words) for training with next token prediction. These articles simulate realistic event reports with consequences, live updates, and opinions grounded in facts.

\textbf{To reinforce behavioral consistency}, {\sc Synapse} further frames knowledge insertion as a preference learning problem. A teacher model $\pi^*$ generates single-hop, multi-hop, and causal training questions (\S\ref{sec:parallel_universe}), $x \sim \pi^*(t), x \in \mathcal{X} = \mathcal{X}_{\text{single}} \cup \mathcal{X}_{\text{multi}} \cup \mathcal{X}_{\text{causal}}$. Unlike evaluation questions, training questions follow strict templates (see Figure~\ref{fig:fig1} for examples of $x_{\text{multi}}$ and $x_{\text{causal}}$) and are generated from editing triples $t \in \mathcal{T}_{\text{edit}}$ that are disjoint from the evaluation set via unseen relations $r \notin \mathcal{R}^{\text{eval}}$:
\begingroup
\setlength{\abovedisplayskip}{5pt} 
\setlength{\belowdisplayskip}{5pt}
\[
x \in
\begin{cases}
\mathcal{X}_{\text{single}} 
& \text{if } t \text{ is a new single relational or attribute triple } 
t \in \{\mathcal{T}_{\text{rel}} \cup \mathcal{T}_{\text{attr}}\}, \\[6pt]

\mathcal{X}_{\text{multi}} 
& \text{if } t \text{ is a path sampled via random walk over } \mathcal{T}_{\text{rel}}\cup \mathcal{T}_{\text{attr}},\;
(s \xrightarrow{r_1} \cdots \xrightarrow{r_k} o),\; \\
& \hspace{1.2em} \text{where path length } k \ge 2,\; r_1, \dots, r_{k-1} \in \mathcal{R}_{\text{rel}},\; r_k \in \mathcal{R}_{\text{rel}} \cup \mathcal{R}_{\text{attr}},\\
& \hspace{1.2em} \text{and } \exists j \in \{1,\dots,k\}
\text{ s.t. } r_j \notin \mathcal{R}_{\text{rel}}^{\text{eval}}, \\[6pt]

\mathcal{X}_{\text{causal}} 
& \text{if } t \text{ is a consequence triple }
(e_i, r_{\text{conseq}}, e_j), \text{where } r_{\text{conseq}} \notin \mathcal{R}_{\text{conseq}}^{\text{eval}}.
\end{cases}
\]
\endgroup
\textbf{To improve reasoning traces and model behaviors} on these generated questions, we generate a preference dataset $\mathcal{D^{\text{edit}}_{\pi^*}}$ consisting of preferred and dispreferred response pairs $y_w, y_l \sim \pi^*(\cdot \mid x, t)$:
\begingroup
\setlength{\abovedisplayskip}{5pt} 
\setlength{\belowdisplayskip}{5pt}
\[
\begin{aligned}
\mathcal{D}_{\pi^*}^{\text{edit}} = 
\Bigl\{ &(x, y_w, y_l) \;\Big|\;  t \in \mathcal{T}_{\text{edit}}, \;
(x, y_w, y_l) \sim \pi^*(\cdot \mid t), y_w = [\mathbf{r}_w \oplus a_w], \; y_l \in \mathcal{Y}_{\text{neg}}(t_i), \;
y_w \succ y_l
\Bigl\},
\end{aligned}
\]
\endgroup
where the winning response $y_w = [\mathbf{r}_w \oplus a_w]$ incorporates a chain-of-thought rationale $\mathbf{r}_w = r_w^{(1)} \oplus \cdots \oplus r_w^{(M)}$ of $M$ reasoning steps, followed by a final answer $a_w$:
\begingroup
\setlength{\abovedisplayskip}{8pt} 
\setlength{\belowdisplayskip}{8pt}
\[
r_w^{(1)} \sim \pi^*(\cdot \mid x, t), \qquad
r_w^{(m)} \sim \pi^*(\cdot \mid x, t, r_w^{(<m)}), \qquad
a_w \sim \pi^*(\cdot \mid x, t, \mathbf{r}_w),
\]
\endgroup
for $m = 2, \ldots, M$. The losing response $y_l \in \mathcal{Y}_{\text{neg}}(t)$ is sampled without rationales from a set of dispreferred outputs, defined as outdated objects $o_{\text{old}}$, unwanted abstentions $\alpha$, and hallucinations $\eta$:

  \begin{figure*}[!htbp]
    \centering
	\includegraphics[width=0.96\textwidth]{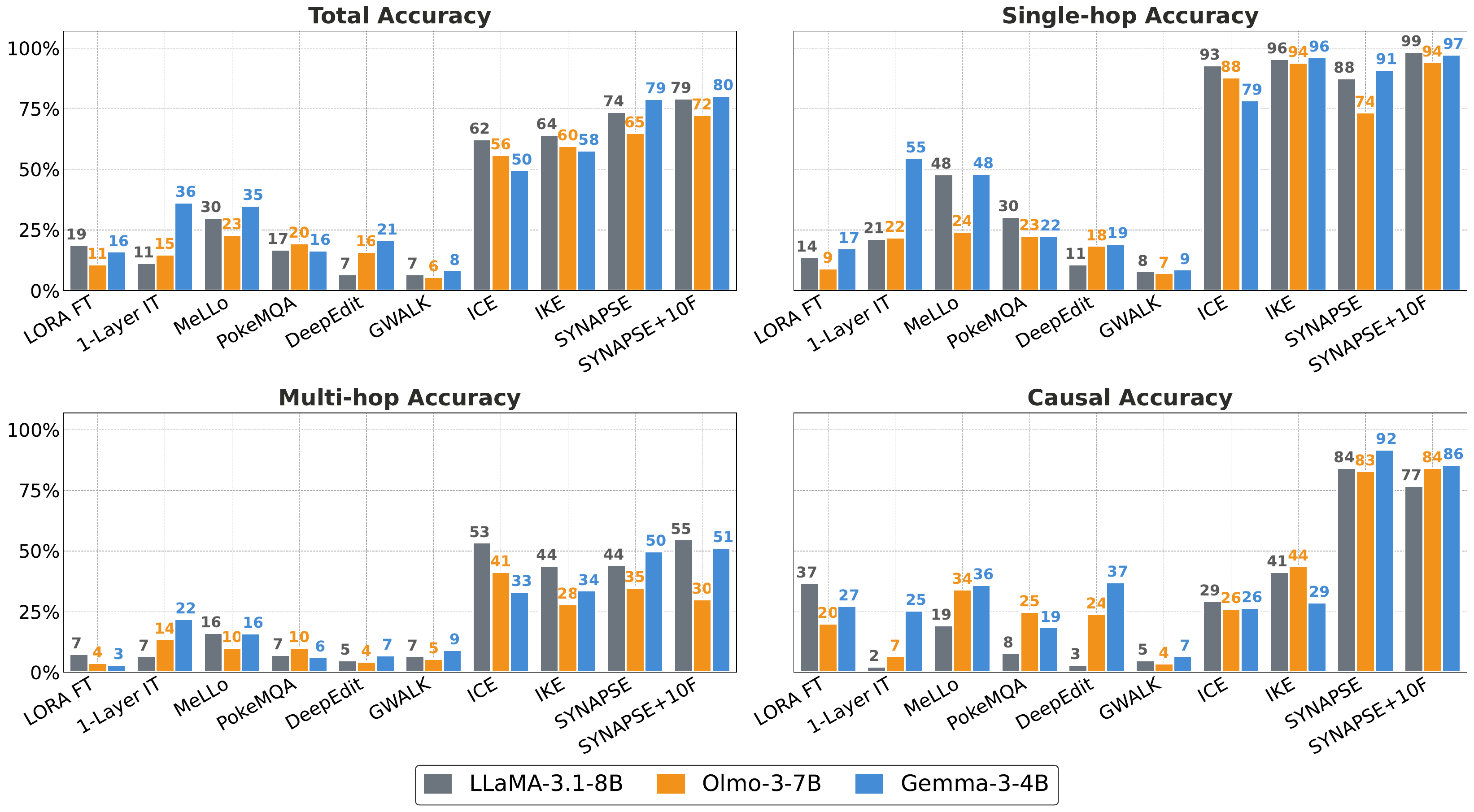}
\vspace{-7pt}
\caption{Question answering performance after inserting 542 facts into LLaMA-3.1-Instruct, OLMo-3-Instruct, and Gemma-3-Instruct. Results are reported for \textbf{total} accuracy, \textbf{single}-hop, \textbf{multi}-hop, and \textbf{causal} accuracy (reasoning over event consequences), rounded to the nearest percent. {\scshape Synapse} consistently achieves the best results on the questions requiring more difficult reasoning. Paired with 10 retrieved facts, {\scshape Synapse} 
achieves the highest accuracies across all question types.}
\label{fig:main_results}
    \vspace{-22pt}
\end{figure*}

\begingroup
\setlength{\abovedisplayskip}{5pt} 
\setlength{\belowdisplayskip}{3pt}
\[
\mathcal{Y}_{\text{neg}}(t) =
\begin{cases}
\{ o_{\text{old}}, \alpha, \eta \} & \text{for pre-existing entities } s \in V_{\text{upd}}, \\[2mm]
\{ \alpha, \eta \} & \text{for newly created entities } s \in V_{\text{new}},
\end{cases}
\]
\endgroup

where $o_{\text{old}}$ are previously stored objects in fact triples $(s, r, o_{\text{old}})$ for pre-existing entities $s \in V_{\text{upd}}$.

\textbf{To prevent hallucinations}, we construct unknown triples $t_{\text{unk}} = (s, r, o) \in \mathcal{T}_{\text{unk}}$ and a preference dataset $\mathcal{D}_{\pi^*}^{\text{unk}}$ incentivizing abstention on questions about involving either (i) unknown entities $V_{\text{unk}} := V \setminus ( V_{\text{new}} \cup V_{\text{upd}} \cup V_{\text{old}} )$, where the object $o$ is an unknown entity in $V_{\text{unk}}$ or an attribute value in $\mathcal{A}$, or (ii) hyperspecific relations $\mathcal{R}_{\text{hyp}}$ of known entities, where $o \in \mathcal{A}$ is an attribute value:
\begingroup
\setlength{\abovedisplayskip}{10pt}
\setlength{\belowdisplayskip}{8pt}
\begin{align*}
\mathcal{T}_{\text{unk}} &= \{ (s, r, o) \mid (s \in V_{\text{unk}}, r \in \mathcal{R}_{\text{edit}}, o \in V_{\text{unk}} \cup \mathcal{A}) \;\lor\; (s \in V_{\text{new}} \cup V_{\text{upd}}, r \in \mathcal{R}_{\text{hyp}}, o \in \mathcal{A}) \}, \\
\mathcal{D}_{\pi^*}^{\text{unk}} &= \bigl\{ (x, \alpha, \eta) \;\big|\; x, \alpha, \eta \sim \pi^*(\cdot \mid t), \; t \in \mathcal{T}_{\text{unk}}, \; \alpha \succ \eta \bigr\}.
\end{align*}
\endgroup

\textbf{To insert new knowledge} into models, we combine the editing and unknown datasets into the final preference dataset $\mathcal{D}_{\pi^*} = \mathcal{D}_{\pi^*}^{\text{edit}} \cup \mathcal{D}_{\pi^*}^{\text{unk}}$. We instantiate the post-training framework with Direct Preference Optimization \citep{rafailov2024directpreferenceoptimizationlanguage}, which optimizes a policy $\pi_\theta$ to maximize the margin of preferred and dispreferred generations on $\mathcal{D}_{\pi^*}$. This setup for {\sc Synapse} models reinforces knowledge acquisition through chain-of-thought reasoning, discourages the generation of incorrect responses or unnecessary abstentions on known information, and enforces abstention on unknown information.
\begingroup
\setlength{\abovedisplayskip}{10pt} 
\setlength{\belowdisplayskip}{8pt}
\[
\begin{aligned}
&\mathcal{L}_{\text{DPO}}(\pi_\theta, \pi_{\text{ref}}; \mathcal{T}) 
= \mathbb{E}_{(x, y_w, y_l) \sim \mathcal{D}_{\pi^*}^{\text{edit}}}\Biggl[
       -\log \sigma \Bigl(
           \beta \log \frac{\pi_\theta(\mathbf{r}_w \oplus a_w \mid x)}
                             {\pi_{\text{ref}}(\mathbf{r}_w \oplus a_w \mid x)}
         - \beta \log \frac{\pi_\theta(y_l \mid x)}
                           {\pi_{\text{ref}}(y_l \mid x)}
       \Bigr)
   \Biggr] \\
&\hspace{7.5em}  + \mathbb{E}_{(x, \alpha, \eta) \sim \mathcal{D}_{\pi^*}^{\text{unk}}} \Biggl[
       -\log \sigma \Bigl(
           \beta \log \frac{\pi_\theta(\alpha \mid x)}
                             {\pi_{\text{ref}}(\alpha \mid x)}
         - \beta \log \frac{\pi_\theta(\eta \mid x)}
                           {\pi_{\text{ref}}(\eta \mid x)}
       \Bigr)
   \Biggr]
\end{aligned}
\]
\endgroup

\textbf{To mitigate excessive divergence} from the original instruction-tuned model and preserve general instruction-following and safety behaviors, we augment with an auxiliary SFT term on response $y_w$:
  \begingroup
\setlength{\abovedisplayskip}{8pt} 
\setlength{\belowdisplayskip}{8pt}
\[
\mathcal{L}_{\text{total}} =
\mathcal{L}_{\text{DPO}} +
\lambda \, \mathbb{E}_{(x,y_w)} \big[ - \log \pi_\theta(y_w \mid x) \big].
\]
\endgroup
\textbf{Verification.}
Sentence-level factual statements are extracted from all synthetic long-form text to construct a unified fact base per event. To ensure consistency in the training data, we apply a unified verification pass over this fact base and the knowledge graph to rewrite all long-form text, ensuring it remains grounded and internally coherent. To mitigate semantic and framing bias, content is generated across balanced perspectives and diverse reasoning structures (e.g., both winning and losing teams, pre- and post-disaster reporting). Preference pairs are simultaneously filtered to remove incorrect, overly specific, internally inconsistent, or unanswerable cases, ensuring both data sources remain mutually consistent without contradictory or degenerate training signals.

% \subsection{GRPO (And Reward Model Design)}
% Here we will look at GRPO, designing a checklist/criterion for the reward model, and seeing how to perform for loewr resource.

\section{Experimental Setup}
\label{sec:experimental_details}

\textbf{Data.} \quad We evaluate experimental setups by inserting 20, 150, 542, and 1,536 facts from our dataset, covering 1, 15, 29, and 41 events, respectively, following prior work on fact insertion for temporal knowledge insertions (e.g., \citep{zhong2024mquakeassessingknowledgeediting}). Question-answering accuracy is reported, where we evaluate on 48 questions (14 causal, 14 multi-hop, 20 single-hop), 353 questions (124 causal, 77 multi-hop, 152 single-hop), 1,314 questions (389 causal, 389 multi-hop, 536 single-hop), and 2,658 questions (1,142 causal, 747 multi-hop, 769 single-hop) for the 20-, 150-, 542-, and 1,536-fact setups, respectively. For benchmark evaluation, binary question accuracy is utilized. Causal questions are marked correct if the model identifies at least one relevant causal event and identifies the correct resulting trend. Both single-hop and multi-hop questions are scored based on the final answer only.

\textbf{Training Details.} \quad We utilize GPT-4.1 \citep{openai2024gpt4technicalreport} to construct synthetic datasets for knowledge updating (see Appendix \ref{app:ablation} for teacher-model ablations) and evaluate performance on LLaMA-3.1-8B Instruct \citep{shah2025reportedcutofflargelanguage}, Olmo-3-7B Instruct \citep{olmo2025olmo3}, and Gemma-3-4B Instruct \citep{gemmateam2025gemma3technicalreport}. For each event, we generate 6--15 articles, totaling 12 (11,489 words), 90 (101,283 words), 210 (235,196 words), and 345 (352,521 words) articles for the 20-, 150-, 542-, and 1,536-fact scenarios, respectively. We create 20 question--preference pairs per single-hop fact and $\sim$60 causal and multi-hop instances per event, yielding 2{,}231, 9{,}454, 28{,}044, and 83{,}385 training instances, with 40\% of pairs enforcing abstention. We further incorporate general preference data from TULU-3 \citep{lambert2025tulu3pushingfrontiers}, downsampled to match each setting (2{,}000, 7{,}000, 28{,}000, and 82{,}792 instances). The final alignment mixture is 30\% $\mathcal{D}_{\text{edit}}$, 20\% $\mathcal{D}_{\text{unk}}$, and 50\% $\mathcal{D}_{\text{gen}}$. We additionally evaluate a setting where {\sc Synapse} and finetuned baselines are provided with 10 retrieved in-context facts. Further details are in Appendix~\ref{app:implementation}.

\textbf{Baselines.} \quad We compare our method against update approaches that directly incorporate $\mathcal{E}$ into the weights of $\pi_\theta$. This includes fine-tuning via LoRA and single-layer instruction tuning \citep{zhang2024comprehensivestudyknowledgeediting} using generated questions and preferred answers from the preference dataset. We also evaluate in-context learning methods that introduce new factual insertions via prompts without modifying weights, including ICE~\citep{cohen2023evaluatingrippleeffectsknowledge} and IKE~\citep{zheng2023editfactualknowledgeincontext}, as well as graph-traversal-based methods: PokeMQA~\citep{gu2024pokemqaprogrammableknowledgeediting}, DeepEdit~\citep{wang2024deepeditknowledgeeditingdecoding}, MeLLo~\citep{zhong2024mquakeassessingknowledgeediting}, and GWalk ~\citep{zhong2025mquakeremastered}, which retrieve facts from knowledge graphs. We further evaluate standard model editing baselines that make small, targeted updates to model weights (e.g., ROME \citep{meng2023locatingeditingfactualassociations}) on LLaMA-3.1-8B (See Appendix \ref{app:additional_results}).

% parameter-based model editing approaches such as ROME~\citep{meng2023locatingeditingfactualassociations} and MEMIT~\citep{meng2023masseditingmemorytransformer}, as well as

% For baselines, we primarily explore commonly utilized methods for knowledge updating in prior work, including in-context learning methods and retrieval-augmented generation (CITE), such as IKE and ICE. We also explore 04.88%step-by-step ICL baselines that calculate for knowledge conflicts, such as PokeMWA, DeepEdit, and MeLLo. Finally, we also look at parameter update methods, primarily with ROME, MEMIT, . We test a baseline model without any knowledge updating to see the rate at which the model will hallucinate on our parallel universe.

\section{Results}
\label{sec:results}
% Traditional model editing approaches, such as ROME and MEMIT, perform particularly poorly, achieving less than 10\% in accuracy. This failure arises from their reliance on modifying \emph{existing} entities in the model’s weights, which does not generalize to our setting involving entirely new entities (e.g., the 2031 Savu Sea Tsunami). While these methods perform well on counterfactual benchmarks with known entities, they are ineffective when the relevant knowledge is absent from the pretrained model.

Figure~\ref{fig:main_results} summarizes results for inserting 542 facts. State-of-the-art counterfactual methods, which rely on graph-based retrieval, perform poorly, achieving only 17.09\% average accuracy and abstaining 72.53\% of the time. While effective on prior benchmarks, they often fail to retrieve relevant facts in our setting. Similarly, LoRA and single-layer finetuning methods only yield 12.02\% accuracy.

 \begin{figure*}[t]
    \centering
	\includegraphics[width=0.98\textwidth]{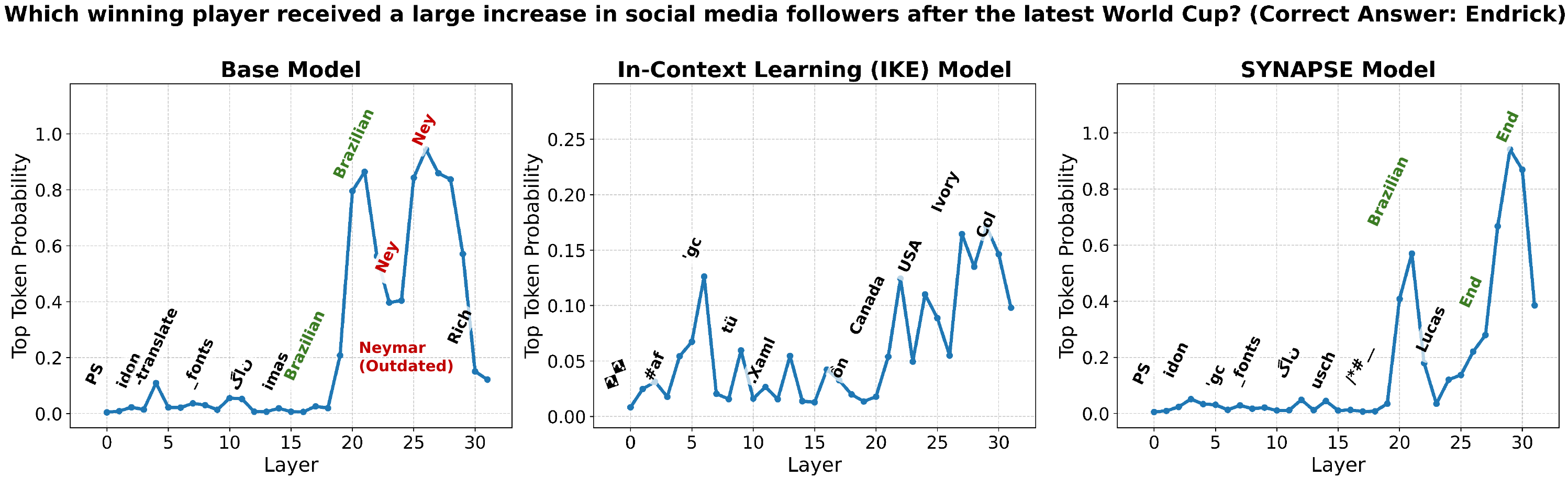}
\vspace{-5pt}
\caption{Token probabilities for a causal question in the \textsc{ParallelEvents} benchmark, comparing the base model, an in-context learning model, and \textsc{Synapse}. Only \textsc{Synapse} consistently produces reasonable answers about event consequences after fact insertions.}
	\label{fig:token_probs_main}
    \vspace{-15pt}
\end{figure*}

Among existing baselines, in-context learning (ICL) methods such as ICE and IKE achieve the strongest performance, with average accuracies of 63.32\% for LLaMA-3.1, 57.73\% for Olmo-3, and 53.73\% for Gemma-3. Given the relatively small number of injected facts, incorporating them into the prompt yields strong results, particularly on single-hop questions (90.86\% accuracy). However, baseline performance drops to 32.65\% on causal questions. Notably, fine-tuning baselines trained on the identical synthetic text and QA pairs do not outperform IKE. Since these baselines use the same synthetic data, their lack of improvement indicates that {\sc Synapse}’s gains stem from the method, not from distributional alignment with the benchmark questions.

In contrast, {\scshape Synapse} achieves the highest overall accuracy on relevant, answerable questions, reaching 73.67\% for LLaMA-3.1, 64.92\% for Olmo-3, and 79.07\% for Gemma-3. Under a paired bootstrap test \citep{berg-kirkpatrick-etal-2012-empirical} with $10^6$ iterations and $\alpha = 0.05$, {\sc Synapse}'s improvement over IKE, the best in-context learning baseline, is statistically significant for all models. An error analysis shows that {\sc Synapse} outperforms ICL methods on questions requiring higher reasoning, especially causal questions, by \textbf{53.73\%} on causal questions, where these baselines abstain 44.13\% of the time. As illustrated in Figure~\ref{fig:token_probs_main}, which shows next-token probabilities, for a question about the consequence of a future World Cup, IKE remains uncertain and inconsistent across layers, reflecting its inability to model event dynamics. In contrast, {\sc Synapse} propagates the correct answer early (around layer 20, e.g., identifying Endrick as Brazilian) and assigns it high probability.

\begin{wrapfigure}{r}{0.46\textwidth}
\vspace{-10pt}
    \centering
    \includegraphics[width=0.48\textwidth]{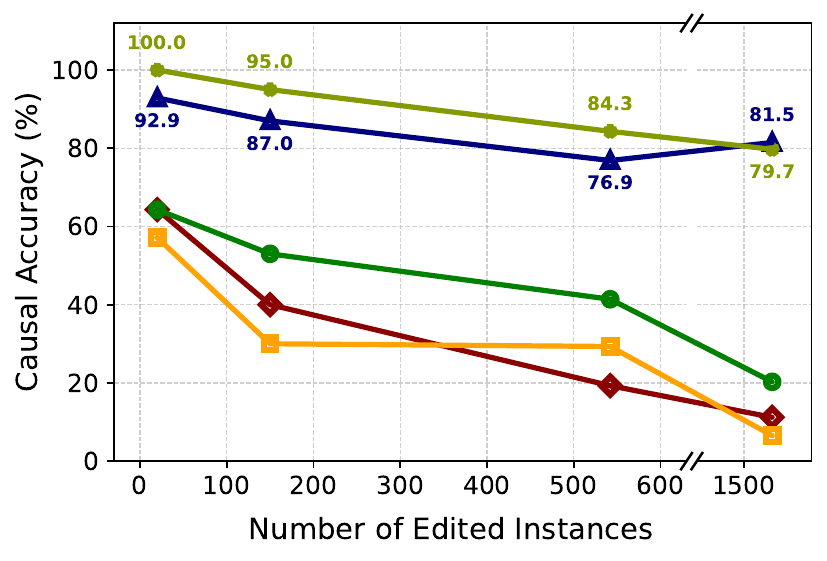}
    \\
    \includegraphics[width=0.48\textwidth]{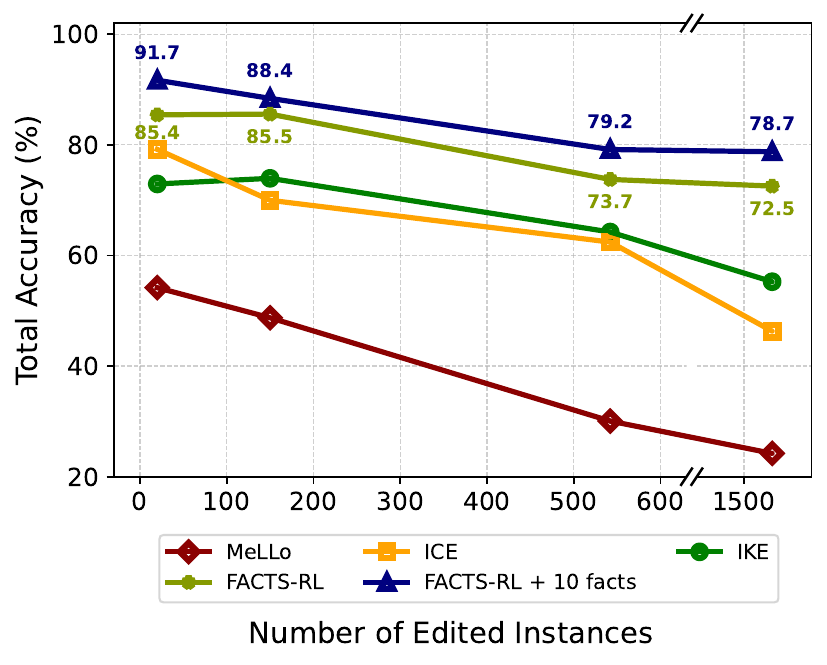}
    \vspace{-15pt}
    \caption{Method performance across fact-editing settings with 150, 542, and 1,536 facts inserted simultaneously. Top: Causal Accuracy. Bottom: Total Accuracy.}
    \label{fig:scaling}
    \vspace{-42pt}
\end{wrapfigure}

However, despite these gains in complex reasoning, {\sc Synapse}'s single-hop accuracy is lower than ICL models. Our analysis attributes this to errors in exact numerical recall, as 64.47\% of cases where IKE outperforms {\sc Synapse} involve numerical attributes, such as the exact wind speed of a typhoon.  To address this, we augment {\sc Synapse} with 10 retrieved facts using the retriever from IKE. Providing the relevant retrieved context eliminates these numerical failures and enables {\sc Synapse} to surpass ICL methods on single-hop questions. We compare {\sc Synapse}+retrieval against matching hybrid setups built on our fine-tuned baselines (Figure \ref{fig:hybrid_results} in Appendix~\ref{app:additional_results}) and find it remains the strongest method overall even in this hybrid setting. {\sc Synapse}'s gains on the hybrid baselines occur mainly on multi-hop and causal questions, indicating that our training method improves model reasoning capabilities on new event information. Compared with all baselines, this hybrid {\sc Synapse} model with 10 retrieved facts achieves top-2 performance on all question types in {\sc ParallelEvents}, showing that retrieval closes the single-hop gap without sacrificing its advantage on complex reasoning.

\section{Ablation Studies}
\label{sec:ablation}

\textbf{Edit Scale.} \quad Figure~\ref{fig:scaling} reports causal and total accuracy for the three best baselines and {\sc Synapse} for LLaMA-3.1-8B as the number of edited facts varies in $\{20, 150, 542, 1,536\}$. Across all models, question-answering accuracy degrades as the number of inserted facts increases. However, on causal accuracy, baseline models deteriorate twice as much as {\sc Synapse}, with an aggregate drop exceeding 40\%. {\sc Synapse} achieves the best performance across all settings, with statistically significant gains in all but the 20-fact case under a paired bootstrap test ($\alpha = 0.05$). Notably, its degradation plateaus from 542 to 1,536 facts, with only a 1.2\% drop, whereas baselines degrade by 10.08\% on average. Overall, {\sc Synapse} maintains strong performance under increasing edit scale, consistently incorporating new knowledge as more events are inserted.

\subsection{Model Generalization}
% \begin{figure}[t]
%     \centering
%     \includegraphics[width=\columnwidth]{images/mmlu_alpaca.pdf}\vspace{-8pt}
% \caption{Results of LLaMA-3.1-8B-Instruct knowledge editing methods evaluated on MMLU-Pro and AlpacaEval. Incorporating TULU-3 as a generalization dataset reduces {\scshape Synapse} model degradation compared to other finetuning methods.}
%     \label{fig:mmlu_alpaca}
%     \vspace{-20pt}
% \end{figure}

\begin{table}[t]
\centering
\setlength{\tabcolsep}{3pt}
\resizebox{0.99\textwidth}{!}{%
\begin{tabular}{lccc ccc ccc}
\toprule

\multirow{2}{*}{\textbf{Dataset}} &
\multicolumn{3}{c}{\textbf{LLaMA-3.1 8B Instruct}} &
\multicolumn{3}{c}{\textbf{Olmo-3 7B Instruct}} &
\multicolumn{3}{c}{\textbf{Gemma-3 4B Instruct}} \\
\cmidrule(lr){2-4} \cmidrule(lr){5-7} \cmidrule(lr){8-10}

& Base & Events Only & {\sc Synapse}
& Base & Events Only & {\sc Synapse}
& Base & Events Only & {\sc Synapse} \\

\midrule

\multicolumn{1}{l}{\textit{Knowledge \& QA}} \\
\quad PopQA & 
\cellcolor[HTML]{EBEBEB}\textbf{32.97\%} & 26.76\% & \cellcolor[HTML]{A9DFBF}31.14\% 
& \cellcolor[HTML]{EBEBEB}\textbf{19.70\%} & \cellcolor[HTML]{A9DFBF}19.07\% & \cellcolor[HTML]{A9DFBF}19.22\% 
& \cellcolor[HTML]{EBEBEB}\textbf{19.88\%} & \cellcolor[HTML]{A9DFBF}17.60\% & \cellcolor[HTML]{A9DFBF}18.89\% \\

\quad MMLU-Pro & 
\cellcolor[HTML]{EBEBEB}\textbf{44.46\%} & 40.35\% & \cellcolor[HTML]{A9DFBF}43.76\% 
& \cellcolor[HTML]{EBEBEB}\textbf{40.94\%} & 35.54\% & \cellcolor[HTML]{A9DFBF}37.36\% 
& \cellcolor[HTML]{EBEBEB}\textbf{38.89\%} & 34.10\% & \cellcolor[HTML]{A9DFBF}36.49\% \\

\quad {\sc ParallelAbstain} & \cellcolor[HTML]{EBEBEB} 81.01\% & 68.99\% & \cellcolor[HTML]{A9DFBF}\textbf{83.33\%} & \cellcolor[HTML]{EBEBEB}\textbf{77.13\%} & 62.79\% & 65.89\% & \cellcolor[HTML]{EBEBEB} 66.67\% & \cellcolor[HTML]{A9DFBF}72.48\% & \cellcolor[HTML]{A9DFBF}\textbf{74.81\%} \\

\midrule

\multicolumn{1}{l}{\textit{Instruction Following}} \\
\quad IFEval & 
\cellcolor[HTML]{EBEBEB}74.15\% & \cellcolor[HTML]{A9DFBF}77.22\% & \cellcolor[HTML]{A9DFBF}\textbf{79.11\%}
& \cellcolor[HTML]{EBEBEB}\textbf{82.62\% }& 76.85\% &  \cellcolor[HTML]{A9DFBF}81.70\%  
& \cellcolor[HTML]{EBEBEB}\underline{78.56\%} & \cellcolor[HTML]{A9DFBF}77.22\% & \cellcolor[HTML]{A9DFBF}\underline{78.56\%}  \\

\quad IFBench & 
\cellcolor[HTML]{EBEBEB}22.79\% & 13.95\% & \cellcolor[HTML]{A9DFBF}\textbf{24.83\%} 
& \cellcolor[HTML]{EBEBEB}\textbf{30.27\%} & 24.49\% & \cellcolor[HTML]{A9DFBF}27.89\% 
& \cellcolor[HTML]{EBEBEB}\textbf{22.11\%} & 11.56\% & \cellcolor[HTML]{A9DFBF}20.41\% \\

\bottomrule
\end{tabular}%
}
\vspace{3pt}
\caption{Comparison of Base, training on event data only, and {\sc Synapse} with TULU-3 across general knowledge and instruction-following datasets. \colorbox[HTML]{A9DFBF}{Green} denotes performance <4\% or better than the base model. {\sc ParallelAbstain} consists of unanswerable questions requiring abstention. {\sc Synapse} with TULU-3 recovers degradation from event-only training, improving by 4.34\% on average.  }
\vspace{-15pt}
\label{tab:generalization}
\end{table}

{\sc Synapse} shows the best performance on new inserted knowledge. To measure if {\sc Synapse} results in catastrophic forgetting of existing knowledge or tasks, we assess general language modeling capabilities of {\sc Synapse} models. We assess general LLM knowledge with MMLU-Pro \citep{wang2024mmluprorobustchallengingmultitask} and PopQA \citep{mallen2023trustlanguagemodelsinvestigating} and instruction following with IFEval \citep{zhou2023instructionfollowingevaluationlargelanguage} and IFBench \citep{pyatkin2025generalizingverifiableinstructionfollowing}. To assess unintended knowledge propagation and safety behaviors of {\sc Synapse}, we construct {\sc ParallelAbstain}, a dataset of 258 questions on (1) temporally adjacent but unknown events occurring both in the timeframe of {\sc ParallelEvents} during 2030--2035 and after 2035, and (2) hyperspecific information (e.g., exact temperature during a hurricane) completely unseen from any training data on the inserted events in {\sc ParallelEvents}.

Table~\ref{tab:generalization} reports results after inserting 542 facts, comparing the base model, finetuning on {\sc Synapse} data only, and finetuning on a mixture of {\sc Synapse} and the general preference set of TULU-3. Finetuning on {\sc Synapse} alone results in notable performance degradations for both existing knowledge and instruction following. In contrast, adding the general preference data TULU-3 largely maintains the abstention and general knowledge capabilities when compared to the base model, yielding an average gain of 4.34\% over training on {\sc Synapse} alone across all models and benchmarks. Notably, training {\sc Synapse} with TULU-3 also results in higher instruction following capabilities than the base LLaMA-3.1-8B-Instruct model. We inspect the model degradations on MMLU-Pro domains and find that the largest drops ($>$5\%) occur in STEM categories (Math, Computer Science, Chemistry), which are underrepresented in {\sc Synapse} and involve technical content such as code.

% \ar{The section title below can be re-written to make it clearer this section is introducing a new method.}

% \begin{table}[t]
% \centering
% \resizebox{0.5\textwidth}{!}{
% \begin{tabular}{l|cccc}
% \toprule
% \textbf{Method} & \textbf{QA} & \textbf{MMLU-Pro} & \textbf{AlpacaEval} & \textbf{Hallucination} \\
% \midrule
% 20\%           & -- & -- & -- & -- \\
% 25\%           & -- & -- & -- & -- \\
% 33\%         & -- & -- & -- & -- \\
% 40\%           & -- & -- & -- & -- \\
% 45\%           & -- & -- & -- & -- \\
% 50\%           & -- & -- & -- & -- \\
% \bottomrule
% \end{tabular}
% }
% \caption{MAKE INTO FIGURE}
% \end{table}

\begin{table*}[htbp]
\centering
\resizebox{0.85\textwidth}{!}{%
\setlength{\tabcolsep}{6pt}
\begin{tabular}{l c cc ccc}
\toprule
Method &
\textbf{Base Model} &
\multicolumn{2}{c}{\textbf{Training Data}} &
\multicolumn{3}{c}{\textbf{Accuracy}} \\
\cmidrule(lr){3-4} \cmidrule(lr){5-7}
& & \textbf{Edited} & \textbf{Synthetic} &
\textbf{Train Edited} & \textbf{Test Edited} & \textbf{Unedited} \\
\midrule
GWalk   & Mistral-7B-Instruct-v0.2 &  &  & 57.14\%  & 33.33\% & 56.28\% \\
\cellcolor[HTML]{EBEBEB}GWalk & \cellcolor[HTML]{EBEBEB}LLaMA-3.1-Instruct &\cellcolor[HTML]{EBEBEB}  & \cellcolor[HTML]{EBEBEB} & \cellcolor[HTML]{EBEBEB}72.00\% & \cellcolor[HTML]{EBEBEB}33.33\% & \cellcolor[HTML]{EBEBEB}\textbf{66.66\%} \\
\midrule
{\sc Synapse}-1 & LLaMA-3.1-Instruct &  \xmark & \xmark & 1.22\% & 2.00\% & 54.13\% \\
\cellcolor[HTML]{EBEBEB}{\sc Synapse}-2 &\cellcolor[HTML]{EBEBEB} LLaMA-3.1-Instruct &  \cellcolor[HTML]{EBEBEB}\cmark & \cellcolor[HTML]{EBEBEB} \xmark & \cellcolor[HTML]{EBEBEB}73.09\% &\cellcolor[HTML]{EBEBEB}76.00\% & \cellcolor[HTML]{EBEBEB}10.04\% \\
{\sc Synapse}-3 & LLaMA-3.1-Instruct & \xmark & \cmark & 0.92\% & 3.00\% & 58.72\% \\
\cellcolor[HTML]{EBEBEB}{\sc Synapse}-4 & \cellcolor[HTML]{EBEBEB}LLaMA-3.1-Instruct & \cellcolor[HTML]{EBEBEB}\cmark & \cellcolor[HTML]{EBEBEB}\cmark &\cellcolor[HTML]{EBEBEB}
\textbf{85.02\%} & \cellcolor[HTML]{EBEBEB}\textbf{77.00\%} & \cellcolor[HTML]{EBEBEB}52.07\% \\
\bottomrule
\end{tabular}
}
\vspace{-3pt}
\caption{Performance of {\scshape Synapse} on MQuAKE-Remastered, a counterfactual knowledge editing benchmark. We compare against GWalk, a graph-traversal ICL baseline, and evaluate {\scshape Synapse} under different subsets of training data. {\scshape Synapse} \textbf{achieves the highest editing accuracy.}}
\vspace{-10pt}
\label{tab:mquake}
\end{table*}

\subsection{Knowledge Insertion on {\scshape MQuAKE-Remastered}}

\textbf{Experimental Setup.}\,\, To evaluate the portability of our method, we also apply {\sc Synapse} to counterfactual editing on LLaMA-3.1-8B-Instruct using {\sc MQuAKE-Remastered} \citep{zhong2025mquakeremastered}. We focus on the Counterfactual-6334 subset and explore a small-scale setting where we insert 100 counterfactual edits of existing entities. The {\sc MQuAKE-Remastered} evaluation set comprises three subsets: \textit{Train Edited}, containing questions that directly probe the edited knowledge the model is explicitly trained on; \textit{Test Edited}, containing unseen questions whose answers change as a consequence of the edits, used to assess whether the model generalizes the edited knowledge to related facts;; and \textit{Unedited}, containing questions about the same or similar entities whose answers are unaffected by the edits, used to assess locality. The preference dataset for {\sc Synapse} is built from the \textit{Train Edited} split, where preferred responses reflect updated counterfactuals and rejected responses the original facts. We also utilize {\sc Synapse} to generate a \textit{Synthetic} Unedited set, constructed to mirror the \textit{Unedited} distribution while remaining non-identical to prevent data leakage, where preferred responses are factual and rejected responses are fictional counterfactuals.

\textbf{Results.} \,\, We report accuracy on all question splits. Table~\ref{tab:mquake} compares our method against GWalk, a state-of-the-art graph-traversing ICL baseline introduced. {\sc Synapse} achieves strong editing performance in counterfactual settings, reaching 85.02\% train and 75.02\% test accuracy, outperforming GWalk on test edits by over 41.67\% and indicating strong edit generalization across shared entities. However, this comes with a trade-off, as \textit{Unedited} accuracy decreases to 52.07\%. The base LLaMA-3.1-Instruct and GWalk also perform poorly on unedited questions (59.01\% on average), highlighting the difficulty of maintaining generalization in counterfactual settings. Finally, we train {\sc Synapse} on different subsets of the training data: \textit{Train Edited} only, \textit{Synthetic Unedited} only, or both. Training on edited data alone yields strong edit generalization but degrades performance on unedited knowledge, whereas training on unedited data alone preserves existing knowledge but performs poorly on the edits. Combining both sets balances edit accuracy with locality and even improves trian and test accuracy by 11.93\% and 1.00\% over training on over \textit{Train Edited} alone, suggesting unedited data also facilitates learning the edits themselves.

\section{Related Work}
\textbf{Knowledge Updating.} Prior work shows that language models degrade under temporal drift, affecting neologisms, entities, recent facts, counterfactual knowledge, and isolated events \citep{zheng2024neobenchevaluatingrobustnesslarge, rijhwani-preotiuc-pietro-2020-temporally, vu2023freshllms, zhong2024mquakeassessingknowledgeediting, park2025textitnewnewssystem2finetuning}. Existing approaches address this via localized knowledge editing \citep{mitchell2022fastmodeleditingscale, meng2023masseditingmemorytransformer}, in-context updates \citep{zheng2023editfactualknowledgeincontext}, retrieval- augmentation \citep{gu2024pokemqaprogrammableknowledgeediting}, and fine-tuning \citep{rozner2024knowledgeeditinglanguagemodels, xiong2025finetuningeffectivesolutionreassessing}. Separately, context distillation \citep{park2025textitnewnewssystem2finetuning, snell2022learningdistillingcontext} and episodic semantic memory \citep{holur2024creatingaiobservergenerative, rajesh2026factretrievalepisodicmemory, rajesh2026paninicontinuallearningtoken} have been proposed as alternative mechanisms for internalizing or storing new knowledge, which we compare against in Appendix~\ref{app:additional_results}. However, all these methods largely target isolated fact corrections.

% \textbf{Synthetic Text Generation.} Synthetic data has been extensively explored for pre-training,  domain adaptation, and post-training objectives such as instruction following, alignment, and reasoning \citep{eldan2023tinystoriessmalllanguagemodels, gururangan2020dontstoppretrainingadapt, lambert2025tulu3pushingfrontiers, muennighoff2025s1simpletesttimescaling, nguyen2025synthetictextgenerationtraining, abdin2024phi4technicalreport}, with analyses focusing on scaling, quality, and novelty \citep{kang2025demystifyingsyntheticdatallm, liu2024bestpracticeslessonslearned}. Synthetic data has also been used for time-series modeling and novel event detection \citep{rousseau2025forgingtimeserieslanguage, huidong-etal-2024-generate}. To our knowledge, no prior work explores synthetic generation for updating LLMs on future events. A more comprehensive discussion of related work is provided in Appendix~\ref{app:related_work}. 
\paragraph{Synthetic Text Generation.} Synthetic data has been extensively explored for pre-training, domain adaptation, and post-training objectives such as instruction following and reasoning \citep{lambert2025tulu3pushingfrontiers, muennighoff2025s1simpletesttimescaling, nguyen2025synthetictextgenerationtraining, abdin2024phi4technicalreport}. Closer to our setting, synthetic generation has been used to construct benchmarks from one-off, fictional but realistic news articles \citep{park2025textitnewnewssystem2finetuning}, hierarchical knowledge graphs of academic subjects (e.g., biology, history) for evaluating how edits and unlearning propagate to related knowledge \citep{luo2026knowledgesmithuncoveringknowledgeupdating}, and realistic episodic memories of local experiences \citep{huet2025episodicmemoriesgenerationevaluation}. Beyond evaluation, prior work also fine-tunes models on synthetic data derived from in-context summaries of these fictional articles \citep{park2025textitnewnewssystem2finetuning}.  However, to our knowledge, no prior work synthetically generates \textit{complex, grounded, real-world event data} for updating LLMs on future events. Further details of related work can be found in Appendix~\ref{app:related_work}.

\section{Conclusion}
We introduce \textsc{ParallelEvents}, a benchmark for evaluating temporal knowledge updating, grounded in plausible future events and designed for reasoning over newly emerging information. Building on this benchmark, we propose \textsc{Synapse}, a scalable continual training framework that leverages synthetic data generation to construct datasets for both pre-training and post-training. Empirically, \textsc{Synapse} outperforms strong baselines by over 14.23\%, demonstrating its effectiveness in internalizing new knowledge while mitigating hallucinations. Moreover, \textsc{Synapse} exhibits strong knowledge insertion capabilities, consistently surpassing baselines in editing performance across all settings, including large-scale scenarios of inserting 1{,}536 facts, as well as evaluations on \textsc{MQuAKE-Remastered}. We hope that releasing our benchmark and method highlights promising directions for future research in continual knowledge integration and adaptive learning systems.

\subsubsection*{Acknowledgments}
We would like to thank Oleksandr Lavreniuk for his help in annotating and refining the {\sc ParallelEvents} dataset. We would also like to thank Jiawei Zhou and Greg Durrett for the valuable discussions. 

% This work is in part supported by the NSF CAREER Awards IIS-2144493 and IIS-2052498. Any opinions, findings, and conclusions or recommendations expressed in this material are those of the authors and do not necessarily reflect the views of the National Science Foundation.

\bibliographystyle{plain}
\bibliography{custom}
%%%%%%%%%%%%%%%%%%%%%%%%%%%%%%%%%%%%%%%%%%%%%%%%%%%%%%%%%%%%

\appendix
\section{Impact Statement}
\label{sec:impact_statement}

This paper explores novel methodologies for knowledge updating in large language models (LLMs), with the goal of addressing temporal knowledge drift and improving model safety and reliability. Such improvements are particularly important for users who rely on LLMs in real-time, high-stakes scenarios, such as search engines, live event tracking, or emergency response systems. By enabling models to incorporate new facts accurately, we aim to reduce the risk of outdated or incorrect information being presented to users.

Our dataset consists entirely of fictional events designed to be realistic, including topics with serious implications, such as natural disasters (with casualty counts) and economic crises in various regions. To ensure plausibility, we leverage real events as prompts to guide the model in generating fictional events with credible outcomes, locations, and details. While this process may still introduce potential biases, each event is carefully reviewed by in-house annotators and one of the authors to minimize harmful content—for example, by removing culturally insensitive or biased names of individuals.

Generating realistic synthetic knowledge also can carry potential misuse risks, such as the creation of convincing misinformation or fabricated ``facts'' that could be used to manipulate downstream models. Our event topics (e.g., natural disasters, sporting events, movies) are large-scale and public in nature rather than targeting specific real entities, which limits the risk of misuse for targeted misinformation or harassment. To further mitigate misuse, we explicitly flag all generated events as synthetic/fictional in the released dataset via metadata tags and clear naming conventions, restrict release to research-oriented licenses, and recommend that any derivative use continue to clearly label content as synthetic to prevent conflation with real-world facts.

A key finding of our work is that, although our methodology reduces temporal knowledge drift, it slightly decreases the model's general LLM capabilities. Nevertheless, model performance remains comparable, and we observe that the largest degradations occur in out-of-domain topics like math and code. These observations highlight the trade-offs between knowledge updating, reasoning capability, and model reliability. Future research could look at knowledge updating while maintaining technical knowledge, like code capabilities.

We emphasize that the deployment of updated LLMs should be done responsibly. Potential risks include the propagation of errors in real-time decision-making or the inadvertent reinforcement of subtle biases in synthetic data. Future research should explore the societal impact of such models, including user studies in urgent scenarios, like live sporting events or impending natural disasters. Moreover, ongoing work should investigate techniques to further ensure fairness, transparency, and accountability when updating model knowledge, particularly when models interact with diverse populations or sensitive information.

Overall, our work aims to enhance the fidelity and utility of LLMs in dynamic real-world contexts, enabling models to provide more accurate, timely, and actionable information. Beyond improving reliability, knowledge-updated LLMs have the potential to empower decision-making in critical domains, support evidence-based policy, and increase public trust in AI systems when deployed responsibly. By carefully integrating new information, these models can contribute to more equitable access to knowledge, help mitigate human error in high-stakes scenarios, and assist underrepresented communities with tailored insights. Furthermore, applications in healthcare, disaster response, education, and governance could directly improve human well-being and societal outcomes. These opportunities highlight the importance of continued research into transparent updating procedures, robust auditing mechanisms, and participatory evaluation methods that engage diverse stakeholders, ensuring that knowledge-updated LLMs provide socially beneficial, ethical, and trustworthy support across a wide range of domains.

\section{Limitations}
\label{app:limitations}

\textbf{Dataset scope and design.} \quad While {\sc ParallelEvents} covers only 6 primary event topics, each event is highly complex; for instance, a hypothetical FIFA World Cup event contains over 40,000 true editable facts, including non-consequential details such as players' spouses. Limited automation is a deliberate design choice to ensure events remain fictional, realistic, and internally consistent, with construction requiring approximately 350 hours overall. To address potential bias, we employ manual verification and diversity augmentation, evenly sampling from automatically produced parameters such as geographic region and severity to prevent overrepresentation of particular entities. Future work should look into extensions of this parallel universe for more events, such as company mergers or the creation of a new commercial product.

{\sc ParallelEvents} spans the years 2030--2035 but is designed with refreshability in mind: for each event, we can systematically advance the associated years or dates through simple functional graph parsing, allowing us to push the dataset's temporal end date forward as needed. For example, a fictional event like ``Hurricane Alicia occurred on August 4th, 2030'' can be updated to ``September 3rd, 2043,'' allowing flexible experimentation across temporal contexts and supporting temporal insertions for new LLMs. Additionally, during dataset construction we compiled a structured list of valid parameters for each event type (e.g., locations, winners, attendance figures, economic costs), which allows us to shuffle and generate new permutations of events that still satisfy real-world constraints, producing fresh event instances not present in any prior release. Together, these mechanisms allow {\sc ParallelEvents} to be periodically refreshed, mitigating contamination risk as newer pretraining corpora are released.

\textbf{Training methodology.} \quad We only explore using DPO in the offline setting, due to its unexplored application for factual insertions of temporal events; this offline setup also avoids the challenge of training a reward model for hypothetical future events, though dynamic updates can be handled by periodically refreshing the training set. Future work should investigate online settings, such as using GRPO for updating codebases~\citep{wu2025recodeupdatingcodeapi}, to enable more dynamic, real-time knowledge integration.

{\sc Synapse} requires the generation of high-quality synthetic data for updating a model's knowledge. While generating high-quality synthetic SFT and preference data incurs some overhead, this cost is substantially lower than continual pretraining, which requires trillions of training tokens. Table \ref{tab:training_stats} provides the statistics for the amount of data needed to be generated for each training split, the generation time, and the training time. For moderate factual insertions, this can be done in a day. We also examine the effect of varying the teacher model used to train {\sc Synapse} (Table \ref{tab:teacher-150} and Table \ref{tab:teacher-542}), and find that teacher strength does affect {\sc Synapse}'s performance. However, even with weaker teacher models, {\sc Synapse} outperforms the strongest baselines in our benchmark.

\textbf{Evaluation scope.} \quad We provide analyses of knowledge editing on this dataset, evaluating settings with 25, 150, 542, and 1,536 inserted facts, as well as a 100-fact setting on {\sc MQuAKE}. These choices are partly motivated by the computational and data-generation costs of larger-scale interventions, which require substantially longer training times. Importantly, we find that, on average, only 20--50 \emph{consequential} facts per event require insertion (e.g., those that affect key outcomes or information like the winner of the World Cup), suggesting that our methodology is well suited for real-time event editing. In this regime, our approach consistently outperforms in-context learning baselines. Due to the computational cost of training models with different subsets of training data, including abstention data and general preference mixtures, hyperparameters, and ablations, we only explore ablations on the 150-fact setting. Additionally, we report results only for LLaMA-3.1-8B, given its superior performance and editability compared to Olmo-3.

\textbf{Sequential updates and catastrophic forgetting.} \quad Our current evaluation of {\sc Synapse} focuses on a batch insertion setting, where the model is updated once with a set of 10 to 1,536 synthetic facts and then evaluated on general benchmarks (e.g., MMLU-Pro, PopQA) alongside the inserted knowledge. This design is consistent with prior knowledge-insertion work, but it leaves open the question of how {\sc Synapse} behaves under multi-round sequential updates, where new facts are inserted incrementally over time. In particular, we have not measured whether earlier-inserted facts degrade as additional rounds of updates are applied. Evaluating {\sc Synapse} under sequential, multi-round updates, including tracking retention of both earlier synthetic insertions and general benchmark performance across rounds, is an important direction for future work.

\textbf{Scale and real-world applicability.} \quad Our evaluation of {\sc Synapse} is conducted in fully synthetic settings, which may not fully capture the challenges of larger-scale or real-world continual learning. Factual performance decreases as the number of new facts grows (10 to 1,536 facts), consistent with other knowledge-insertion work, though our instruction-tuning preference set helps retain instruction-following ability. We expect this degradation to compound further when knowledge insertions are combined with task-specific fine-tuning or scaled to hundreds of events. Real-world continual learning also introduces challenges beyond our synthetic setting, such as noisy or contradictory updates, which likely require a contradiction-resolution step before integration. We view scaling {\sc Synapse} to noisier knowledge domains as important future work.

% Finally, we do not report \textsc{GWalk}~\citep{zhong2025mquakeremastered} on our baseline. Additionally, we do not include NMKE~\citep{liu2025editlessachievemore}, a new sparse neuron model editing methodology for factual updating, as the codebases for both methods have not yet been released. We will add these as important baselines once the codebases become available.

\section{Extended Related Work}
\label{app:related_work}
\paragraph{Knowledge Updating.} 
Prior work has identified model degradation due to temporal drift across multiple dimensions, including neologisms \citep{zheng2024neobenchevaluatingrobustnesslarge}, named entities \citep{rijhwani-preotiuc-pietro-2020-temporally, liu2024ecbdevidencecenteredbenchmarkdesign, pmlr-v267-thede25a}, recent facts \citep{vu2023freshllms, kasai2024realtimeqawhatsanswer, zhong2024mquakeassessingknowledgeediting}, and counterfactual knowledge \citep{meng2023locatingeditingfactualassociations, li2023evaluatingdependenciesfactediting, zhang2024comprehensivestudyknowledgeediting, zhong2024mquakeassessingknowledgeediting, zhong2025mquakeremastered}. Closely related work also constructs hypothetical-but-plausible future events, but evaluates whether a model internalizes a given fact and its immediate implication in isolation \citep{park2025textitnewnewssystem2finetuning}; in contrast, our events are embedded within a dense Wikidata-style knowledge graph, enabling us to evaluate whether a fact's consequences are correctly propagated across a broader web of existing knowledge.

A substantial body of work has explored knowledge editing approaches, including localized model edits via auxiliary hypernetworks \citep{mitchell2022fastmodeleditingscale} or direct modification of transformer weights associated with specific facts \citep{meng2023locatingeditingfactualassociations, meng2023masseditingmemorytransformer, liu2025editlessachievemore, wang2024wiserethinkingknowledgememory, fang2025alphaeditnullspaceconstrainedknowledge}. In-context learning methods for injecting updated knowledge at inference time \citep{zheng2023editfactualknowledgeincontext, qi2025incontexteditinglearningknowledge} and retrieval- or graph-based mechanisms that provide relevant facts from structured knowledge sources \citep{zhong2024mquakeassessingknowledgeediting, gu2024pokemqaprogrammableknowledgeediting, wang2024deepeditknowledgeeditingdecoding, zhong2025mquakeremastered} have also been explored. More recently, reinforcement learning has been investigated for integrating knowledge updates across code APIs \citep{wu2025recodeupdatingcodeapi}, general factual knowledge, and counterfactuals \citep{rozner2024knowledgeeditinglanguagemodels}, alongside complementary strategies such as single-layer fine-tuning \citep{xiong2025finetuningeffectivesolutionreassessing} and masked text fine-tuning \citep{zhang2024comprehensivestudyknowledgeediting}. There has also been prior work in studying knowledge unlearning \citep{zhang2024negativepreferenceoptimizationcatastrophic, maini2024tofutaskfictitiousunlearning, fan2025simplicityprevailsrethinkingnegative, shi2024musemachineunlearningsixway, li2024wmdpbenchmarkmeasuringreducing} acquisition and memorization in synthetic texts \citep{kirchenbauer2026fictionalqadatasetstudyingmemorization, huang-etal-2024-demystifying, biderman2023emergentpredictablememorizationlarge}. In research closely related to ours, data generation has been utilized to finetune models for continual learning on general benchmarks \citep{luo2026knowledgesmithuncoveringknowledgeupdating} and standalone news articles \citep{park2025textitnewnewssystem2finetuning}. Despite this progress, existing work largely focuses on correcting simple or isolated knowledge updates, like singular news articles, with no systematic study of temporal drift induced by complex, realistic future events.

A separate line of work models temporally-structured experience explicitly as episodic memory, rather than relying on undifferentiated context windows or parametric weight updates, with benchmarks and non-parametric structured-memory frameworks proposed for multi-hop reasoning, link prediction, and event anticipation over such memories \citep{huet2025episodicmemoriesgenerationevaluation, rajesh2026factretrievalepisodicmemory, holur2024creatingaiobservergenerative, rajesh2026paninicontinuallearningtoken}. Context distillation methods, which internalize in-context information into model parameters via self-generated fine-tuning data, have also been studied for knowledge insertion, including protocols for closing the gap between fine-tuning and in-context learning \citep{snell2022learningdistillingcontext, park2025textitnewnewssystem2finetuning}.

Prior work has also explored other facets of knowledge updating, including knowledge forgetting via earlier temporal cutoffs that test a model’s ability to provide time-conditioned information \citep{gao-etal-2025-prompts}, as well as real-time question answering using lifelong snapshots of Wikipedia \citep{vu2023freshllms, kasai2024realtimeqawhatsanswer, pmlr-v267-thede25a}. While these interfaces provide useful datasets for studying LLM learning dynamics through in-context learning or evaluating isolated, one-off updates to factual triples, they fail to capture the complex, interdependent, and evolving nature of event-based knowledge updates that will occur in the future.

\paragraph{Synthetic Text Generation.} There has been substantial work exploring the synthetic generation of pre-training data for small language models \citep{eldan2023tinystoriessmalllanguagemodels}, domain adaptation in general tasks \citep{gururangan2020dontstoppretrainingadapt, yang2024syntheticcontinuedpretraining}, and code \citep{gunasekar2023textbooksneed}. These methods often iteratively generate new data from a small seed set of real examples. Other approaches explore synthetic data generation via rephrasing human-written text \citep{maini2024rephrasingwebrecipecompute}, training models on linguistic rules derived from formal languages \citep{hu2025circuitschomskyprepretrainingformal}, and prompting models to encourage more diverse generations \citep{patel2024datadreamertoolsyntheticdata, chen2024diversitysyntheticdataimpact}. Prior research has also studied the generation of synthetic post-training data for instruction following \citep{lambert2025tulu3pushingfrontiers, wang2023farcamelsgoexploring}, alignment \citep{ge2025scalingsyntheticdatacreation}, and reasoning \citep{muennighoff2025s1simpletesttimescaling}. Analyses of synthetic data generation largely focus on scaling laws \citep{kang2025demystifyingsyntheticdatallm}, data quality \citep{liu2024bestpracticeslessonslearned}, and data novelty \citep{yang2024syntheticcontinuedpretraining}. Some work utilizes synthetic data throughout the entire training pipeline, spanning both pre-training and post-training, particularly for code-based models \citep{abdin2024phi4technicalreport}. Most closely related to our work, prior research has leveraged synthetic data generation for time-series data \citep{rousseau2025forgingtimeserieslanguage} and novel event detection \citep{huidong-etal-2024-generate}. 

Closer still, synthetic generation has recently been used to construct benchmarks and training signal specifically for knowledge updating: \citep{park2025textitnewnewssystem2finetuning} use context distillation over synthetically generated documents to integrate new knowledge into model weights; \citep{luo2026knowledgesmithuncoveringknowledgeupdating} synthesize edits to probe model editing and unlearning; and \citep{huet2025episodicmemoriesgenerationevaluation} construct synthetic episodic memories to evaluate LLMs' temporal and contextual recall. These approaches, however, target synthetically invented or static facts rather than complex, grounded, real-world events. To the best of our knowledge, no prior work has explored the synthetic generation of temporal event data grounded in real-world occurrences, particularly for updating LLMs with knowledge of such events.

\begin{table}[t]
\centering
\resizebox{0.55\textwidth}{!}{
\begin{tabular}{lcccc}
\toprule
IKE & Total Accuracy & Single Hop & Multi-Hop & Causal\\
\midrule

\multicolumn{5}{c}{\textbf{LLaMA-3.1-8B-Instruct}} \\
\midrule
10  & 57.99\% & 91.23\% & 31.62\% & 38.56\% \\
32  & 64.23\% & 95.52\% & 43.96\% & 41.39\% \\
64  & 63.62\% & 93.84\% & 38.05\% & 38.05\%  \\
128 & 62.57\% & 93.09\% & 40.36\% & 42.67\% \\
542 & 62.41\% & 92.91\% & 53.47\% & 29.31\% \\

\midrule
\multicolumn{5}{c}{\textbf{Olmo-3-7B-Instruct}} \\
\midrule
10  & 58.22\% & 92.72\% & 31.62\% & 37.28\% \\
32  & 59.59\% & 94.03\% & 28.02\% & 43.70\% \\
64  & 58.98\% & 93.10\% & 26.99\% & 43.96\% \\
128 & 59.36\% & 93.10\% & 29.82\% & 42.41\% \\
542 & 55.86\% & 87.87\% & 41.39\% & 26.22\% \\

\bottomrule
\end{tabular}
}
\caption{{\sc ParallelEvents} accuracy for IKE when prompted with differing amounts of retrieved facts. For both models, providing 32 facts yields the highest overall accuracy.}
\label{tab:ike_scaling}
\end{table}

\begin{table}[t]
\centering
\resizebox{0.9\textwidth}{!}{
\begin{tabular}{l|c|c|c|c}
\toprule
\textbf{\# of facts} & \textbf{\# of Events} & \textbf{\# of Tokens} & \textbf{Generation time (hours)} & \textbf{Training time (hours, 2 H100s)} \\
\midrule
20 & 1 & 89{,}099 & 0.50 & 0.4 \\
150 & 15 & 1{,}336{,}471 & 6.50 & 2.2 \\
542 & 29 & 3{,}738{,}524 & 18.17 & 8.1 \\
1{,}536 & 41 & 10{,}438{,}174 & 50.74 & 22.5 \\
\bottomrule
\end{tabular}
}
\caption{Generation and Training time statistics for {\sc Synapse} across 20, 150, 542, and 1,536 fact insertion cases. Both the preference dataset and the long form text are included for each training split. Tokens are measured using the LLaMA-3.1 Tokenizer.}
\label{tab:training_stats}
\vspace{-20pt}
\end{table}

\section{Benchmark Details}
\label{app:dataset}

Figure \ref{fig:page_counts} reports Wikipedia page counts for subject--object co-occurrence pairs $(s,o)$ from factual triples in existing benchmarks and {\sc ParallelEvents}, serving as a proxy for how strongly such rigid facts are represented in pre-training data. We find that entity pairs in {\sc MQuAKE-CF} and {\sc CounterFact} \citep{meng2023locatingeditingfactualassociations} frequently co-occur across large numbers of documents, with 38.43\% and 10.16\% of pairs appearing in more than 1{,}000 Wikipedia pages, respectively. For example, the edit $(\text{United Kingdom}, \text{part of}, \text{Europe}) \rightarrow (\text{United Kingdom}, \text{part of}, \text{Oceania})$ involves entities that co-occur in 120{,}340 Wikipedia articles, indicating a strong pre-existing association that makes downstream queries inconsistent or difficult to resolve after editing; similarly, in {\sc CounterFact}, the edit $(\text{Mozambique}, \text{continent}, \text{Africa}) \rightarrow (\text{Mozambique}, \text{continent}, \text{Antarctica})$ corresponds to 34{,}815 co-occurring articles. Empirically, across 42 related questions per edited fact, LLaMA-3.1-8B-Instruct achieves only 17.86\% accuracy after editing with in-context prompting, suggesting strong conflicts between edits and pre-training knowledge. In contrast, {\sc ParallelEvents} contains far fewer highly co-occurring entity pairs. For instance, a synthetic event in which South Whittier becomes an incorporated city in Los Angeles County has the factual triple (South Whittier, incorporated city, Los Angeles County), and South Whittier and Los Angeles County only co-occur in 123 shared articles. Thus, the future worlds of {ParallelEvents} reduce pre-existing associations and thereby mitigate unintended inconsistencies introduced by factual edits.

In total, our dataset covers natural disasters, general elections, sporting events, economic crises, and new city incorporations. The full dataset includes one creative work with a movie; one global economic crisis; four city incorporations in the United States and China; six one-off natural disasters, including a blizzard, a city fire, a hurricane, a sinkhole, and two tornadoes; and ten cause-and-effect disaster pairs. These include six random events—namely an earthquake and conflagration, an earthquake and derailment, an earthquake and oil spill, a thunderstorm and flash floods, a volcano and landslide, and a volcano and tsunami—and four seasonal events, including a blizzard and avalanche, a drought and wildfire, flooding and landslide, and a typhoon and flooding. The dataset also includes three sporting events: the 2030 NBA season, the 2035 NFL season (including the Super Bowl and halftime show performance), and the full 2034 FIFA World Cup, including all players, coaches, and referees. Players in the FIFA World Cup are split evenly between real current players and fictional names. We also cover one UK general election, with both a close result and landslide result variant, where the leaders of all participating political parties are fictional individuals. The dataset only contains entities that participate in at least one factual triple requiring an update.

For consistency checks, 2 in-house annotators (including an author) verified events for realism, entity-level attribute issues, uniqueness, temporal consistency, and stereotypes/bias in names and countries, enforcing unique names, consistent tournament brackets, and valid age statistics for real players. Election vote counts used human-guided generation to maintain realistic ranges for existing parties, as fully synthetic generation of these numerical attributes was inconsistent. We doubly annotate 20\% of the events and calculate the inter-annotator agreement of selecting a JSON field for editing to be Cohen's $\kappa = 0.8298$. 26.67\% of events were discarded and regenerated (incoherence, unrealism, excessive similarity), and a further 14.3\% of events generated during knowledge graph construction were filtered for being too similar to prior events. 35\% of entities in accepted events were manually edited for logical consistency, most commonly for entity uniqueness (e.g., duplicate coaches), temporal ordering, and invalid tournament brackets. In the final dataset, the most frequent relation is ``start time'' (29,324), and the most frequent entity is ``association football'' (2,627).

Potential stereotypes or negative biases associated with names and countries are also reviewed by an in-house annotator. For the election scenarios, we utilize human labor combined with random generation to determine vote counts for each party while maintaining realistic ranges for existing parties, as fully synthetic generation of these numerical attributes was found to be inconsistent. For relations, the most frequent relation is “start time” (29,324), and the most frequent entity is “association football” (2,627). For generated knowledge graphs, 14.3\% of events are filtered for being too similar to prior events. 35\% of entities in the dataset are manually edited for logical consistency.

For evaluation, each fact-insertion setting uses a different subset of events. The 20-fact setting explores a single natural disaster and its effects; the 150-fact setting explores 15 natural disasters; the 542-fact setting explores 29 events, consisting of 26 natural disasters and three sporting events plus the halftime show; and the 1,536-fact setting explores all 41 events. Multi-hop questions in our benchmark average 3.06 hops (2/3/4-hop: 26.9\% / 39.9\% / 26.0\%), while causal questions average 1.66 hops (a step in a chain of actions) across 40+ consequence types. We utilize LLM-as-a-judge to evaluate all knowledge-insertion methods: two LLM judges (LLaMA-3.1-70B-Instruct, GPT-OSS-20B) independently assess semantic equivalence, with disagreements (5.67\% of answers) sent to human adjudication. On a separate 500-pair sample, human agreement with the judge consensus reached $\kappa = 0.937$, indicating strong agreement.

  \begin{figure*}[t]
    \centering
	\includegraphics[width=\textwidth]{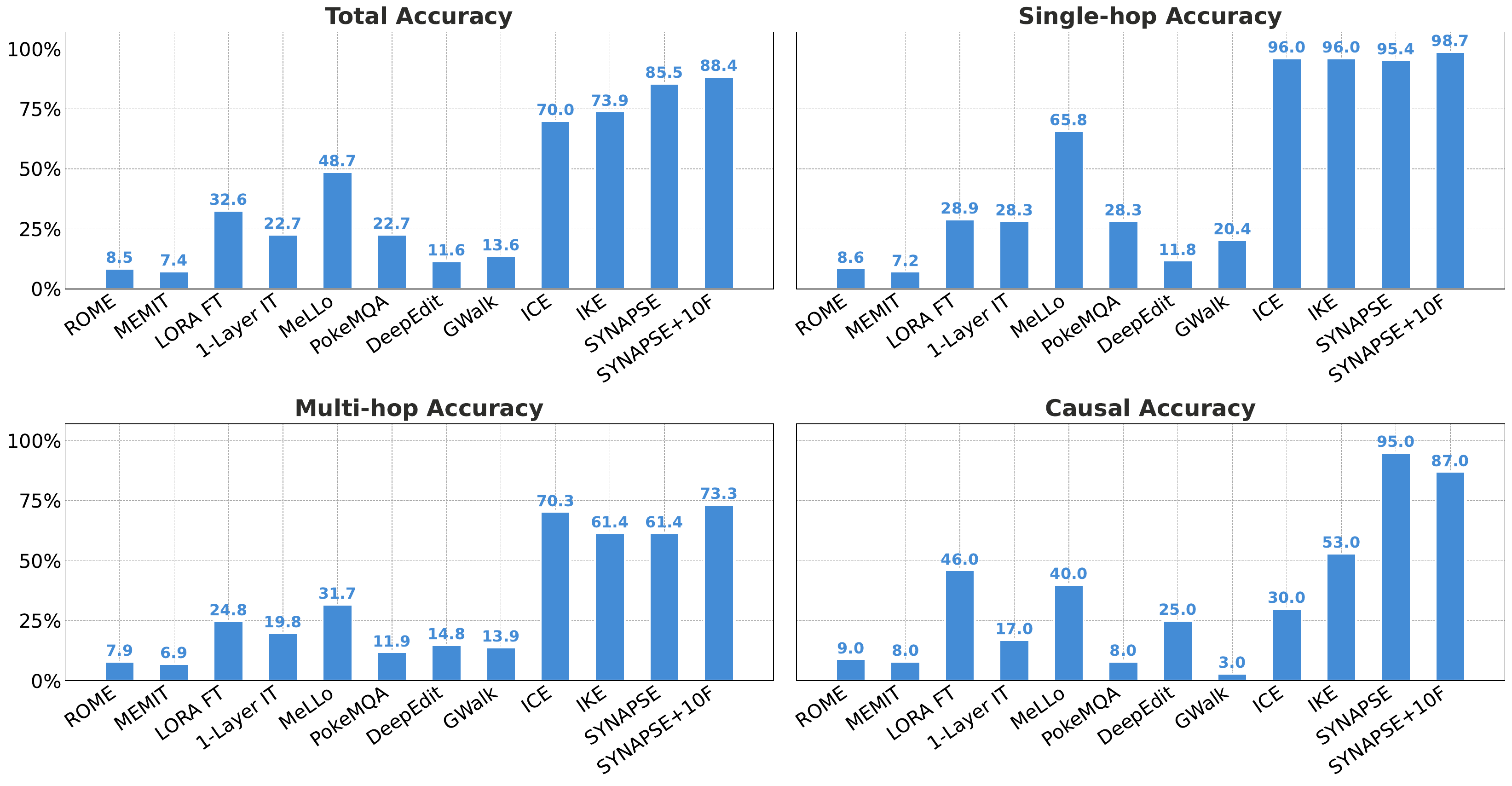}
\vspace{-20pt}
\caption{Question answering performance after inserting 150 facts into LLaMA-3.1-Instruct. Results are reported for \textbf{Total} accuracy, \textbf{Single}-hop, \textbf{Multi}-hop, and \textbf{Causal} accuracy (reasoning over event consequences), rounded to the nearest percent. {\scshape Synapse} \textbf{with 10 retrieved facts achieves the highest accuracies across all question types.}}
\label{fig:main_results_150}
    \vspace{-10pt}
\end{figure*}

\begin{table}[t]
\centering
\setlength{\tabcolsep}{10pt}
\resizebox{0.8\textwidth}{!}{%
\begin{tabular}{lccccc}
\toprule
\textbf{Model Name} & \textbf{Overall Avg.} & \textbf{Single-Hop} & \textbf{Multi-Hop} & \textbf{Causal} \\
\midrule
\multicolumn{5}{l}{\textit{LLaMA-3.1-8B-Instruct}} \\
\quad ROME \citep{meng2023locatingeditingfactualassociations} & 7.63\% & 6.72\% & 4.88\% & 13.37\% \\
\quad MEMIT	\citep{meng2023masseditingmemorytransformer} & 9.22\% & 11.01\% & 5.14\% & 12.85\% \\
% \quad PMET	\citep{li2024pmetprecisemodelediting} & \cellcolor[HTML]{F1948A}0.25\% & 0.19\% & 0.26\% & 0.26\% & 0.39\% \\
\quad WISE \citep{wang2024wiserethinkingknowledgememory} & 6.87\% & 7.46\% & 2.31\% & 12.60\%  \\
\quad AlphaEdit	\citep{fang2025alphaeditnullspaceconstrainedknowledge} & 1.53\% & 0.93\% & 1.03\% & 1.03\%  \\
\midrule
\quad {\scshape Synapse} & \underline{74.11\%}$^2$ & 87.50\% & \textit{44.22\%}$^3$ & \textbf{84.32\%}$^1$  \\
\quad {\scshape Synapse} \textbf{+ 10 Facts} & \textbf{79.83\%}$^1$ & \textbf{98.51\%}$^1$ & \textbf{54.76\%}$^1$ & \underline{76.86\%}$^2$\\

\bottomrule
\end{tabular}%
}
\caption{Results of Model Editing baselines on the 542 fact-insertion case for LLaMA-3.1-8B Instruct. Model editing methods as a whole perform poorly on {\sc ParallelEvents}.}
\vspace{-25pt}
\label{tab:model_editing}
\end{table}

\section{Baseline Implementation Details}
\label{app:baseline_implementation}

The finetuning baselines are trained using the same hyperparameters as the SFT phase of {\scshape Synapse}. Namely, we utilize 8 A40 GPUs, train for 10 epochs, use a warmup ratio of 0.06, a weight decay of 0.1, and a learning rate of $1 \times 10^{-5}$. All models are finished training within 8 hours.

For the in-context baselines, we utilize all hyperparameters in their respective setups, including using their respective retrieval models like \texttt{Facebook Contriever}. Table~\ref{tab:ike_scaling} reports the results for IKE using \{10, 32, 64, 128, 542\} retrieved facts for LLaMA-3.1-8B-Instruct and OLMo-3-7B-Instruct. We focus on the setting with 32 retrieved facts, as it achieves the highest overall accuracy across all configurations. For fast inference, all methods are generated with VLLM.

We also explore model editing with ROME \citep{mitchell2022fastmodeleditingscale}, MEMIT \citep{mitchell2022fastmodeleditingscale}, WISE \cite{wang2024wiserethinkingknowledgememory}, and AlphaEdit \cite{fang2025alphaeditnullspaceconstrainedknowledge} in the 542-fact case and report results in Table~\ref{tab:model_editing}. We adopt hyperparameters from prior work for LLaMA-3.1-8B \citep{yoon2024biggereditbatchsize}. We do not report model editing results for Olmo-3-8B-Instruct or Gemma-3-4B-Instruct due to the model’s recent release, incompatibilities with the existing software stack, and the potential time required to identify optimal hyperparameters, such as the model editing layer. We utilize the EasyEdit repository \citep{zhang2024comprehensivestudyknowledgeediting} for all of these methods.

\section{{\sc Synapse} Implementation Details}
\label{app:implementation}

For the generation of long-form text and preference datasets, {\sc Synapse} conditions only on the set of natural-language facts, similar to all other baselines. No method has access to the underlying graph in {\sc ParallelEvents}. We generate multiple news article styles and content about the event, including incremental updates, retrospective analyses, and articles with both optimistic and pessimistic tones. When generating the preference data, single-hop questions are constructed by explicitly rephrasing the facts. Causal questions in the DPO set are also generated through multiple prompts to elicit logical consequences that would emerge from an event. For multi-hop questions, we utilize sampled random walks. Utilizing the graph structure of the event, we identify relations $r_{\text{hop}}$ that are explicitly relations between entities and perform random walks on the graph to obtain fact chains. These fact chains are provided to the model to construct the multi-hop question and generate the chain-of-thought rationale in the winning response. The losing response is explicitly created by an incorrect jump in the fact chain. We explicitly define new final relations compared to the evaluation set, ensuring that the training set does not have contamination with the test questions. We also empirically verify that the questions are disjoint between the training and evaluation sets. We find that the synthetically generated questions in the DPO set overlap with evaluation questions in only 1.79\% of instances.

During synthetic dataset creation, verification filtering is conducted with the teacher model. We leverage both the knowledge graph and an evolving set of facts maintained in a knowledge base. For the long-form text, we first prompt the model to extract relevant facts so that we can append them to the knowledge base. Then, we provide the knowledge base to the model and instruct it to remove inconsistent facts and rephrase the sentences so that they remain consistent with the growing fact base. Overall, 30.89\% of paragraphs in the long-form text are rewritten to maintain alignment with the knowledge graph and knowledge base. For preference data, GPT-4.1 is prompted to provide a binary decision for whether a (question, preferred) pair is logical, given the set of event facts relevant to the question-answer pair and the fact base extracted from the long text. Any question answer pairs that are unlikely, incorrect, inconsistent (e.g., if insurance premiums increase in September 2032, no other instance may claim they remain stable), or unanswerable (e.g., if there is no reference answer to a question asking about the mayor in 2035) are filtered out. Overall, 1.01\% of causal preference pairs and 32.33\% of multi-hop preference pairs are filtered out. 

To mitigate semantic distortion and framing bias in the synthetic data, we employ a combination of diversity augmentation and post-generation validation. During generation, LLMs first produce a structured set of event parameters---such as geographic region, severity, and contextual conditions---which are evenly sampled and enforced to prevent overrepresentation of particular entities or perspectives. This includes producing balanced narratives across event outcomes (e.g., coverage of both winning and losing teams in sports events, or neutral reporting before and after disasters) and across multiple reasoning structures and question types, preventing consistent framing biases. The verification step also reduces bias by removing unverifiable or hallucinated details. To further reduce the tendency to overgeneralize or respond confidently in uncertain cases, 40\% of generated samples involve unrelated events or unanswerable queries, encouraging the model to abstain when information is insufficient.

For {\scshape Synapse}, we set attention dropout to 0.1 for both models during next-token prediction. We utilize a learning rate of $1 \times 10^{-5}$, and models are trained for 10 epochs. Additionally, we train the LM head in our experiments. We use a warmup ratio of 0.06, a weight decay of 0.1, and train using DeepSpeed. We then train the model using preference learning with a sigmoid loss, set $\beta = 0.1$, and add an auxiliary cross-entropy loss on the preferred responses with a weight of 0.1. The model is trained for 5 epochs with a learning rate of $1 \times 10^{-6}$. For Olmo-3-7B-Instruct, we increase the number of epochs to 10. Based on empirical analyses, Olmo-3-7B is more extensively instruct-tuned compared to LLaMA-3.1-8B. Consequently, behaviors such as chain-of-thought rationales in the winning response emerge more quickly in LLaMA-3.1-8B, whereas some training setups for Olmo-3-Instruct do not exhibit these behaviors even after 5 epochs. We utilize LLaMAFactory for both models, setting Olmo-3's configuration to Olmo-2 to ensure software compatibility. For preference learning, {\scshape Synapse} is first trained on our event-specific preference dataset, and then trained on the general preference dataset for realignment.

Both setups in {\scshape Synapse} are evaluated with the development set of our parallel universe question set, where we select the best-performing epoch. For the general preference set, we utilize a small sample of MMLU-Pro data as the development set. We train with either 8 A40 GPUs or 2 H100 GPUs in all fact settings. Table \ref{tab:training_stats} provides the number of tokens, generation time, and training time needed for all training splits. {\sc Synapse}'s framework relies on pre-generated synthetic data grounded in verified facts, which prioritizes safety and controlled learning. While this restricts immediate updates, data generation is relatively efficient: inserting hundreds of facts and training can be completed within ~26 hours for moderate scales as seen in the table. This offline setup enables careful study of how much data is needed for knowledge integration and abstention behavior, following prior work \cite{lambert2025tulu3pushingfrontiers, olmo2025olmo3}.

\section{Evaluation Details}
For our question-answering dataset, we utilize LLaMA-3.1-70B as the LLM-as-a-judge. All evaluations are run twice, with any disagreements being resolved through human adjudication. To determine the effectiveness of LLM-as-ajudge, we compare annotator agreement with an author and the LLM. On 500 random QA pairs, human agreement with LLM-as-a-judge achieves Cohen’s $\kappa$ = 0.937, indicating strong agreement with LLM judgments. For the abstention set, an author annotates the abstention rate. For MMLU-Pro, IFEval, and IFBench, we follow the standard evaluation methodology as presented in their paper \citep{wang2024mmluprorobustchallengingmultitask}. For PopQA, we utilize string matching with the provided answer set and its aliases, reporting question-answering accuracy. Evaluations are conducted on the full test sets of all general datasets. 

\begin{figure*}[t]
    \centering
	\includegraphics[width=\textwidth]{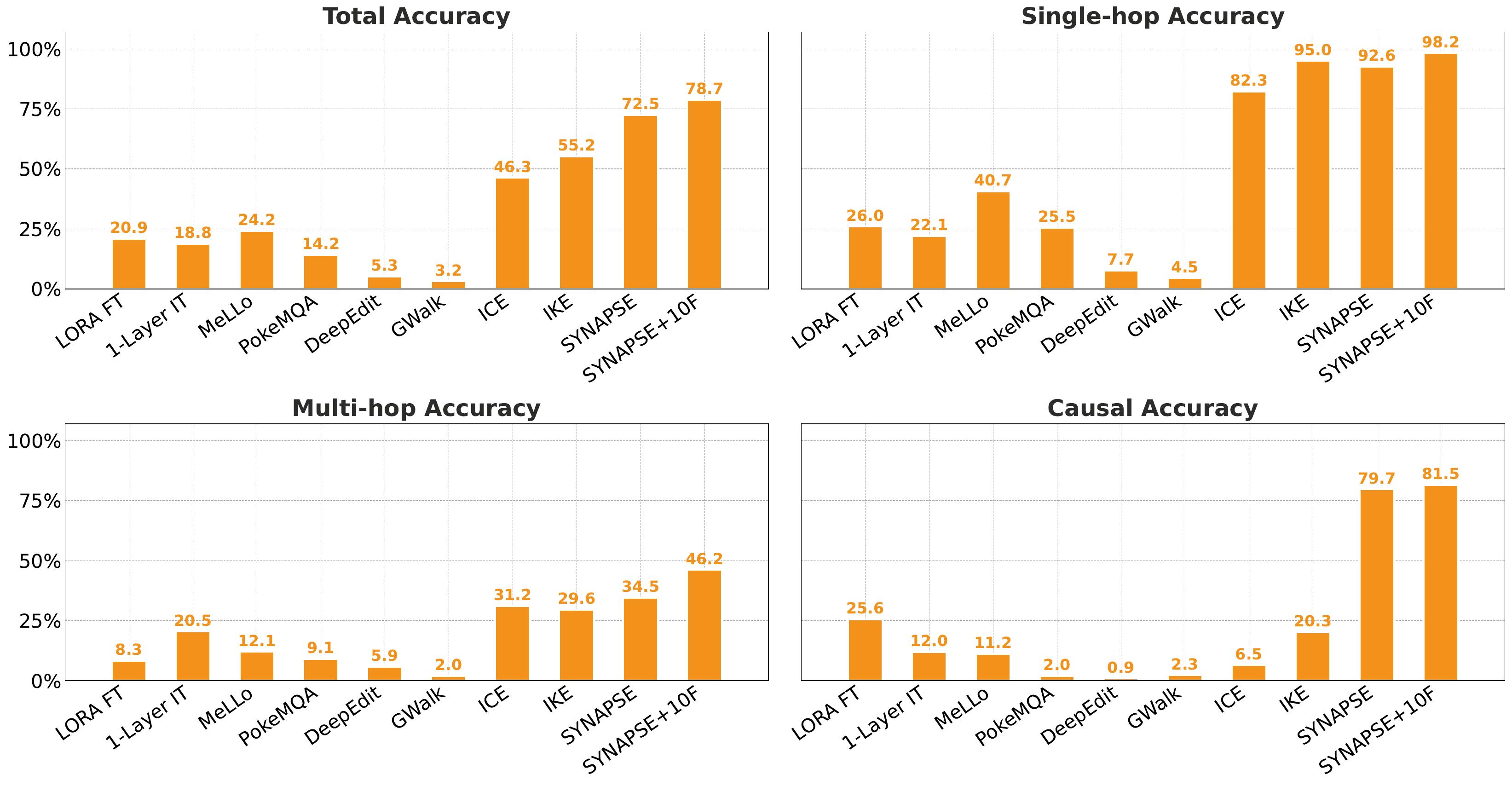}
    \vspace{-15pt}
\caption{Question answering performance after inserting 1,536 facts into LLaMA-3.1-Instruct. Results are reported for \textbf{Total} accuracy, \textbf{Single}-hop, \textbf{Multi}-hop, and \textbf{Causal} accuracy (reasoning over event consequences), rounded to the nearest percent. {\scshape Synapse} \textbf{with 10 retrieved facts achieves the highest accuracies across all question types.}}
\label{fig:main_results_1000}
    \vspace{-10pt}
\end{figure*}

\section{Additional Results}
\label{app:additional_results}

Table \ref{tab:model_editing} compares model-editing methods with {\sc Synapse} on LLaMA-3.1-8B-Instruct for the 542 fact insertion setting; we do not report results for other models due to unavailable hyperparameters. We evaluate ROME \citep{mitchell2022fastmodeleditingscale}, MEMIT \citep{mitchell2022fastmodeleditingscale}, WISE \cite{wang2024wiserethinkingknowledgememory}, and AlphaEdit \cite{fang2025alphaeditnullspaceconstrainedknowledge}. Overall, model editing performs poorly, achieving only 6.31\% total accuracy, although it attains relatively higher performance on causal questions. Inspecting outputs, we find these methods fail to incorporate new events and often produce incoherent responses, likely due to the prevalence of unseen entities in our benchmark. Because model editing modifies existing entity associations, it is largely ineffective in this setting. Model editing methods also increase hallucinations, with a low abstention rate of 4.85\% on unanswerable questions in our dataset.

Figure~\ref{fig:main_results_150} presents the full results for inserting 150 facts into LLaMA-3.1-8B. Due to the substantial computational cost of scaling experiments across all models, we report scaling results only for LLaMA-3.1-8B. Overall trends remain consistent with other settings, with {\scshape Synapse} outperforming all baselines, largely driven by significant gains in causal reasoning accuracy. {\scshape Synapse} achieves an overall accuracy of 85.55\% and a causal accuracy of 95.00\% for 150 facts. 

\begin{table}[t]
\centering
\resizebox{0.65\textwidth}{!}{
\begin{tabular}{lccccc}
\toprule
\textbf{Topic} & \textbf{Number} & \textbf{ICE} & \textbf{IKE} & \textbf{Synapse} & \textbf{Synapse+10 Facts} \\
\midrule
Disaster  & 25.54\% & 46.59\% & 62.91\% & 65.13\% & 81.90\% \\
Sport     & 35.51\% & 46.21\% & 57.10\% & 74.92\% & 80.36\% \\
City      & 17.09\% & 38.36\% & 40.13\% & 76.05\% & 70.29\% \\
Election  & 11.75\% & 53.55\% & 57.74\% & 73.87\% & 77.42\% \\
Movie     & 5.76\% & 58.55\% & 56.58\% & 65.79\% & 81.58\% \\
Crisis    & 4.36\% & 40.87\% & 46.09\% & 69.57\% & 80.00\% \\
\bottomrule
\end{tabular}
}
\caption{Topic breakdown of question answering accuracy ICE, IKE, {\sc Synapse}, and {\sc Synapse} with 10 facts for inserting 1,536 facts into the model. {\sc Synapse} shows the least amount of question answering variability across question topic.}
\vspace{-30pt}
\label{tab:topic_accuracy}
\end{table}

Performance across all methods is markedly higher than in the 542-fact insertion setting, as expected. Among the baselines, ICE and IKE perform best, achieving an average total accuracy of 71.96\%. Finetuning baselines exhibit performance comparable to graph-traversal methods, generally characterized by high abstention rates and relatively low total accuracy. In contrast, ROME and MEMIT are the worst-performing methods, with an average total accuracy of 7.94\% and an average abstention rate of 17.15\%.

For 150 facts, we also test a factual unlearning method, SimNPO \citep{fan2025simplicityprevailsrethinkingnegative}, which utilizes negative preference optimization to unlearn information while having a retain set for learned information. We utilize the dispreferred set and preferred set of {\sc Synapse} data as unlearned and retain sets, respectively. We find that model performance on {\sc ParallelEvents} is comparable to {\sc Synapse}, at 82.44\% overall, 92.76\% for single-hop, 56.44\% for multihop, and 93.00\% for causal questions. However, MMLU-Pro general accuracy decreases to 14.86\%, and the abstention rate on {\sc ParallelAbstain} is only 52.86\%, indicating unwanted side effects with this unlearning method.

Figure~\ref{fig:main_results_1000} presents the full results for inserting 1,536 facts into LLaMA-3.1-8B.  {\scshape Synapse} achieves an overall accuracy of 72.50\% accuracy and a causal accuracy of 79.7\% for 1,536 facts. Providing 10 facts in the prompt to {\sc Synapse} further improves performance across setups, increasing model performance to 78.70\% overall, 98.2\% for single-hop questions, 46.2\% for multi-hop questions, and 83.50\% for causal questions.

\begin{figure}[t]
    \centering
    \includegraphics[width=\textwidth]{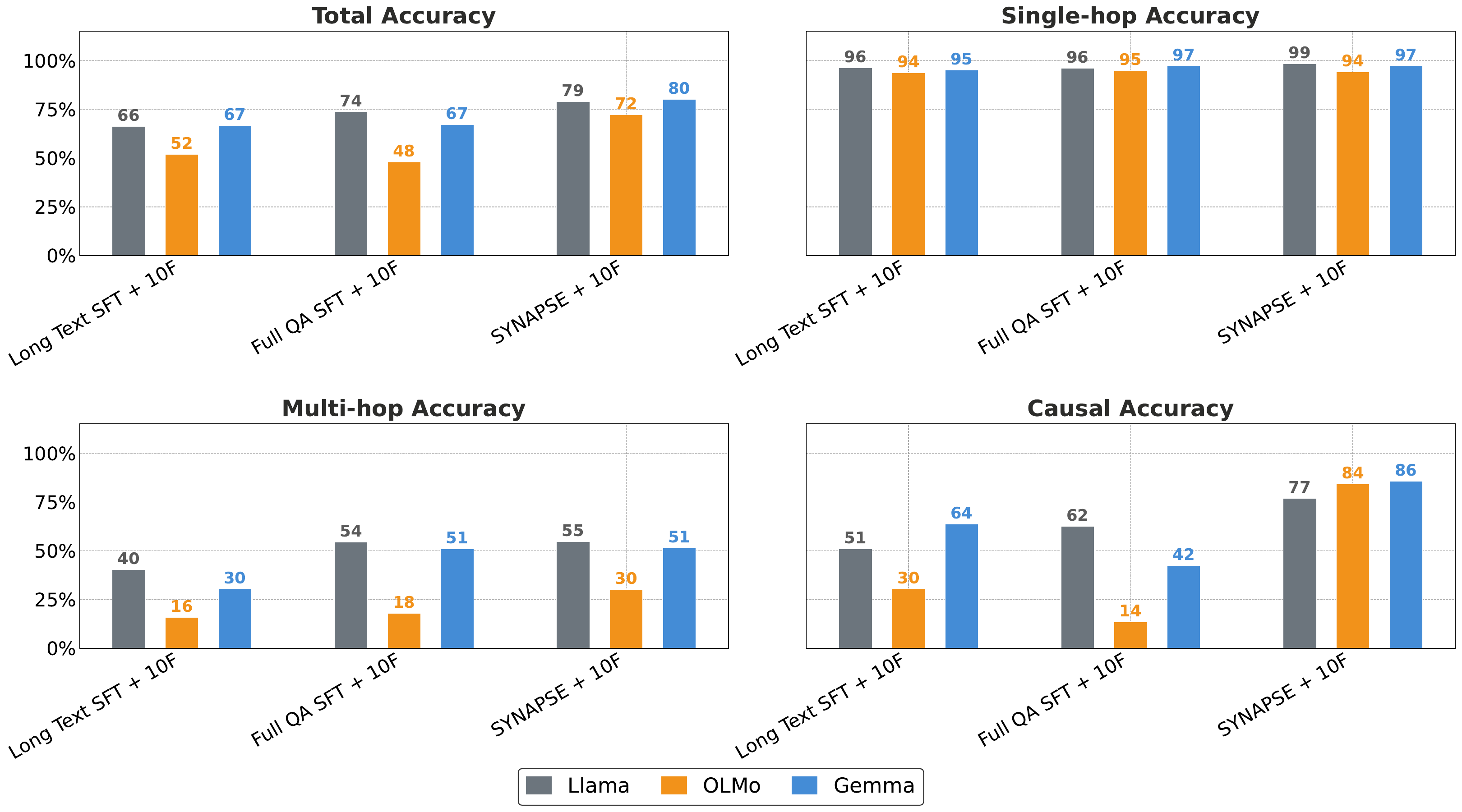}
    \caption{Accuracy of hybrid (Finetuning + RAG) methods across Llama, OLMo, and Gemma, broken down by question type. {\sc Synapse} + 10 Facts matches or exceeds the fine-tuned baselines under identical retrieval augmentation, with the largest margins on Multi-hop and Causal questions.}
    \label{fig:hybrid_results}
\end{figure}

To compare with the hybrid setup of {\sc Synapse} plus 10 retrieved facts, we additionally ran matching hybrid setups for our fine-tuned baselines (Long Text SFT and Full QA SFT), pairing each with the same retrieved facts used for {\sc Synapse}. As shown in Figure~\ref{fig:hybrid_results}, {\sc Synapse} outperforms the baseline hybrids across all three base models (Llama, OLMo, Gemma), with the largest gains concentrated in the more difficult multi-hop and causal reasoning categories. This indicates that {\sc Synapse}'s advantage is not simply subsumed by retrieval augmentation and persists when all methods are given equal access to retrieved context.

\subsection{Comparing In-Context Learning with {\sc Synapse}}

Table \ref{tab:topic_accuracy} presents question answering accuracy across ICE, IKE, {\sc Synapse}, and {\sc Synapse} with 10 retrieved facts for the 1,536 fact editing case on LLaMA-3.1-8B-Instruct. Sport questions constitute the largest share at 35.51\% of all questions, reflecting the breadth and complexity of the 2034 FIFA World Cup event. ICE and IKE exhibit substantial variability across topics, with accuracy fluctuations of 20.19\% and 22.78\%, respectively, whereas {\sc Synapse} and {\sc Synapse}+10 remain considerably more consistent, with fluctuations of only 10.92\% and 11.61\%. Across all topic categories, both {\sc Synapse} models outperform their in-context learning counterparts.

 \begin{figure*}[t]
    \centering
	\includegraphics[width=\textwidth]{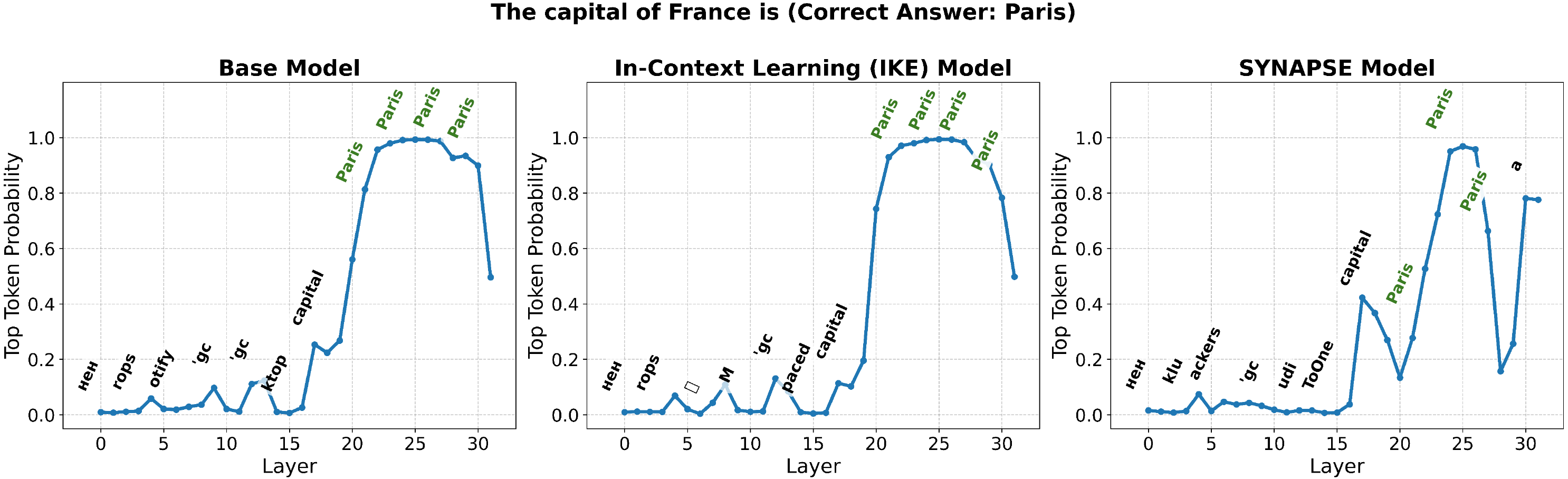}
    \includegraphics[width=\textwidth]{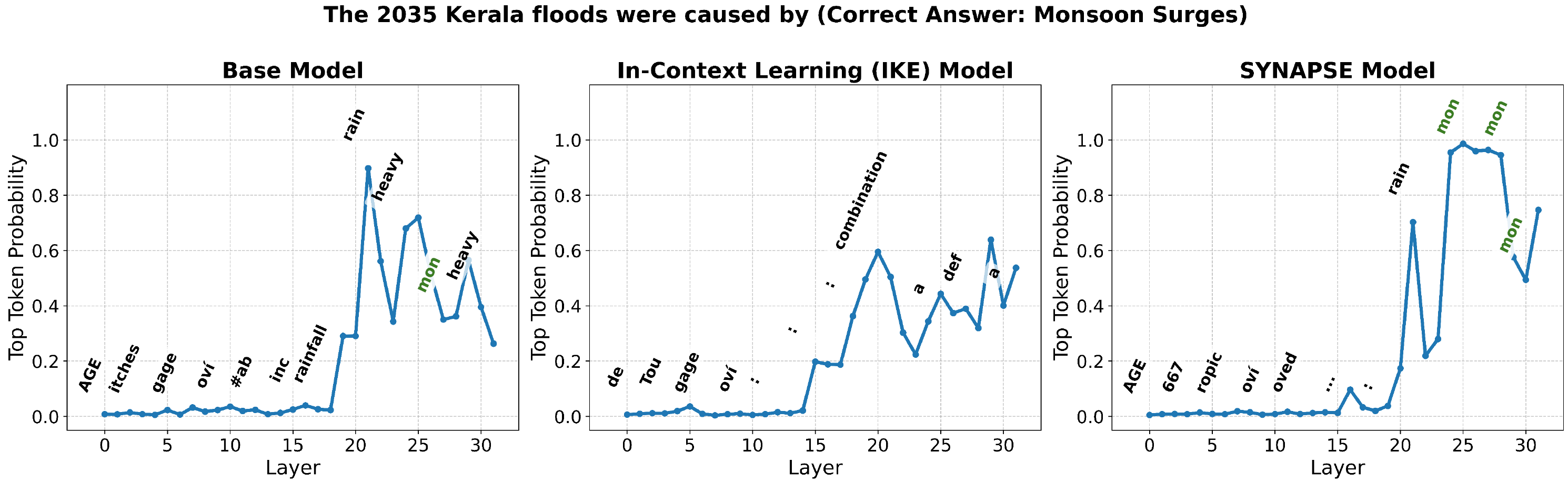}
    \includegraphics[width=\textwidth]{images/tokenprobs2.pdf}
	\caption{Token probabilities for a static question, single hop question, and causal question in the \textsc{ParallelEvents} benchmark, comparing the base model, an in-context learning model, and \textsc{Synapse}. Only \textsc{Synapse} consistently produces reasonable answers about event consequences after fact insertions.}
	\label{fig:token_probs}
    \vspace{-10pt}
\end{figure*}

We perform an error analysis comparing IKE and ICE with {\sc Synapse}. The 14.23\% gain in model performance largely comes from gains on causal questions and some multi-hop questions, as {\sc Synapse} better encodes these question dynamics. While {\sc Synapse} shows some rates of abstaining on relevant questions, ICE and IKE overwhelmingly abstain on causal questions. Among the questions where in-context retrieval performs better, 64.5\% are single-hop queries involving exact numeric recall (e.g., event attendance).These results suggest that parameter updates improve the model’s ability to reason on new facts, while retrieval-based methods remain stronger for exact fact lookup, motivating our experiments with providing {\sc Synapse} with 10 facts.

We also perform an error analysis comparing base models with {\sc Synapse} on MMLU-Pro. The largest drops ($>$5\%) occur in STEM categories (Math, Computer Science, Chemistry), which are underrepresented in {\sc Synapse}'s training distribution and involve technical content absent from synthetic event data: at least 15.61\% of computer science questions contain mathematical jargon, code syntax, or LaTeX, and 30.94\% for math questions contain such jargon. These categories degrade primarily because such technical content falls outside the distribution of {\sc Synapse}'s generated data. History also sees moderate drops ($\sim$4\%), which we address by augmenting training with synthetically generated preference pairs grounded in verified historical facts sourced from Helpsteer-2. This yields a 3.68\% accuracy gain on historical questions for LLaMA-3.1-8B while maintaining strong performance on new knowledge.

Figure \ref{fig:token_probs} presents next-token probability metrics across layers of LLaMA-3.1-8B-Instruct for the base model, IKE, and {\sc Synapse} for three question types: a static question, a single-hop question, and a causal question about the outcome of a future World Cup. For the static question, all models exhibit high token probabilities for the correct answer, with {\sc Synapse} showing some variation across layers but remaining highly confident throughout. On the single-hop question, {\sc Synapse} is the only model to answer directly with high probability, while the In-Context Learning model produces a range of plausible but uncertain candidates and the base model effectively guesses. Finally, for the causal question, the base model's outdated knowledge leads it to predict an answer such as ``Neymar,'' whereas {\sc Synapse} correctly propagates new information through to layer 20, identifying Endrick as Brazilian and assigning him high probability; IKE, by contrast, remains uncertain and inconsistent across layers, reflecting its inability to model event dynamics and causal consequences.

\subsection{Generalization}
Table~\ref{tab:generalization_150} presents results on the general benchmark for the 150-fact and 1,536-fact insertion settings using LLaMA-3.1-8B-Instruct. Consistent with the 542-fact setting, training exclusively on event data leads to substantial performance degradation across all models in both settings, with the 1,536-fact insertion setting exhibiting the greatest decline. Overall, inserting a larger number of facts correlates with increased degradation in model performance. Interestingly, the abstention rate for the 1,536 fact insertion is higher than that of the 150 fact insertion due to the increased number of data points that promote abstentions. In the 150-fact setting, augmenting training with 7{,}000 TULU-3 samples improves performance across all evaluated datasets relative to the base model, demonstrating the benefit of incorporating general preference data for broader generalization. Because the proportion of event-only data is smaller in this setting, the resulting degradation is also substantially reduced. For the 1,536-fact setting, {\sc Synapse} augmented with TULU data achieves a 9.7\% performance improvement over training solely on event data. The largest gains stem from improved instruction-following performance, underscoring the value of combining general preference datasets with instruction-following objectives. Nevertheless, despite these improvements, performance remains below that of the base model, reflecting the challenges associated with inserting facts at this scale.

\subsection{Comparison with Episodic Memory and Context Distillation Methods}
\label{app:context-distillation-comparison}

We compare {\sc Synapse} against two distinct lines of prior work for injecting new knowledge without full parametric updates to the base model's reasoning: (1) \textbf{episodic memory} frameworks, which structure temporally-organized experience explicitly and supply it to the model via prompting at inference time, rather than relying on undifferentiated context or weight updates, and (2) \textbf{context distillation} methods, which internalize in-context information into model parameters via self-generated fine-tuning data. For episodic memory, we evaluate \textbf{Panini} \citep{rajesh2026paninicontinuallearningtoken} on both raw factual sentences and full synthetic articles. For context distillation, we evaluate \textbf{Context Distillation} \citep{snell2022learningdistillingcontext}, which discards the teacher's intermediate reasoning and trains the student only on final answers, and \textbf{System2-Finetuning} \citep{park2025textitnewnewssystem2finetuning},which trains on generated questions, implications, and contextual summaries. All methods use the same teacher model and synthetic question set for the 542 factual insertion setting.

\begin{table}[t]
\centering
\small
\setlength{\tabcolsep}{2pt}
\resizebox{\textwidth}{!}{%
\begin{tabular}{l |cccc |cccc| cccc}
\toprule
& \multicolumn{4}{c|}{Llama} & \multicolumn{4}{c|}{OLMo} & \multicolumn{4}{c}{Gemma} \\
\cmidrule(lr){2-5} \cmidrule(lr){6-9} \cmidrule(lr){10-13}
Method & Total & Single & Multi & Cause & Total & Single & Multi & Cause & Total & Single & Multi & Cause \\
\midrule
GSW (Factual Sentences)     & 37.1\% & 48.3\% & 13.6\% & 45.0\% & 33.3\% & 49.6\% & 11.1\% & 33.2\% & 29.5\% & 51.7\% & 6.9\%  & 21.6\% \\
GSW (Full Articles) & 42.8\% & 55.0\% & 18.5\% & 50.1\% & 44.1\% & 50.0\% & 20.1\% & 60.2\% & 32.0\% & 44.2\% & 11.1\% & 36.0\% \\
Context Distillation        & 45.0\% & 65.3\% & 13.6\% & 48.3\% & 28.5\% & 46.6\% & 10.5\% & 21.6\% & 44.8\% & 67.7\% & 16.2\% & 41.7\% \\
System2-Finetuning          & 63.5\% & 76.7\% & 34.7\% & 74.0\% & 48.6\% & 58.2\% & 34.5\% & 49.6\% & 60.2\% & 75.2\% & 31.9\% & 67.9\% \\
\midrule
\textsc{Synapse} (ours) & \textbf{73.7\%} & \textbf{87.5\%} & \textbf{44.2\%} & \textbf{84.3\%} & \textbf{64.9\%} & \textbf{73.5\%} & \textbf{35.0\%} & \textbf{83.0\%} & \textbf{79.1\%} & \textbf{91.1\%} & \textbf{49.9\%} & \textbf{91.8\%} \\
\bottomrule
\end{tabular}%
}
\caption{Comparison of {\sc Synapse} with an episodic-memory baseline (GSW) and two context-distillation baselines (Context Distillation, System2-Finetuning), across Llama, OLMo, and Gemma. {\sc Synapse} outperforms all baselines across nearly every model and metric.}
\label{tab:gsw-system2-comparison}
\end{table}

Under this setup, Panini lags behind both fine-tuning approaches, particularly on multihop and causal questions, with a larger gap on raw factual sentences than on full synthetic articles. Context Distillation, which trains only on the teacher's final answers, performs especially poorly on multihop questions (13.45\% average vs.\ 43.02\% for {\sc Synapse}), suggesting that discarding the teacher's intermediate reasoning prevents step-by-step reasoning from being internalized when it must be grounded in external factual knowledge rather than purely procedural or self-contained tasks. This may reflect a difference in setting from the original work, where such internalization was observed. System2-Finetuning, which retains richer intermediate signals in the form of implications and contextual summaries, improves substantially over both GSW and Context Distillation but remains below {\sc Synapse} across all metrics, most notably on causal reasoning. Notably, System2-Finetuning does not consistently outperform IKE, indicating that any potential stylistic alignment with the benchmark does not drive performance on these models.

Notably, we evaluate PANINI off-the-shelf, using its default QA-pair retrieval rather than fully reconciling the underlying episodic memory over the event chain before retrieval. We did not implement this modified pipeline, so the results above should be read as a lower bound on PANINI's potential performance rather than a definitive assessment of episodic memory as a class of approaches. We view PANINI and related episodic-memory frameworks as complementary to {\sc Synapse}, and leave a full reconciliation-based comparison to future work.

\subsection{Providing {\scshape Synapse} data to retrieval baselines}
A natural question is whether {\sc Synapse}'s gains over retrieval-based baselines stem from differences in data access rather than the method itself. To isolate this, we re-ran IKE and MELLO with access to the same synthetic data used by {\sc Synapse}, including reasoning-annotated content generated during dataset construction.

Table \ref{tab:retrieval-synthetic-ablation} shows that even with equivalent data access, both baselines substantially underperform {\sc Synapse} across Llama, OLMo, and Gemma. Additionally, these methods perform worse than IKE and MeLLo without the synthetic data due to less retrieval noise. IKE + synthetic is the stronger of the two, but still trails {\sc Synapse} by a wide margin, while MeLLo + synthetic performs consistently worse. The gap is most pronounced on multi-hop and causal subsets, where both baselines struggle to exceed 20\% accuracy in most settings. This suggests that access to reasoning-annotated content alone is not sufficient to close the gap: retrieval methods struggle to elicit or leverage the provided reasoning pathways effectively, whereas {\sc Synapse} internalizes them through parameter updates.

\begin{table}[t]
\centering
\small
\setlength{\tabcolsep}{3pt}
\resizebox{\textwidth}{!}{%
\begin{tabular}{l |cccc|cccc|cccc}
\toprule
& \multicolumn{4}{c|}{Llama} & \multicolumn{4}{c|}{OLMo} & \multicolumn{4}{c}{Gemma} \\
\cmidrule(lr){2-5} \cmidrule(lr){6-9} \cmidrule(lr){10-13}
Method & Total & Single & Multi & Cause & Total & Single & Multi & Cause & Total & Single & Multi & Cause \\
\midrule
IKE + synthetic  & 53.6\% & 89.6\% & 18.5\% & 39.1\% & 32.8\% & 64.2\% & 5.4\%  & 17.0\% & 43.2\% & 86.4\% & 8.2\% & 18.5\% \\
MELLO + synthetic & 28.5\% & 36.8\% & 19.28\% & 26.5\% & 23.4\% & 22.0\% & 8.7\%  & 40.1\% & 20.5\% & 28.6\% & 8.2\% & 21.6\% \\
\bottomrule
\end{tabular}%
}
\caption{Retrieval baselines (IKE, MELLO) given access to the same synthetic data as {\sc Synapse}, including reasoning-annotated content. Even with equivalent data access, both baselines substantially underperform {\sc Synapse}, particularly on multi-hop and causal reasoning.}
\label{tab:retrieval-synthetic-ablation}
\end{table}

\section{Loss-Based Analysis}
\label{app:loss-retain-forget}

To obtain a complementary view of how knowledge updating affects new and pre-existing knowledge, we conduct a loss-based analysis of {\sc Synapse} and supervised finetuned baselines.

\paragraph{Setup.} For a fine-tuned model $\pi_\theta'$ and its base model $\pi_{\theta_0}$, we compute token-averaged NLL loss $\ell(x,y)$ under teacher forcing on two sets: (i) \textbf{new facts} — 500 QA pairs sampled from the \textsc{ParallelEvents} fine-tuning data, and (ii) \textbf{old
facts} — 300 QA pairs on real historical events disjoint from
\textsc{ParallelEvents} and never seen during fine-tuning. We define $\Delta(x,y) = \ell_{\pi_\theta'}(x,y) - \ell_{\pi_{\theta_0}}(x,y)$; negative $\Delta$ indicates acquisition (expected for new facts), positive $\Delta$
indicates forgetting (undesirable for old facts). For old facts, we restrict to the subset the base model already knew confidently (loss below the sample median, $N{=}151$), since forgetting is only meaningful for facts previously
known. We report mean $\Delta$, the fraction moving in the expected direction, and a paired Wilcoxon signed-rank test.

\begin{table}[h]
\centering
\small
\begin{tabular}{lcccc}
\toprule
& \multicolumn{2}{c}{Old facts (retention)} & \multicolumn{2}{c}{New facts (acquisition)} \\
\cmidrule(lr){2-3} \cmidrule(lr){4-5}
Method & Mean $\Delta\ell$ $\downarrow$ & \% worse $\downarrow$ & Mean $\Delta\ell$ $\uparrow$ & \% improved $\uparrow$ \\
\midrule
SFT Baseline & $+0.81$ & $94.7\%$ & $+0.09$ & $26.2\%$ \\
\textsc{Synapse} & $\mathbf{+0.47}$ & $\mathbf{68.9\%}$ & $\mathbf{-1.46}$ & $\mathbf{93.8\%}$ \\
\bottomrule
\end{tabular}
\caption{Loss-level retention (old facts, $N{=}151$) and acquisition (new
facts, $N{=}500$). All differences significant under paired Wilcoxon test
($p < 10^{-7}$).}
\label{tab:loss-retain-forget}
\end{table}

\paragraph{Discussion.} The SFT baseline's acquisition delta is near zero ($+0.09$, $26.2\%$ improved), indicating standard supervised training induces only marginal, inconsistent confidence gains on facts it was trained to insert. \textsc{Synapse} does not trade acquisition for retention: it achieves both substantially stronger acquisition ($-1.46$ vs.\ $+0.09$) and less forgetting ($+0.47$ vs.\ $+0.81$; $68.9\%$ vs.\ $94.7\%$ of facts affected). We attribute this to \textsc{Synapse}'s contrastive DPO signal, which sharpens acquisition, combined with general-preference mixing, which better anchors prior knowledge than SFT's undifferentiated objective. As a loss-level probe, this measures confidence shifts rather than answer correctness, and complements the accuracy-based evaluations in the main paper as an early-warning diagnostic for degradation.

\section{Additional Ablation Studies}
\label{app:ablation}

\begin{table}[h]
\centering
\small
\setlength{\tabcolsep}{3pt}
\resizebox{\textwidth}{!}{%
\begin{tabular}{l|cccc|cccc|cccc}
\toprule
& \multicolumn{4}{c|}{Llama} & \multicolumn{4}{c|}{OLMo} & \multicolumn{4}{c}{Gemma} \\
\cmidrule(lr){2-5} \cmidrule(lr){6-9} \cmidrule(lr){10-13}
Teacher Model & Total & Single & Multi & Cause & Total & Single & Multi & Cause & Total & Single & Multi & Cause \\
\midrule
GPT-4.1        & 85.5\% & 95.4\% & 61.4\% & 95.0\% & 80.2\% & 86.8\% & 56.4\% & 94.0\% & 86.7\% & 98.0\% & 57.4\% & 99.0\% \\
OLMo-3-32B     & 82.4\% & 94.1\% & 60.4\% & 86.0\% & 77.3\% & 86.8\% & 49.5\% & 91.0\% & 82.7\% & 95.4\% & 55.5\% & 91.0\% \\
% OLMo-3-7B      & 74.5\% & 85.5\% & 49.5\% & 83.0\% & 75.1\% & 83.6\% & 46.5\% & 91.0\% & 75.4\% & 84.21\% & 46.5\% & 91.0\% \\
IKE (baseline)  & 73.9\% & 96.0\% & 61.4\% & 53.0\% & 67.1\% & 90.8\% & 47.5\% & 51.0\% & 66.9\% & 98.0\% & 46.5\% & 40.0\% \\
\bottomrule
\end{tabular}%
}
\caption{Analysis results of changing the teacher model used for {\sc Synapse} in the 150 fact insertion setting. All teacher model setups outperform IKE, the best baseline method.}
\label{tab:teacher-150}
\end{table}

\subsection{Teacher Model Sensitivity}

To assess whether {\sc Synapse}'s gains depend on access to a strong proprietary teacher, we vary the teacher model used for synthetic data generation, replacing GPT-4.1 with the open-source OLMo-3-32B-Instruct, while holding the rest of the pipeline fixed. We evaluate under both the 150-fact and 542-fact insertion settings, comparing against IKE, the strongest retrieval baseline, on Llama, OLMo, and Gemma student models.

In the 150-fact setting (Table~\ref{tab:teacher-150}), performance degrades gradually as teacher strength decreases from GPT-4.1 to OLMo-3-32B, with the largest drops concentrated in Single-hop and Multi-hop accuracy. Notably, Causal accuracy remains high even with the smallest teacher, and {\sc Synapse} with OLMo-3-32B largely outperforms the best baseline overall on Total accuracy across all three student models. This suggests that the causal reasoning gains from synthetic generation are fairly robust to teacher strength, while multi-hop performance benefits more from a stronger teacher.

In the 542-fact setting (Table~\ref{tab:teacher-542}), we observe a similar pattern with OLMo-3-32B as a teacher: {\sc Synapse} continues to substantially outperform IKE overall, particularly on Causal and Multi-hop reasoning. These results indicate that {\sc Synapse}'s benefits are not contingent on a specific proprietary teacher, and that the causal reasoning improvements in particular are robust to teacher strength, while raw single- and multi-hop performance show the most sensitivity to teacher capability.

\begin{table}[h]
\centering
\small
\setlength{\tabcolsep}{3pt}
\resizebox{\textwidth}{!}{%
\begin{tabular}{l|cccc|cccc|cccc}
\toprule
& \multicolumn{4}{c|}{Llama} & \multicolumn{4}{c|}{OLMo} & \multicolumn{4}{c}{Gemma} \\
\cmidrule(lr){2-5} \cmidrule(lr){6-9} \cmidrule(lr){10-13}
Method & Total & Single & Multi & Causal & Total & Single & Multi & Causal & Total & Single & Multi & Causal \\
\midrule
GPT-4.1 (teacher)   & 73.7\% & 87.5\% & 44.2\% & 84.3\% & 64.9\% & 73.5\% & 35.0\% & 83.0\% & 79.1\% & 91.0\% & 49.9\% & 91.8\% \\
OLMo-32B (teacher)  & 70.5\% & 84.9\% & 44.0\% & 77.1\% & 61.6\% & 71.8\% & 32.1\% & 77.1\% & 75.3\% & 87.7\% & 45.0\% & 88.7\% \\
IKE (baseline)      & 64.2\% & 95.5\% & 44.0\% & 41.4\% & 59.6\% & 94.0\% & 28.0\% & 43.7\% & 57.8\% & 96.3\% & 33.7\% & 28.8\% \\
\bottomrule
\end{tabular}%
}
\caption{Analysis results of changing the teacher model used for {\sc Synapse} in the 542 fact insertion setting. All teacher model setups outperform IKE, the best baseline method.}
\label{tab:teacher-542}
\end{table}

\subsection{{\scshape Synapse} training ablations}
We compare our method on LLaMA-3.1-8B against several ablations to assess each component’s contribution. Specifically, we evaluate models trained with only next-token prediction, only preference optimization, instruction tuning on preferred responses from the DPO dataset, and without the general TULU preference dataset, all under the \emph{150-fact insertion setting}. Table~\ref{tab:ablation} shows that combining pre-training and post-training objectives achieves the best performance, with \textbf{85.55\%} accuracy on related questions. The full {\sc Synapse} setup with TULU also improves on general datasets and achieves an abstention rate of 82.86\%. Incorporating TULU-3 with {\scshape Facts-RL} as a general preference set notably improves multi-hop reasoning by 6.93\%. In contrast, the other variants achieve only an average accuracy of 68.75\% on {\sc ParallelEvents} with much lower abstention (e.g., 1.43\% for next-token prediction alone). 

\begin{table}[t]
\centering
\setlength{\tabcolsep}{3pt}
\resizebox{0.8\textwidth}{!}{%
\begin{tabular}{lccc ccc}
\toprule
\multirow{2}{*}{\textbf{Dataset}} &
\multicolumn{3}{c}{\textbf{150 Facts}} & \multicolumn{3}{c}{\textbf{1,536 Facts}} \\
\cmidrule(lr){2-4}\cmidrule(lr){5-7}
& Base & Event Only & {\sc Syn.}+TULU & Base & Event Only & {\sc Syn.}+TULU \\
\midrule
\multicolumn{1}{l}{\textit{Knowledge \& QA}} \\
\quad PopQA & 
\cellcolor[HTML]{EBEBEB}\textbf{32.97\%} & \cellcolor[HTML]{A9DFBF}29.04\% & \cellcolor[HTML]{A9DFBF}\textbf{33.35\%} & \cellcolor[HTML]{EBEBEB}32.97\% & \cellcolor[HTML]{A9DFBF} 29.14\% & \cellcolor[HTML]{A9DFBF} 31.82\% \\
\quad MMLU-Pro & 
\cellcolor[HTML]{EBEBEB}44.46\% & \cellcolor[HTML]{A9DFBF}41.02\% & \cellcolor[HTML]{A9DFBF}\textbf{45.11\%} & \cellcolor[HTML]{EBEBEB}\textbf{44.46\%} & 39.47\% & \cellcolor[HTML]{A9DFBF}41.83\% \\
\quad {\sc ParallelAbstain} & 
\cellcolor[HTML]{EBEBEB}\underline{82.86\%}  & 70.00\% & \cellcolor[HTML]{A9DFBF}\underline{82.86\%} & \cellcolor[HTML]{EBEBEB}81.01\% & \cellcolor[HTML]{A9DFBF}\underline{83.72\%} & \cellcolor[HTML]{A9DFBF}\underline{83.72\%} \\
\midrule
\multicolumn{1}{l}{\textit{Instruction Following}} & & & \\
\quad IFEval & 
\cellcolor[HTML]{EBEBEB}74.15\% & \cellcolor[HTML]{A9DFBF}77.03\% & \cellcolor[HTML]{A9DFBF}\textbf{77.22\%} & \cellcolor[HTML]{EBEBEB}\textbf{74.15\%} & 38.63\% & \cellcolor[HTML]{A9DFBF}72.09\% \\
\quad IFBench & 
\cellcolor[HTML]{EBEBEB}22.79\% & 10.54\% & \cellcolor[HTML]{A9DFBF}\textbf{23.81\%} & \cellcolor[HTML]{EBEBEB}\textbf{22.79\%} & 11.33\% & \cellcolor[HTML]{A9DFBF}21.33\% \\
\bottomrule
\end{tabular}%
}
\caption{Comparison of Base, {\sc Synapse}, and {\sc Synapse} with TULU variants across general datasets and models for the 150 and 1,536 fact insertion settings. \colorbox[HTML]{A9DFBF}{Green} denotes performance <4\% or better than the base model. {\sc ParallelAbstain} consists of unanswerable questions requiring abstention. {\sc Synapse} with TULU-3 recovers degradation from event-only training, improving by 6.94\% in the 150 fact insertion setting and 9.70\% on average in the 1,536 fact insertion setting. }
\vspace{-15pt}
\label{tab:generalization_150}
\end{table}

\begin{table}[t]
\setlength{\tabcolsep}{3pt}
\centering
\resizebox{0.55\textwidth}{!}{
\begin{tabular}{l|ccccc}
\toprule
\textbf{Method} & \textbf{Total} & \textbf{Single} & \textbf{Multi} & \textbf{Causal} \\
\midrule
% IKE & 73.94\% & 44.46\% & 19.66\% & \textbf{82.86\%} \\
% DPO Only           & 70.54\% & 10.58\% & -- & 52.86\% \\
% SFT Only          & 70.54\% & 38.68\% & 13.75\% & 1.43\% \\
% Instruct-Tune         & 65.16\% & 43.60\% & 0.19\% & 37.14\%  \\
% % TULU-3 IT         & - & - & -- & - \\
% {\scshape Synapse} & 82.15\% & -- & -- & 70.00\% \\\midrule
% \makecell[l]{\scshape Synapse \\ + TULU 3} & 85.55\% & 45.25\% & 23.72\% & 70.00\% \\
IKE & 73.94\% & \textbf{96.05\%} & \textbf{70.03\%}  & 30.00\%  \\
Preference Learning Only           & 70.54\% & 87.50\%  & 30.69\% & 85.00\% \\
SFT Only          & 70.54\% & 80.26\% & 42.57\% & 84.00\% \\
Instruct-Tune         & 65.16\% & 81.58\% & 24.75\% & 81.00\%  \\
% TULU-3 IT         & - & - & -- & - \\
{\scshape Synapse} Data Only & 82.15\% & 91.44\% & 54.46\% & \textbf{96.00\%} \\\midrule
\makecell[l]{\scshape Synapse \\ + TULU 3} & \textbf{85.55\%} & 95.39\% & 61.39\% & 95.00\% \\
\bottomrule
\end{tabular}
}
\caption{Ablation study of individual components in the {\scshape Synapse} pipeline, compared against IKE, the strongest in-context learning baseline. {\scshape Synapse} consistently outperforms variants trained on individual components. Including TULU-3 as a generalization dataset further improves performance.}
\vspace{-10pt}
\label{tab:ablation}
\end{table}

\subsection{Abstention Data Mixtures}
\begin{figure}[t]
    \centering
    \includegraphics[width=0.65\columnwidth]{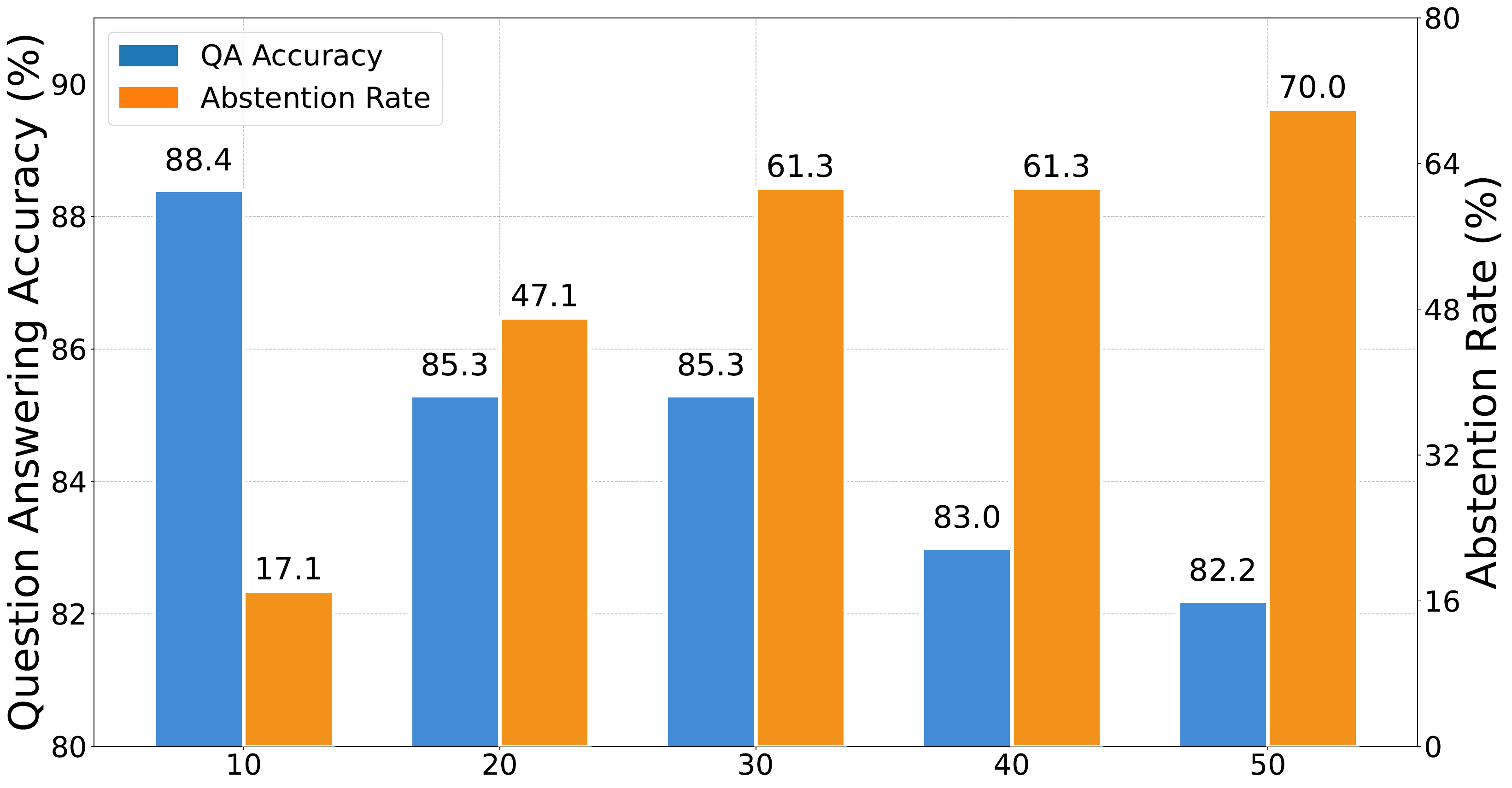}\vspace{-5pt}
\caption{Question accuracy (left axis) and abstention rate (right axis) of LLaMA-3.1-8B-Instruct when inserting 150 facts using {\scshape Synapse}, as a function of the ratio of abstention data that encourages abstention on \textbf{unknown events and entities}.}
    \label{fig:abstention_plot}
\end{figure}

We empirically investigate how much abstention data is required to effectively insert a fact into a model. Specifically, we analyze general question-answering accuracy and abstention behavior using 10\%, 20\%, 30\%, 40\%, and 50\% of data drawn from $\mathcal{D}_{\text{unk}}$, without training on TULU-3. Figure~\ref{fig:abstention_plot} presents results for LLaMA-3.1-8B in the 150-fact insertion setting. As expected, we observe an \textit{inverse relationship} between abstention and question-answering accuracy: as the proportion of abstention-focused data increases, the model abstains more frequently while overall question-answering performance declines. Notably, this degradation in accuracy is relatively modest, with only a 6.2\% decrease, whereas the abstention rate increases by nearly 53\%. This result underscores the importance of incorporating abstention data when performing parameter updates for new facts.

\begin{table}[t]
\centering
\resizebox{0.5\textwidth}{!}{
\begin{tabular}{l|cccc}
\toprule
\textbf{\# Articles} & \textbf{Total} & \textbf{Single Hop} & \textbf{Multi} & \textbf{Causal} \\
\midrule
1 & 44.76\% & 57.24\% & 11.18\% & 60.00\% \\
2 & 43.63\% & 52.63\% & 21.78\% & 52.00\% \\ 
3 & 54.39\% & 61.84\% & 21.78\% & 76.00\% \\ 
4 & 66.29\% & 71.71\% & \textbf{44.55\%} & 80.00\% \\ 
5 & 69.12\% & \textbf{85.53\% }& 33.66\% & 80.00\% \\ 
6 & \textbf{70.54\%} & 80.26\% & 42.57\% & \textbf{84.00\%} \\
\bottomrule
\end{tabular}
}
\caption{Question-answering accuracy of finetuned models trained using the next-token prediction component of {\scshape Synapse}, as a function of the number of generated articles trained per event. Performance improves as the amount of textual data per event increases, saturating around 5–6 articles.}
\label{tab:sft_mixture}
\end{table}

\begin{table*}[t]
\centering
\setlength{\tabcolsep}{3pt}
\resizebox{\textwidth}{!}{
\begin{tabular}{l|cccccc}
\toprule
\textbf{Dropout \%} &
\textbf{Total} $\uparrow$ & \textbf{Single} $\uparrow$ & \textbf{Multi} $\uparrow$ & \textbf{Causal} $\uparrow$ & \textbf{Abstain} $\uparrow$ & \textbf{MMLU-Pro} $\uparrow$
% \textbf{MMLU-Pro} &
% \textbf{AlpacaEval} &
\\
\midrule
TULU-3 $\rightarrow$ \text{{\sc Synapse}-Preference} & 81.02\% & 92.11\% & 50.50\% & 95.00\% & 70.00\% & 43.49\% \\
{\sc Synapse}-Preference $\rightarrow$ \text{TULU-3} & 85.55\% & 95.39\% & 61.39\% & 95.00\% & 70.00\% & 45.25\%  \\
\text{Helpsteer2} $\rightarrow$ \text{{\sc Synapse}-Preference} & 82.72\% & 91.45\% & 55.45\% & 97.00\% & 68.57\% & 44.32\% \\
\text{{\sc Synapse}-Preference} $\rightarrow$ \text{Helpsteer2} & 84.99\% & 94.74\% & 58.42\% & 97.00\% & 42.86\% & 42.95\% \\
Alternating Steps of {\sc Syn}-Preference and Helpsteer2 & 87.82\% & 94.74\% & 70.30\% & 95.00\% & 10.00\% & 44.07\% \\
Alternating Steps of {\sc Syn}-Preference and TULU-3 & 89.52\% & 95.39\% & 72.28\% & 98.00\% & 2.86\% & 46.08\% \\
Alternating Checkpoints of {\sc Syn}-Preference and TULU-3 & 77.90\% & 86.84\% & 48.51\% & 94.00\% & 82.86\% & 44.37\% \\\bottomrule

\end{tabular}
}
\caption{Ablation results for different training strategies using our Preference set in {\scshape Synapse} (called Event-Preference) and general preference datasets (TULU-3 and Helpsteer2) on LLaMA-3.1-8B in the 150-fact setting. The table reports total question accuracy, abstention accuracy, and MMLU-Pro accuracy, highlighting the effects of training order and alternating data on model performance. Ultimately, we opt to train models on Event-Preference first and then TULU-3 for both the 150 fact and 542 fact setting.}

\label{tab:preference_analyses}
\end{table*}

\begin{table*}[t]
\centering
\setlength{\tabcolsep}{3pt}
\resizebox{0.45\textwidth}{!}{
\begin{tabular}{l|ccccc}
\toprule
\textbf{Dropout \%} &
\textbf{Total} $\uparrow$ & \textbf{Single} $\uparrow$ & \textbf{Multi} $\uparrow$ & \textbf{Causal} $\uparrow$
% \textbf{MMLU-Pro} &
% \textbf{AlpacaEval} &
\\
\midrule
0\%  & 67.42\%  & 82.24\% & 45.55\% & 67.00\%  \\
10\%  & 70.54\% & 80.26\% & 42.57\% & 84.00\%  \\
30\% & 64.87\% & 83.55\% & 37.62\% & 64.00\% \\
50\% & 68.27\% & 82.90\% & 42.57\% & 72.00\% \\\bottomrule
\end{tabular}
}
\caption{Effect of attention dropout on LLaMA-3.1-8B-Instruct in the next-token prediction component of {\scshape Synapse}. A 10\% attention dropout rate yields the best performance.}
\label{tab:attention_dropout}
\end{table*}

\subsection{Supervised Finetuning Data Mixtures}
We report results for the next-token prediction component of {\scshape Synapse}. Specifically, we vary the number of generated articles per event used for training in the 150-fact setting for LLaMA-3.1-8B-Instruct. Table~\ref{tab:sft_mixture} summarizes the results. As expected, we observe a general trend in which increasing the number of articles per event leads to higher question-answering accuracy, although the marginal performance gain between five and six articles is substantially smaller. Overall, these results suggest that using approximately 4--6 articles per event, corresponding to roughly 4{,}000--6{,}000 words, is sufficient for inserting preliminary knowledge about events into the model when doing pre-training.

\subsection{Attention Dropout}
\label{app:attention_dropout}

We explore the impact of attention dropout on performance in the supervised finetuning setup for LLaMA-3.1-8B-Instruct. Table~\ref{tab:attention_dropout} reports the full results. We find that the best performance is achieved with a 10\% attention dropout rate, yielding a total accuracy of 70.54\% and a causal accuracy of 84.00\%. In comparison, the other settings yield much lower results.

\subsection{General Preference Datasets}

We also experiment with different general preference optimization datasets, namely Helpsteer2 \citep{wang2024helpsteer2opensourcedatasettraining}. We explore several training setups combining the general preference set and our Preference set (named {\sc Synapse}-Preference): 

\begin{itemize}
    \item Training on TULU-3/Helpsteer2 first, then on {\sc Synapse}-Preference.
    \item Training on {\sc Synapse}-Preference first, then on TULU-3/Helpsteer2.
    \item Training on both datasets by alternating the data at each gradient step.
    \item Training on {\sc Synapse}-Preference and TULU-3 while alternating the data seen across checkpoints, where each checkpoint is logged every 100 steps.
\end{itemize}

These experiments are conducted on LLaMA-3.1-8B in the 150-fact setting, and we also report MMLU-Pro accuracy and abstention rate on {\sc ParallelAbstain} to simultaneously show general performance changes.

Table~\ref{tab:preference_analyses} presents the results of these ablations. Overall, Helpsteer2 performs worse than TULU-3. While both datasets improve event question-answering accuracy, Helpsteer2 does not enhance MMLU-Pro performance as effectively as TULU-3, indicating that TULU-3 is a better general preference set for maintaining general knowledge understanding. In general, training on Event-Preference first and then on the general preference set yields the best results, achieving higher abstention accuracy while maintaining question-answering performance and improving MMLU-Pro accuracy.

Interestingly, training by alternating steps of Event-Preference and the general preference set results in very poor abstention accuracy (10\% or less). We hypothesize that when the datasets are combined at the step level, the proportion of data encouraging abstention becomes diluted, leading to lower performance. However, when alternating at the checkpoint level, the model sees enough abstention-focused data while still learning to answer relevant questions, producing the best overall results. This setup achieves a total question accuracy of 77.90\% and an abstention rate of 82.85\%. 

Despite these improvements, training with alternating checkpoints is slower and yields smaller gains in question-answering and MMLU-Pro accuracy. Therefore, we report results primarily from the setup where the model is trained on Event-Preference first, followed by TULU-3.

\newpage

\section{Licenses}
\label{appendix:license}
We list the details and licenses for the datasets and models used in this paper.

\textbf{Datasets}
\begin{itemize}
    \item MQuAKE-Remastered: \textcolor{blue}{\url{https://huggingface.co/datasets/henryzhongsc/MQuAKE-Remastered}}, Creative Commons Attribution 4.0: \textcolor{blue}{\url{https://huggingface.co/datasets/choosealicense/licenses/blob/main/markdown/cc-by-4.0.md}}
    \item MQuAKE: \textcolor{blue}{\url{https://github.com/princeton-nlp/MQuAKE}}, MIT License: \textcolor{blue}{\url{https://github.com/princeton-nlp/MQuAKE/blob/main/LICENSE}}
    \item CounterFact: \textcolor{blue}{\url{https://github.com/kmeng01/rome}}, MIT License: \textcolor{blue}{\url{https://github.com/kmeng01/rome/blob/main/LICENSE}}
\end{itemize}

\textbf{Methods}
\begin{itemize}
    \item PokeMQA: \textcolor{blue}{\url{https://github.com/Hengrui-Gu/PokeMQA}}, MIT License: \textcolor{blue}{\url{https://github.com/Hengrui-Gu/PokeMQA/blob/main/LICENSE}}
    \item DeepEdit: \textcolor{blue}{\url{https://github.com/wangywUST/DeepEdit}}, Creative Commons Attribution 4.0: \textcolor{blue}{\url{https://wangywust.github.io/deepedit.io/}}
    \item MeLLo: \textcolor{blue}{\url{https://github.com/princeton-nlp/MQuAKE}}, MIT License: \textcolor{blue}{\url{https://github.com/princeton-nlp/MQuAKE/blob/main/LICENSE}}
    \item GWalk: \textcolor{blue}{\url{https://huggingface.co/datasets/henryzhongsc/MQuAKE-Remastered}}, Creative Commons Attribution 4.0: \textcolor{blue}{\url{https://huggingface.co/datasets/choosealicense/licenses/blob/main/markdown/cc-by-4.0.md}}
    \item IKE: \textcolor{blue}{\url{https://github.com/Zce1112zslx/IKE}}, Creative Commons Attribution NonCommercial ShareAlike 4.0: \textcolor{blue}{\url{https://arxiv.org/abs/2305.12740}}
    \item ROME: \textcolor{blue}{\url{https://github.com/zjunlp/EasyEdit/tree/main/easyeditor/models/rome}}, MIT License: \textcolor{blue}{\url{https://github.com/zjunlp/EasyEdit/blob/main/LICENSE}}
    \item MEMIT: \textcolor{blue}{\url{https://github.com/zjunlp/EasyEdit/tree/main/easyeditor/models/memit}}, MIT License: \textcolor{blue}{\url{https://github.com/zjunlp/EasyEdit/blob/main/LICENSE}}
    \item WISE: \textcolor{blue}{\url{https://github.com/zjunlp/EasyEdit/tree/main/easyeditor/models/wise}}, MIT License: \textcolor{blue}{\url{https://github.com/zjunlp/EasyEdit/blob/main/LICENSE}}
    \item AlphaEdit: \textcolor{blue}{\url{https://github.com/zjunlp/EasyEdit/tree/main/easyeditor/models/alphaedit}}, MIT License: \textcolor{blue}{\url{https://github.com/zjunlp/EasyEdit/blob/main/LICENSE}}
    \item SimNPO: \textcolor{blue}{\url{https://github.com/OPTML-Group/Unlearn-Simple}}, MIT License: \textcolor{blue}{\url{https://github.com/OPTML-Group/Unlearn-Simple/blob/main/LICENSE}}
\end{itemize}

\textbf{Models}
\begin{itemize}
    \item LLaMA-3.1-Instruct: \textcolor{blue}{\url{https://huggingface.co/meta-llama/Llama-3.1-8B-Instruct}}, Llama 3.1 Community License: \textcolor{blue}{\url{https://huggingface.co/meta-llama/Llama-3.1-70B-Instruct/blob/main/LICENSE}}
    \item Olmo-3-Instruct: \textcolor{blue}{\url{https://huggingface.co/allenai/Olmo-3-7B-Instruct}}, Apache License v.2: \textcolor{blue}{\url{https://huggingface.co/allenai/Olmo-3-7B-Instruct/blob/main/README.md}}
    \item Gemma-3-IT: \textcolor{blue}{\url{https://huggingface.co/google/gemma-3-4b-it}}, Gemma License: \textcolor{blue}{\url{https://ai.google.dev/gemma/terms}}
    \item GPT-4.1: \textcolor{blue}{\url{https://developers.openai.com/api/docs/models/gpt-4.1}}, OpenAI Terms of Use: \textcolor{blue}{\url{https://openai.com/policies/terms-of-use}}

\end{itemize}

\section{Prompts}
\begin{tcolorbox}[
    colback=white, % Background color
    fontupper=\footnotesize,
    colframe=black, % Border color
    coltitle=white, % Title text color
    fonttitle=\bfseries, % Title font style
    title={Fictional Knowledge Graph Entity Generation Prompt}, % Title of the box
    sharp corners=southwest, % Sharp corners on one side
    boxrule=0.5mm, % Thickness of the border
    width=\textwidth, % Make the box fit the page width
    breakable % Allows the box to break across pages
]
Here is a reference \{entity\_type\}:
\{reference\_json\}

\vspace{10pt}

Here is an hypothetical wikidata JSON of a \{related\_entity\_type\}:
\{related\_entity\_json\}

\vspace{10pt}

Using this as a reference, create a fictional \{entity\_type\} that results from the \{related\_entity\_type\}. You must include a "has cause" entry referencing the \{related\_entity\_type\}. Format the JSON to be consistent with the example reference \{entity\_type\}. Output only the final JSON.
\end{tcolorbox}

\begin{tcolorbox}[
    colback=white, % Background color
    fontupper=\footnotesize,
    colframe=black, % Border color
    coltitle=white, % Title text color
    fonttitle=\bfseries, % Title font style
    title={Wikidata JSON Question Generation Prompt}, % Title of the box
    sharp corners=southwest, % Sharp corners on one side
    boxrule=0.5mm, % Thickness of the border
    width=\textwidth, % Make the box fit the page width
    breakable % Allows the box to break across pages
]
Here is the complete JSON of a wikidata entry for a hypothetical \{entity\_type\} (disaster / election / sporting event).
\{object\_json\}

\vspace{10pt}

Construct 5 single hop questions about this JSON. These single-hop questions should be a meaningful question, not logistical (i.e. wikidata ID or the coordinate location of the event).

\vspace{10pt}

Do not construct questions that can be easily answered from just the question itself. For instance, do not construct questions asking which country was affected by a natural disaster.

\vspace{10pt}

For existing real-world entities referenced in the JSON (e.g. a country, or an office like Prime Minister), construct questions about what is NEW as of this fictional event (e.g. the new prime minister or governing party), not static/historical facts about that entity (e.g. its capital).

\vspace{10pt}

Return the entries as a python list of tuples with the questions and answers paired together.
\end{tcolorbox}

\begin{tcolorbox}[
    colback=white, % Background color
    fontupper=\footnotesize,
    colframe=black, % Border color
    coltitle=white, % Title text color
    fonttitle=\bfseries, % Title font style
    title={Long-Form Article Generation Prompt}, % Title of the box
    sharp corners=southwest, % Sharp corners on one side
    boxrule=0.5mm, % Thickness of the border
    width=\textwidth, % Make the box fit the page width
    breakable % Allows the box to break across pages
]
Here are facts about \{subject\_description\} (a natural disaster / sporting event or figure / newly established city / election or political turning point / political leader / film).

\vspace{10pt}

\{facts\}

\vspace{10pt}

Construct a long-form article discussing this subject. The article should primarily introduce and explain \{subject\}, offering context and significance without assuming the reader already understands it.

\vspace{10pt}

While you may add broadly reasonable context (e.g. the region entered a period of growth, the team was seen as a rising contender, the campaign followed a period of tension), you must NOT hallucinate exact details not provided in the fact seed (e.g. precise statistics, vote totals, box-office numbers, named individuals not included in the facts, or specific timelines not explicitly stated).

\vspace{10pt}

You may include general \{domain\} context, institutional or community reaction, governance/competitive/industry environment, and plausible future implications. Write it in the tone of a professional feature article of roughly \{500--1000\} words.

\vspace{10pt}

The article should be in the following style: \{style\}. DO NOT USE ANY REAL NAMES UNLESS THE NAMES ARE PROVIDED IN THE FACT LIST; otherwise use [NAME] to anonymize individuals. Do not use quotes in the article.

\vspace{10pt}

Note: styles that are inherently outcome-specific (e.g. finals recaps, retrospective legacy pieces, policy breakdowns) must center on the winning/central party; broader narrative styles may cover surrounding regions, losing parties, or secondary actors when facts are provided for them.
\end{tcolorbox}

\begin{tcolorbox}[
    colback=white, % Background color
    fontupper=\footnotesize,
    colframe=black, % Border color
    coltitle=white, % Title text color
    fonttitle=\bfseries, % Title font style
    title={Long-Form Fact Extraction \& Consistency Rewrite Prompt}, % Title of the box
    sharp corners=southwest, % Sharp corners on one side
    boxrule=0.5mm, % Thickness of the border
    width=\textwidth, % Make the box fit the page width
    breakable % Allows the box to break across pages
]
Here are a list of facts established for a given event:

\{fact\_list\}
\vspace{10pt}
Here is a full article text:

\{article\}
\vspace{10pt}
Now, take the following news article paragraph:

\{paragraph\}
\vspace{10pt}
Given this paragraph, determine if this paragraph is consistent with the current set of facts. That is, determine if there are no existing contradictions that exist with this paragraph.
\vspace{10pt}
Rewrite the paragraph to address any potential issues with consistency with the **EXISTING** fact list and the overall **ARTICLE AS A WHOLE**. Keep the style of the existing paragraph. If the paragraph is fine as is, you may return the paragraph text. Return the paragraph in <paragraph></paragraph> tags.
\vspace{10pt}
Then, extract meaningful facts from this modified paragraph and add it to the established fact list such that it is a consistent fact list. Only extract the meaningful facts, such as broad day-by-day developments, effects, reactions, etc. to the fact. Do not extract individual excerpts or stories. These facts must include the dates or numbers mentioned in the fact if it contains it.
\vspace{10pt}
Return the list of newly extracted facts, in <python></python> tags as a python formatted list.
\vspace{10pt}
Additionally return yes if the paragraph is modified in its **CONTENT** or no if the paragraph remained as is in <score></score> tags.
\end{tcolorbox}

\begin{tcolorbox}[
    colback=white, % Background color
    fontupper=\footnotesize,
    colframe=black, % Border color
    coltitle=white, % Title text color
    fonttitle=\bfseries, % Title font style
    title={Single-Hop DPO Pair Generation Prompt}, % Title of the box
    sharp corners=southwest, % Sharp corners on one side
    boxrule=0.5mm, % Thickness of the border
    width=\textwidth, % Make the box fit the page width
    breakable % Allows the box to break across pages
]
Here is a fact about \{event\_type\}:
**\{FACT\_SENTENCE\}**

\vspace{10pt}

Your task is to create \{5--10\} training triples, each consisting of a question (Q), a preferred answer, and a reject (dispreferred) answer.

\vspace{10pt}

- Q must reference the new event, either (a) EXPLICITLY, by naming the event/date, or (b) IMPLICITLY, using phrasing like "latest," provided the new event is indeed the most recent of its kind. If the fact concerns basic, static information about a person (e.g. birthday, gender), Q does not need to mention the event at all.

- The preferred answer is the correct answer to Q -- the new fact -- rephrased and simplified as appropriate.

- The reject answer is incorrect for Q. Use EITHER (a) the superseded fact from a similar but older, now-outdated event, OR (b) the same fact with a deliberate error injected (wrong person, event, or time).

- All Q / preferred / reject triples must be mutually distinct, and no two reject answers may repeat.

- Both preferred and reject answers should avoid explicitly naming the new event.

\vspace{10pt}

In the preferred answer, open with: "This is a single hop fact about [ENTITY], which exists in my knowledge base. [ENTITY FACT]." The dispreferred answer must NOT include that framing sentence.

\vspace{10pt}

Respond with only a python list of tuples:
[
(Question, Prefer, Reject),
(Question, Prefer, Reject)
]
\end{tcolorbox}

\begin{tcolorbox}[
    colback=white, % Background color
    fontupper=\footnotesize,
    colframe=black, % Border color
    coltitle=white, % Title text color
    fonttitle=\bfseries, % Title font style
    title={Multi-Hop DPO Pair Generation Prompt}, % Title of the box
    sharp corners=southwest, % Sharp corners on one side
    boxrule=0.5mm, % Thickness of the border
    width=\textwidth, % Make the box fit the page width
    breakable % Allows the box to break across pages
]
Here are facts about an event.
\vspace{10pt}

\{facts\}
\vspace{10pt}

Construct 20 Direct Preference Optimization pairs of \{2-hop / 3-hop\} relations about this hypothetical event, following the provided list of hop skeletons:
\vspace{10pt}

\{skeletons\}
\vspace{10pt}

QUESTION CONSTRUCTION:

- The question must not be answerable without knowledge of the event, and must not reference any entity outside the event chain. 

- Single out entities with modifiers (the northernmost, the richest, the most populous, ...) rather than naming them directly.

- In the QUESTION, mention only the disaster TYPE and the month/year -- never the full disaster name or exact location (e.g. "2085 Earthquake" is NOT wanted).

- Follow the skeleton form: "What is the [RELATION] of the [RELATION] (of the [RELATION]) of [FINAL EVENT]?"

- Use only STATIC relations between entities (e.g. not "president of," which changes every 4 years).
\vspace{10pt}

ANSWER CONSTRUCTION -- choose ONE dispreferred mode for this batch:
\vspace{10pt}

(a) HALLUCINATED-RELATION dispreferred: both preferred and dispreferred use explicit chain-of-thought, naming each intermediate entity in turn ("Let's think about this. Out of [EVENT RELATION], the most populous country is China. The [RELATION] of China is ..."). The dispreferred answer substitutes a plausible but WRONG intermediate entity and follows it through to a wrong final answer, stated with full confidence -- never call it out as incorrect.
\vspace{10pt}

(b) ABSTENTION dispreferred: the preferred answer uses the same chain-of-thought structure with real intermediate entities. The dispreferred answer instead abstains, citing a knowledge cutoff or no record of such an event -- it must not name the event, the disaster, or any real fact from the seed, though it may offer vague, clearly speculative guesses.
\vspace{10pt}

Return a python list of tuples (question, preferred, dispreferred); both answers should be complete, multi-sentence chain-of-thought.
\end{tcolorbox}

\begin{tcolorbox}[
    colback=white, % Background color
    fontupper=\footnotesize,
    colframe=black, % Border color
    coltitle=white, % Title text color
    fonttitle=\bfseries, % Title font style
    title={Causal DPO Pair Generation Prompt}, % Title of the box
    sharp corners=southwest, % Sharp corners on one side
    boxrule=0.5mm, % Thickness of the border
    width=\textwidth, % Make the box fit the page width
    breakable % Allows the box to break across pages
]
Here are facts about an event.
\vspace{10pt}

\{facts\}
\vspace{10pt}

[Optionally: Here are causal facts already covered about this event: \{causal\_facts\} -- construct only NEW consequences not already covered.]
\vspace{10pt}

Construct 20 Direct Preference Optimization pairs of consequential relations about this hypothetical event.
\vspace{10pt}

QUESTION: phrase generally, without naming the event -- only the consequence, location, and month/year. Use either form:
\vspace{10pt}

- "Why did [CONSEQUENCE, with LOCATION] occur in [DATE]?"

- "What happened to [OBJECT OF CONSEQUENCE] in [LOCATION AND DATE]?"
\vspace{10pt}

Prefer nuanced consequences over simple death/injury counts (e.g. displacement, job loss, ecological/biodiversity change).
\vspace{10pt}

ANSWER -- choose ONE dispreferred mode for this batch:
\vspace{10pt}

(a) HALLUCINATED-NEGATION dispreferred: both preferred and dispreferred use chain-of-thought. The preferred answer reasons about the date/location, recalls the real disaster, causally connects it to the consequence in the question, and closes with a one-sentence summary. The dispreferred answer reasons similarly but concludes NO such disaster occurred and the consequence did not happen -- it must never name the disaster or use euphemisms for it ("the event," "the occurrence").
\vspace{10pt}

(b) ABSTENTION dispreferred: the preferred answer is constructed as above. The dispreferred answer instead abstains, citing a knowledge-base cutoff or no record of such a disaster, and likewise must never name it (it may offer a clearly hypothetical guess).
\vspace{10pt}

Both responses must be multiple sentences to preserve chain-of-thought reasoning. Return a python list of tuples (question, preferred, dispreferred).
\end{tcolorbox}

\begin{tcolorbox}[
    colback=white, % Background color
    fontupper=\footnotesize,
    colframe=black, % Border color
    coltitle=white, % Title text color
    fonttitle=\bfseries, % Title font style
    title={Hyperspecific Abstention DPO Pair Generation Prompt}, % Title of the box
    sharp corners=southwest, % Sharp corners on one side
    boxrule=0.5mm, % Thickness of the border
    width=\textwidth, % Make the box fit the page width
    breakable % Allows the box to break across pages
]
STEP 1 -- Question construction
\vspace{10pt}

Here are facts about an event.
\{facts\}
\vspace{10pt}

Here are existing hyperspecific questions already written for this and other events:
\{existing\_questions\}
\vspace{10pt}

Construct new single-hop, multi-hop, and causal questions about THIS event that ask for information NOT PROVIDED by the facts. Bad (answerable from the facts): "What is the highest mountain in the country where [EVENT] occurred?" Good (a genuinely unknown, overly specific detail): "What was the temperature on the day of the event?", or a multihop chain ending on an unprovided attribute, e.g. "What is the official flower of the city that suffered the second-most economic damage from [EVENT]?" Do not construct questions answerable from the facts, and do not duplicate any existing question's content. Paraphrase each question once more with different wording. Return just a python list of question strings.
\vspace{10pt}

STEP 2 -- Preferred / dispreferred construction
\vspace{10pt}

Here are the relevant facts we DO have about this event:
\{fact\_string\}
\vspace{10pt}

Here are questions about this event whose answers are NOT in the facts:

\{questions\}
\vspace{10pt}

For each question, construct a preferred and a dispreferred answer.
\vspace{10pt}

- PREFERRED: classify the question type, then explicitly name what IS known about the event (5 or so representative facts, not an exhaustive list) and abstain, stating that the knowledge base lacks the specific granularity to answer. Template: "This is a [TYPE] question about an event that occurred in [DATE] regarding [INFO TYPE]. Although I am aware that [EVENT] took place, including [KNOWN INFO TYPES], my knowledge base does not contain sufficiently detailed or granular information to determine [QUESTION INFO]. As a result, I am unable to answer this question due to the lack of specific details in my available sources."
\vspace{10pt}

- DISPREFERRED: answer the question anyway, hallucinating a plausible but fabricated specific fact, stated with full confidence.
\vspace{10pt}

Return a python list of tuples (preferred, dispreferred).
\end{tcolorbox}

\begin{tcolorbox}[
    colback=white, % Background color
    fontupper=\footnotesize,
    colframe=black, % Border color
    coltitle=white, % Title text color
    fonttitle=\bfseries, % Title font style
    title={Outside-Scope Abstention DPO Pair Generation Prompt}, % Title of the box
    sharp corners=southwest, % Sharp corners on one side
    boxrule=0.5mm, % Thickness of the border
    width=\textwidth, % Make the box fit the page width
    breakable % Allows the box to break across pages
]
STEP 1 -- Question construction
\vspace{10pt}

Here are facts about a known event, a list of already-used questions, and a list of already-used "other events":
\{facts\} / \{existing\_questions\} / \{other\_events\}
\vspace{10pt}

Construct new single-hop, multi-hop, and causal questions about COMPLETELY SEPARATE events of the same general type (e.g. another disaster, another tournament) that appear in NEITHER list and are NOT covered by our knowledge base. Split roughly half between invented events dated 2030--2035 (inside the model's training window but never actually seen) and half dated 2036 onward (after the stated knowledge cutoff of December 2035). Each question must explicitly name its invented event. Return just a python list of question strings.
\vspace{10pt}

STEP 2 -- Preferred / dispreferred construction
\vspace{10pt}

Given these questions, construct DPO pairs (question, preferred, dispreferred):
\vspace{10pt}

- PREFERRED: classify the question type, then abstain.

  - For invented events dated 2030--2035: state that the knowledge base has no record of any such event causing this phenomenon; optionally offer a vague, clearly-labeled hypothesis.

  - For invented events dated 2036 onward: state plainly that this postdates the model's knowledge cutoff (December 2035) and therefore cannot be answered.
\vspace{10pt}

- DISPREFERRED: hallucinate an answer with full confidence. Open with "Let's think about what happened in [LOCATION] in [TIME]. The [INVENTED EVENT NAME] occurred. [FABRICATED FACT]" and answer the question using that fabricated fact, never flagging it as invented.
\vspace{10pt}

Return a python list of tuples (question, preferred, dispreferred).
\end{tcolorbox}

\newpage
\section*{NeurIPS Paper Checklist}

%%% BEGIN INSTRUCTIONS %%%
The checklist is designed to encourage best practices for responsible machine learning research, addressing issues of reproducibility, transparency, research ethics, and societal impact. Do not remove the checklist: {\bf The papers not including the checklist will be desk rejected.} The checklist should follow the references and follow the (optional) supplemental material.  The checklist does NOT count towards the page
limit. 

Please read the checklist guidelines carefully for information on how to answer these questions. For each question in the checklist:
\begin{itemize}
    \item You should answer \answerYes{}, \answerNo{}, or \answerNA{}.
    \item \answerNA{} means either that the question is Not Applicable for that particular paper or the relevant information is Not Available.
    \item Please provide a short (1--2 sentence) justification right after your answer (even for \answerNA). 
   % \item {\bf The papers not including the checklist will be desk rejected.}
\end{itemize}

{\bf The checklist answers are an integral part of your paper submission.} They are visible to the reviewers, area chairs, senior area chairs, and ethics reviewers. You will also be asked to include it (after eventual revisions) with the final version of your paper, and its final version will be published with the paper.

The reviewers of your paper will be asked to use the checklist as one of the factors in their evaluation. While \answerYes{} is generally preferable to \answerNo{}, it is perfectly acceptable to answer \answerNo{} provided a proper justification is given (e.g., error bars are not reported because it would be too computationally expensive'' or ``we were unable to find the license for the dataset we used''). In general, answering \answerNo{} or \answerNA{} is not grounds for rejection. While the questions are phrased in a binary way, we acknowledge that the true answer is often more nuanced, so please just use your best judgment and write a justification to elaborate. All supporting evidence can appear either in the main paper or the supplemental material, provided in appendix. If you answer \answerYes{} to a question, in the justification please point to the section(s) where related material for the question can be found.

IMPORTANT, please:

%%% END INSTRUCTIONS %%%

\begin{enumerate}

\item {\bf Claims}
    \item[] Question: Do the main claims made in the abstract and introduction accurately reflect the paper's contributions and scope?
    \item[] Answer: \answerYes{} % Replace by \answerYes{}, \answerNo{}, or \answerNA{}.
    \item[] Justification: Results in Section \ref{sec:results} and Ablation studies in Section \ref{sec:ablation} reflects the claims made in Section \ref{sec:introduction}.

\item {\bf Limitations}
    \item[] Question: Does the paper discuss the limitations of the work performed by the authors?
    \item[] Answer: \answerYes{} % Replace by \answerYes{}, \answerNo{}, or \answerNA{}.
    \item[] Justification: Found in Appendix \ref{app:limitations}.

\item {\bf Theory assumptions and proofs}
    \item[] Question: For each theoretical result, does the paper provide the full set of assumptions and a complete (and correct) proof?
    \item[] Answer: \answerNA{} % Replace by \answerYes{}, \answerNo{}, or \answerNA{}.
    \item[] Justification: No Theoretical proofs.

    \item {\bf Experimental result reproducibility}
    \item[] Question: Does the paper fully disclose all the information needed to reproduce the main experimental results of the paper to the extent that it affects the main claims and/or conclusions of the paper (regardless of whether the code and data are provided or not)?
    \item[] Answer: \answerYes{} % Replace by \answerYes{}, \answerNo{}, or \answerNA{}.
    \item[] Justification: Section \ref{sec:method}, Section \ref{sec:experimental_details}, Appendix \ref{app:implementation}, and Appendix \ref{app:baseline_implementation} provide all the information needed to reproudce the main experimental results in the paper.

\item {\bf Open access to data and code}
    \item[] Question: Does the paper provide open access to the data and code, with sufficient instructions to faithfully reproduce the main experimental results, as described in supplemental material?
    \item[] Answer: \answerYes{} % Replace by \answerYes{}, \answerNo{}, or \answerNA{}.
    \item[] Justification: The benchmark is submitted along with the paper. All experimental details are provided to reproduce the experimental results.

\item {\bf Experimental setting/details}
    \item[] Question: Does the paper specify all the training and test details (e.g., data splits, hyperparameters, how they were chosen, type of optimizer) necessary to understand the results?
    \item[] Answer: \answerYes{} % Replace by \answerYes{}, \answerNo{}, or \answerNA{}.
    \item[] Justification: See Section \ref{sec:experimental_details} and Appendix \ref{app:implementation} for {\sc Synapse} details and Appendix \ref{app:baseline_implementation} for Baseline method implementation details.

\item {\bf Experiment statistical significance}
    \item[] Question: Does the paper report error bars suitably and correctly defined or other appropriate information about the statistical significance of the experiments?
    \item[] Answer: \answerYes{} % Replace by \answerYes{}, \answerNo{}, or \answerNA{}.
    \item[] Justification: Section \ref{sec:results} and Section \ref{sec:ablation} performs statistical significance tests on the main results.

\item {\bf Experiments compute resources}
    \item[] Question: For each experiment, does the paper provide sufficient information on the computer resources (type of compute workers, memory, time of execution) needed to reproduce the experiments?
    \item[] Answer: \answerYes{} % Replace by \answerYes{}, \answerNo{}, or \answerNA{}.
    \item[] Justification: See Section \ref{app:implementation} for the Implementation details that provide the full compute resources.
    
\item {\bf Code of ethics}
    \item[] Question: Does the research conducted in the paper conform, in every respect, with the NeurIPS Code of Ethics \url{https://neurips.cc/public/EthicsGuidelines}?
    \item[] Answer: \answerYes{} % Replace by \answerYes{}, \answerNo{}, or \answerNA{}.
    \item[] Justification: Anonymity is fully preserved, and an ethical impact statement is provided in Section \ref{sec:impact_statement}.

\item {\bf Broader impacts}
    \item[] Question: Does the paper discuss both potential positive societal impacts and negative societal impacts of the work performed?
    \item[] Answer: \answerYes{} % Replace by \answerYes{}, \answerNo{}, or \answerNA{}.
    \item[] Justification: See ethical discussion in the Impact Statement in Section \ref{sec:impact_statement}.
    
\item {\bf Safeguards}
    \item[] Question: Does the paper describe safeguards that have been put in place for responsible release of data or models that have a high risk for misuse (e.g., pre-trained language models, image generators, or scraped datasets)?
    \item[] Answer: \answerYes{} % Replace by \answerYes{}, \answerNo{}, or \answerNA{}.
    \item[] Justification: See ethical discussion in the Impact Statement in Section \ref{sec:impact_statement}.

\item {\bf Licenses for existing assets}
    \item[] Question: Are the creators or original owners of assets (e.g., code, data, models), used in the paper, properly credited and are the license and terms of use explicitly mentioned and properly respected?
    \item[] Answer: \answerYes{} % Replace by \answerYes{}, \answerNo{}, or \answerNA{}.
    \item[] Justification: See Appendix \ref{appendix:license}.

\item {\bf New assets}
    \item[] Question: Are new assets introduced in the paper well documented and is the documentation provided alongside the assets?
    \item[] Answer: \answerYes{} % Replace by \answerYes{}, \answerNo{}, or \answerNA{}.
    \item[] Justification: Section \ref{sec:parallel_universe} and Apepndix \ref{app:dataset} for the documentation of the benchmark.

\item {\bf Crowdsourcing and research with human subjects}
    \item[] Question: For crowdsourcing experiments and research with human subjects, does the paper include the full text of instructions given to participants and screenshots, if applicable, as well as details about compensation (if any)? 
    \item[] Answer: \answerNA{} % Replace by \answerYes{}, \answerNo{}, or \answerNA{}.
    \item[] Justification: There is no crowdsourcing experiments or research with human subjects.

\item {\bf Institutional review board (IRB) approvals or equivalent for research with human subjects}
    \item[] Question: Does the paper describe potential risks incurred by study participants, whether such risks were disclosed to the subjects, and whether Institutional Review Board (IRB) approvals (or an equivalent approval/review based on the requirements of your country or institution) were obtained?
    \item[] Answer: \answerNA{} % Replace by \answerYes{}, \answerNo{}, or \answerNA{}.
    \item[] Justification: There is no human subject research done i this paper.

\item {\bf Declaration of LLM usage}
    \item[] Question: Does the paper describe the usage of LLMs if it is an important, original, or non-standard component of the core methods in this research? Note that if the LLM is used only for writing, editing, or formatting purposes and does \emph{not} impact the core methodology, scientific rigor, or originality of the research, declaration is not required.
    %this research? 
    \item[] Answer: \answerYes{} % Replace by \answerYes{}, \answerNo{}, or \answerNA{}.
    \item[] Justification: Section \ref{sec:parallel_universe} and Section \ref{sec:method} describe the synthetic processes in depth that we utilize for the paper.

\end{enumerate}

\end{document}